\documentclass[letterpaper]{article} 
\usepackage[]{aaai2026}  
\usepackage{times}  
\usepackage{helvet}  
\usepackage{courier}  
\usepackage[hyphens]{url}  
\usepackage{graphicx} 
\usepackage{natbib}  
\usepackage{caption} 
\usepackage{algorithm}
\usepackage{algorithmic}
\usepackage{amsfonts}
\usepackage{amsmath} 
\usepackage{booktabs} 
\usepackage{multirow} 
\usepackage{enumitem}
\usepackage{booktabs}
\usepackage{makecell}
\usepackage{graphicx}
\usepackage{subcaption}

\usepackage{xcolor} 
\newcommand{\boldres}[1]{{\textbf{\textcolor{red}{#1}}}}
\newcommand{\secondres}[1]{{\underline{\textcolor{blue}{#1}}}}

\usepackage{newfloat}
\usepackage{listings}
\DeclareCaptionStyle{ruled}{labelfont=normalfont,labelsep=colon,strut=off} 
\floatstyle{ruled}
\newfloat{listing}{tb}{lst}{}
\floatname{listing}{Listing}
\title{OccamVTS: Distilling Vision Models to 1\% Parameters \\ for Time Series Forecasting}
\author{
    Sisuo Lyu\textsuperscript{1},
    Siru Zhong\textsuperscript{1},
    Weilin Ruan\textsuperscript{1},
    Qingxiang Liu\textsuperscript{1},\\
    Qingsong Wen\textsuperscript{2},
    Hui Xiong\textsuperscript{1},
    Yuxuan Liang\textsuperscript{1}\thanks{Corresponding author}
}
\affiliations{
    \textsuperscript{\rm 1}The Hong Kong University of Science and Technology (Guangzhou), China\\
    \textsuperscript{\rm 2}Squirrel Ai Learning, USA\\
    \{sisuolyu, siruzhong, yuxliang\}@outlook.com; xionghui@ust.hk; \\
    \{rwlinno, qingxiangliu737, qingsongedu\}@gmail.com

}

\usepackage{bibentry}

\begin{document}

\maketitle

\begin{abstract}

Time series forecasting is fundamental to diverse applications, with recent approaches leveraging large vision models (LVMs) to capture temporal patterns through visual representations. We reveal that while vision models enhance forecasting performance, 99\% of their parameters are unnecessary for time series tasks. Through cross-modal analysis, we find that time series align with low-level textural features but not high-level semantics, which can impair forecasting accuracy. We propose OccamVTS, a knowledge distillation framework that extracts only the essential 1\% of predictive information from LVMs into lightweight networks. Using pre-trained LVMs as privileged teachers, OccamVTS employs pyramid-style feature alignment combined with correlation and feature distillation to transfer beneficial patterns while filtering out semantic noise. Counterintuitively, this aggressive parameter reduction improves accuracy by eliminating overfitting to irrelevant visual features while preserving essential temporal patterns. Extensive experiments across multiple benchmark datasets demonstrate that OccamVTS consistently achieves state-of-the-art performance with only 1\% of the original parameters, particularly excelling in few-shot and zero-shot scenarios.
\end{abstract}

\textbf{Code} --- https://github.com/sisuolv/OccamVTS


\section{Introduction}

Time series forecasting (TSF) is a fundamental task in machine learning, underpinning a wide range of critical applications, including energy demand prediction, financial market analysis, meteorological modeling, and traffic flow optimization~\cite{idrees2019prediction,kiyasseh2021clocs,xu2021anomaly,bi2023accurate}. The objective of TSF is to anticipate future values based on historical observations of one or more temporally evolving variables. Despite its broad utility, TSF presents significant challenges due to inherent characteristics such as non-stationarity, long-range dependencies, stochastic noise, and the simultaneous presence of localized patterns and global trends across multiple temporal scales.

Deep learning has emerged as the dominant paradigm for TSF, with state-of-the-art models consistently achieving strong performance across diverse benchmarks. These methods operate directly on numerical sequences or their frequency-domain counterparts, enabling rich representations of complex temporal dynamics. Convolutional neural networks (CNNs) and Transformers, for example, are particularly adept at capturing both short-term fluctuations and long-term dependencies~\cite{ismail2019deep}; advanced models such as Autoformer~\cite{wu2021autoformer} and FEDformer~\cite{zhou2022fedformer} explicitly decompose sequences into trend and seasonal components~\cite{bai2018empirical,hatami2018classification,li2019logtrans,liu2022pyraformer,wen2023transformers}. 
Distinct from other modalities that rely on abstract semantics, these architectures effectively model time series as structured numerical signals, ultimately grounding predictions in precise and interpretable temporal behaviors.

\begin{figure}[t]
    \centering
    \includegraphics[width=\linewidth]{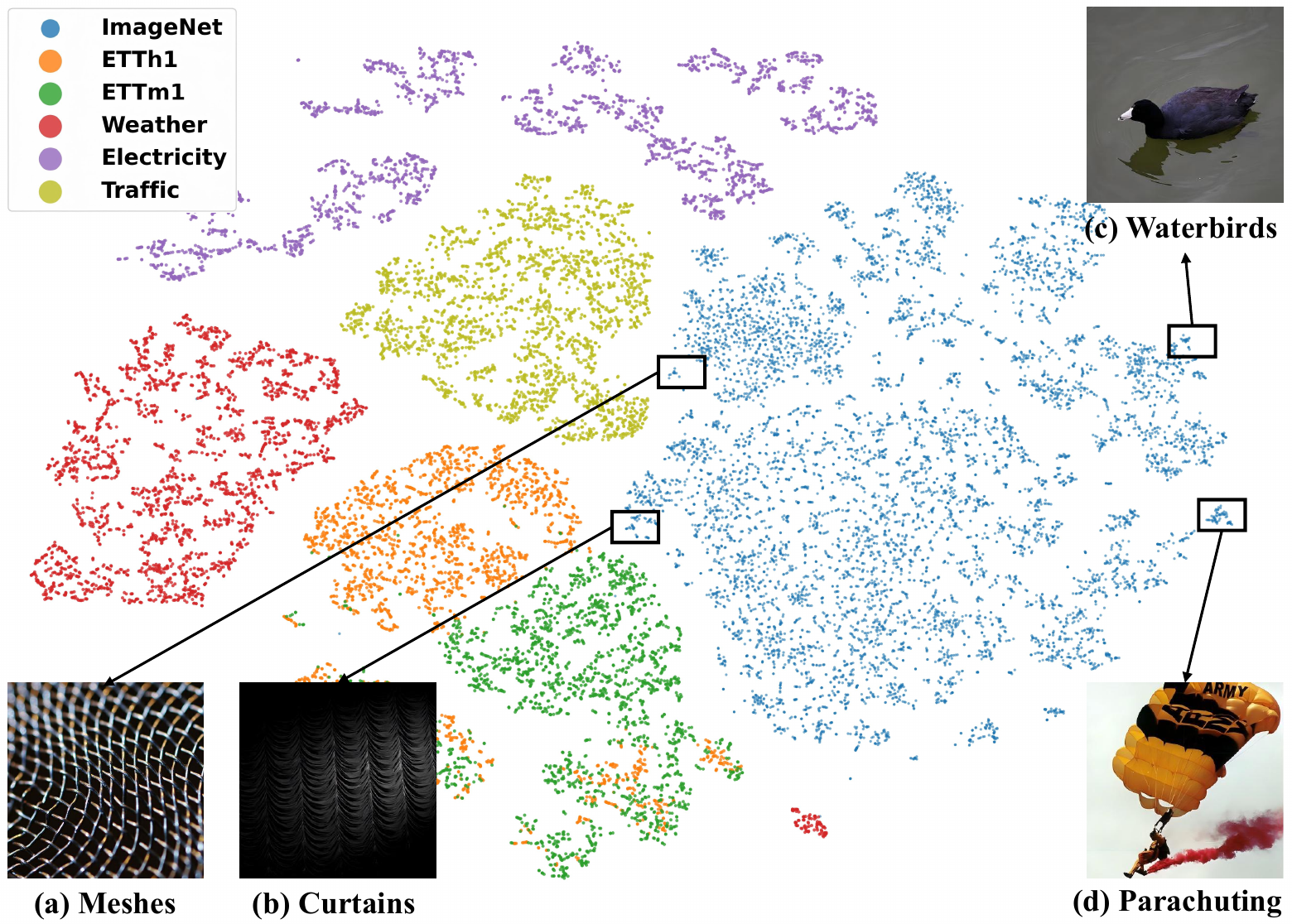}

    \caption{Modality visualization of images (ImageNet) and time series (ECL, Weather, Electricity, ETT) via the MAE encoder. (a)-(d): Original image samples extracted from the corresponding boxes in the t-SNE plot.}
    \label{fig:dataset3_1}

\end{figure}

In parallel, vision models have recently gained significant traction as an alternative and intriguing approach to TSF, driven by the observation that humans can often see meaningful patterns in time series plots. For instance, a seasoned bond trader may visually recognize a shift in trend or momentum in a price curve and form predictive judgments based on that perception. This intuitive alignment between visual understanding and temporal inference has inspired researchers to transform time series data into image-like representations, such as spectrograms, recurrence plots, or frequency-enhanced textures \cite{ni2025harnessing}, and directly leverage pre-trained vision models for downstream time series analysis tasks. Representative examples include TimesNet \cite{wu2022timesnet}, which recasts sequences into two-dimensional formats for CNN processing; VisionTS \cite{chen2024visionts}, fine-tuning Masked Autoencoders (MAE) on visualized time series \cite{he2022masked}; and TimeVLM \cite{zhong2025time}, leveraging vision-language models for multimodal forecasting. These approaches repurpose vision models' capability to detect edges, gradients, and frequency textures for identifying temporal structures and dynamics.

However, repurposing large vision models (LVMs) for TSF introduces substantial redundancy and misalignment. These models are originally architected for image data rich in semantic content -- a modality that fundamentally diverges from the purely numerical signals of time series. To investigate this misalignment and understand which visual features benefit time series forecasting, we conduct t-SNE visualization of features extracted by MAE from ImageNet~\cite{deng2009imagenet} and four time-series benchmarks (ECL, Weather, Electricity, ETT), as shown in Figure \ref{fig:dataset3_1}. The results reveal that while some visual features exhibit overlap and similarity with time series data, others show significant distributional differences between the two modalities. Fig.\ref{fig:dataset3_1} (a) and (b) lie at the interface of temporal and visual features, exhibiting similar texture patterns, while Fig.\ref{fig:dataset3_1} (c) and (d) appear in distant image clusters, demonstrating richer semantic content. Further analysis uncovers a distinct binary differentiation pattern: time-series features align closely with texture-rich images such as metal meshes and stage curtains, yet diverge markedly from semantically complex scenes like waterbirds and parachuting activities. This pattern suggests that vision models capture low-level textural features relevant to time series, while their high-level semantic representations not only add unnecessary complexity but also impair forecasting by introducing overfitting to irrelevant visual features. In limited-data scenarios, this redundancy becomes particularly detrimental as models latch onto spurious visual patterns instead of essential temporal dynamics.

Two fundamental modality mismatches underlie this problem. The first is the positional sensitivity paradox, which arises since natural images reward translation invariance, whereas time series signals critically depend on absolute temporal position. The second is the semantic representation mismatch, which stems from vision backbones that pursue high-level object semantics absent from purely numerical temporal data. These intrinsic differences manifest as three-dimensional redundancy when deploying vision models for time series forecasting: (1) Computational inefficiency from architectures over-optimized for image resolution; (2) Representational redundancy where most parameters focus on high-level semantic distinction rather than essential temporal patterns; (3) Objective misalignment where classification features directly conflict with regression tasks. This redundancy not only wastes computational resources but also risks negative transfer through semantic noise, interfering with trend prediction. These findings prompt a pressing central question: \textit{How can we retain the useful inductive biases of vision models while eliminating components that are redundant or even detrimental to TSF?}

We answer in the affirmative with \textbf{OccamVTS}: a novel cross-modal knowledge-distillation framework that transfers only the most salient \textbf{1\%} of predictive information from off-the-shelf vision models into compact forecasting networks. Drawing inspiration from Occam's razor, OccamVTS systematically prunes unnecessary complexity while preserving essential core temporal cues. Unlike existing efforts that rely on direct fine-tuning (as in VisionTS) or architectural adaptation (as in TimeVLM), our approach employs native pre-trained LVMs as privileged teachers, guiding lightweight student models via carefully designed pyramid-style feature alignment and selective distillation. This strategy differs in three critical ways: (1) It avoids inheriting architectural constraints by directly distilling from unmodified vision backbones; (2) It introduces a hierarchical pyramid-style alignment mechanism to precisely map spatial features to temporal representations; (3) It explicitly targets redundancy, with empirical results showing that removing up to 99\% of vision model parameters not only retains performance, but can actually improve accuracy by systematically mitigating overfitting to irrelevant features. Through this principled simplification, OccamVTS effectively captures frequency textures and gradient-like structures while eliminating semantic noise, ultimately achieving more with less.
\section{Related Work}

\begin{figure*}[t!]
    \centering
    \includegraphics[width=0.95\textwidth]{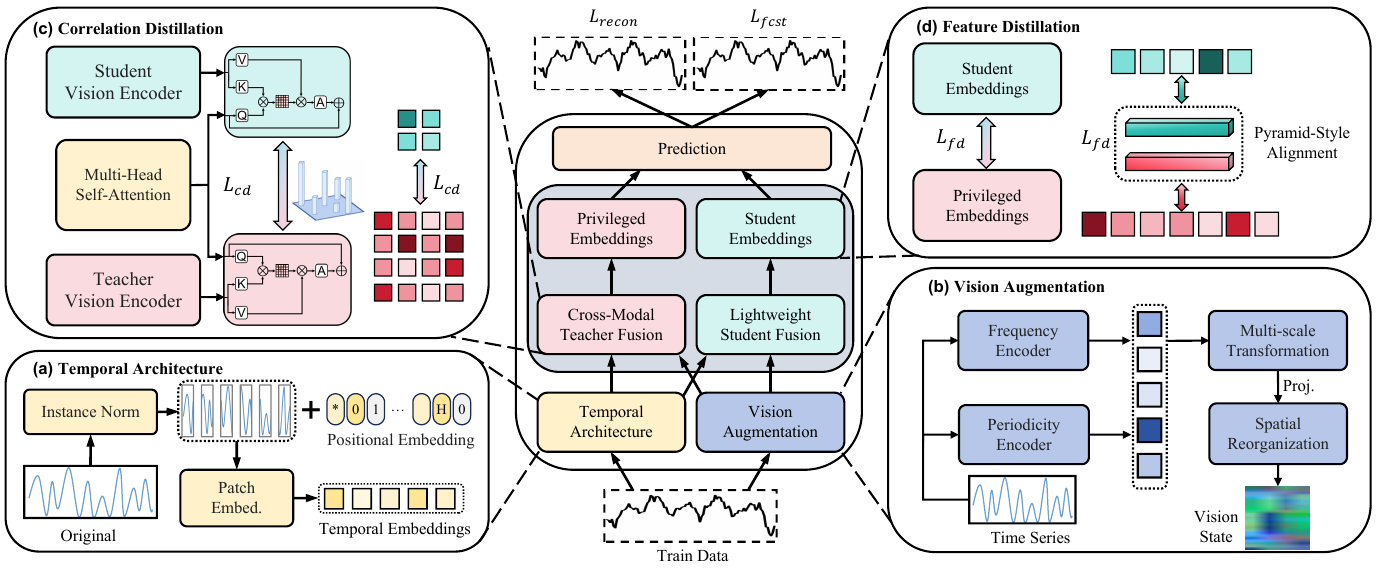}
    \caption{Overview of the OccamVTS framework.}
    \label{fig:framework}
\end{figure*}

Transforming time series data into visual representations has emerged as a popular paradigm in forecasting research. However, existing approaches have yet to systematically investigate which visual features genuinely benefit forecasting performance and which introduce harmful noise. Early efforts converted sequences into images and trained convolutional neural networks (CNNs) from scratch to perform forecasting tasks \cite{wang2015encoding,sezer2018algorithmic,li2020forecasting,sood2021visual,semenoglou2023image}. 


As this line of research has progressed, structural-level and model-level innovations have both played key roles. On the structural side, TimesNet decomposes time series into multi-periodic components and rearranges them into 2D tensors \cite{wu2023timesnet}, while TimeMixer and its successor TimeMixer++ apply mixing operations across time and feature dimensions to capture patterns at multiple scales \cite{wang2024timemixer,wang2024timemixer++}. Meanwhile, model-level adaptations harness the power of vision backbones pre-trained on natural images: BEiT was first repurposed for forecasting \cite{zhou2023one}, VisionTS demonstrated the effectiveness of Masked Autoencoders (MAE) on grayscale time-series images \cite{chen2024visionts}, TimeVLM introduced multimodal forecasting with vision-language models \cite{zhong2025time}, and LDM4TS leveraged multi-view diffusion models for enhanced prediction \cite{ruan2025vision}.

However, off-the-shelf vision backbones remain laden with parameters irrelevant to time-series data: the helpful low-level textures are often diluted by object-centric semantic features. The key challenge is to preserve these predictive cues while simultaneously shedding harmful redundancy. OccamVTS confronts this issue by selectively distilling the crucial 1\% of visual knowledge and pruning the remaining 99\%, thereby consistently boosting forecasting accuracy.

\section{Methodology}

To address the severe parameter redundancy when applying vision models to time series forecasting, we propose OccamVTS, a knowledge distillation framework that selectively transfers essential visual knowledge to lightweight temporal models. As illustrated in Figure~\ref{fig:framework}, the framework comprises three core components:

\begin{itemize} 
    \item \textbf{Cross-Modal Representation Module.} Extracts temporal features through transformer-based patch embeddings while transforming time series into visual augmentations via multi-scale convolutions and frequency encoding. This dual representation reveals complementary patterns: temporal dependencies in 1D sequences and texture-like patterns in 2D space that enhance forecasting.
    
    \item \textbf{Teacher-Student Model Module.} Implements an asymmetric design where a frozen pretrained LVM serves as teacher with only its fusion and prediction heads trained to produce privileged supervisory signals. The lightweight student encoder is jointly trained through forecasting and distillation losses to learn cross-modal patterns from the teacher. After training, only the efficient student model is retained for deployment.
    
    \item \textbf{Knowledge Distillation Module.} Transfers predictive knowledge through pyramid-style feature alignment and selective attention distillation, ensuring the student replicates only forecasting-relevant behaviors. Correlation distillation aligns temporal attention while feature distillation employs a composite objective to match fused representations. Adaptive loss weighting dynamically balances knowledge transfer with prediction accuracy.
\end{itemize}

\subsection{Cross-Modal Representation Module}

This module extracts complementary temporal and visual representations through dual parallel pathways. See Appendix~\ref{appendix:cross_modal} for complete algorithmic details.

\noindent\textbf{Temporal Feature Extraction.}  Given an input time series $x = [x_1, \dots, x_T]$, where $x\in\mathbb{R}^{B\times T\times C}$ denotes a batch of multivariate sequences, $B$ is batch size, $T$ sequence length, and $C$ the number of variables. We employ a patch-based transformer architecture. The sequence is segmented into overlapping patches of length $L$ with stride $s$, each projected into a $d$-dimensional embedding space with positional encodings. These embeddings are processed through $L_{\text{enc}}$ transformer layers:
\begin{equation}
\mathbf{h}^{(\ell)} = \text{TransformerBlock}(\mathbf{h}^{(\ell-1)}), 
\end{equation}
where $\ell = 1, \ldots, L_{\text{enc}}$, $\mathbf{h}^{(0)}$ is the patch embedding sequence, and $L_{\text{enc}}$ is the number of encoder blocks. We then pool the $T'$ patch tokens to obtain a sequence-level embedding $\mathbf{h}_{\text{pool}} \in \mathbb{R}^{B\times d_{\text{model}}}$, where $T'$ is the number of temporal patches. This pooled embedding serves as the temporal query in cross-modal fusion.

\noindent\textbf{Visual Augmentation for Time Series.} Our analysis reveals that time series data align with texture-rich visual patterns rather than high-level semantic content. This motivates a transformation pipeline that emphasizes low-level visual features to exploit pre-trained vision models:


\textit{Pattern Enhancement.} We augment the time series with frequency and periodicity features:
\begin{equation}
\mathbf{X}_{\text{aug}} = \text{concat}[\mathbf{x}, \text{FFT}(\mathbf{x}), \text{PE}(\mathbf{x})],
\end{equation}
where $\mathbf{X}_{\text{aug}}\in\mathbb{R}^{B\times L\times C\times 3}$ stacks the raw window, frequency, and periodic channels for each length-$L$ window, and we instantiate the two operators as
\begin{equation}
\begin{aligned}
\text{FFT}(x_{\text{enc}}) = \Biggl\lvert\sum_{t=0}^{L-1} x_{\text{enc}}(t) \cdot e^{-2\pi i k t / L}\Biggl\lvert,\\
\text{PE}(t) = \left[ \sin\left(\frac{2\pi t}{P}\right), \cos\left(\frac{2\pi t}{P}\right) \right] .
\end{aligned}
\end{equation}

Here $\mathbf{x}_{\text{enc}}$ denotes the normalized length-$L$ window used for spectral analysis, $t$ is the time index within a window, $k$ the frequency index, $L$ the transform length, and $P$ the dataset-specific periodicity hyperparameter (Appendix~\ref{appx:dataset_details}). In practice, we use the normalized magnitude of $\mathrm{FFT}(\cdot)$.

\textit{Multi-Scale Transformation}. The augmented features undergo hierarchical convolutions. First, lightweight depthwise–pointwise 1D blocks to capture short-range local dependencies, then shallow 2D layers on the (channel$\times$time) grid to create spatial patterns. This yields intermediate tensors $\mathbf{F}_{\text{multi-scale}}\in\mathbb{R}^{B\times C\times h\times w}$ before resizing, enabling local variations and global trends to interact as visual textures that pre-trained vision encoders are biased to detect.

\textit{Spatial Reorganization}. Features are transformed into 2D images through bilinear interpolation and normalized to $[0, 255]$, resulting in $I_{\text{visual}} \in \mathbb{R}^{B \times C \times H_{\!img} \times W_{\!img}}$:
\begin{equation}
\mathbf{I}_{\text{visual}} = \text{Normalize}(\text{Interpolate}(\mathbf{F}_{\text{multi-scale}})), 
\end{equation}
where $H_{\!img}, W_{\!img}$ denote image height/width (in pixels) after resizing and should not be confused with the forecasting horizon $H$. Concretely, for a target pixel $(x,y)$ we use bilinear interpolation

\begin{equation}
I(x, y) = \sum_{i=1}^{2}\sum_{j=1}^{2} I(x_i, y_j)\, w_{ij},
\end{equation}
where $(x_i,y_j)$ are the four nearest neighbors and $w_{ij}$ are distance-based weights; min-max normalization then scales intensities to the vision backbone’s range

\begin{equation}
I_{\text{norm}} = 255 \cdot \dfrac{I_{\text{raw}} - \min(I_{\text{raw}})}{\max(I_{\text{raw}}) - \min(I_{\text{raw}}) + \varepsilon}.
\end{equation}

This interpolation-normalization pipeline aligns pseudo-images with pre-trained vision encoders, revealing temporal patterns invisible in 1D sequences while leveraging minimal vision capabilities necessary for time series analysis.

\subsection{Teacher-Student Model Design}

\noindent\textbf{Cross-Modal Teacher Model.}
The teacher model combines the temporal features $\mathbf{h}_T$ and visual augmentation $I_{\mathrm{visual}}$ from the Cross-Modal Representation Module to produce privileged embeddings for forecasting.

\textit{Visual Feature Extraction.} The teacher employs frozen large pre-trained vision backbones $\mathcal{V}(\cdot)$ to extract visual features:
\begin{equation}
\mathbf{F}_{\mathrm{vis}}^{T} = \text{GlobalAvgPool}(\mathcal{V}(I_{\mathrm{visual}})) \cdot \mathbf{W}_{\text{proj}}^v,
\end{equation}
where $\text{GlobalAvgPool}(\cdot)$ removes spatial dimensions, $\mathbf{W}_{\text{proj}}^v$ projects the backbone output to the fusion dimension, $\mathbf{F}_{\mathrm{vis}}^{T} \in \mathbb{R}^{B\times d_{\mathrm{fus}}}$ and $d_{\mathrm{fus}}$ is the fusion dimension for aligning different modalities. We deliberately employ global pooling to prevent noisy spatial alignment between synthetic images and natural-image priors, extracting only aggregate visual features relevant to temporal patterns. Despite containing predominantly redundant parameters for time series tasks, these large models enable the teacher to explore which visual patterns genuinely benefit forecasting.

\textit{Cross-Modal Fusion and Privileged Representation Learning.}
To combine modalities, we apply cross-attention where temporal features query visual representations:
\begin{equation}
\begin{split}
    \mathbf{Q} &= \mathbf{h}_T\,\mathbf{W}_{Q}, \quad
    \mathbf{K} = \mathbf{F}_{\mathrm{vis}}^{T}\,\mathbf{W}_{K}, \quad
    \mathbf{V} = \mathbf{F}_{\mathrm{vis}}^{T}\,\mathbf{W}_{V}, \\
    \mathbf{A} & = \text{Attention}(\mathbf{Q}, \mathbf{K}, \mathbf{V}) = \mathrm{softmax}\Bigl(\frac{\mathbf{Q}\,\mathbf{K}^\top}{\sqrt{d_{k}}}\Bigr)\,\mathbf{V}.
\end{split}
\end{equation}

Here $\mathbf{W}_Q\!\in\!\mathbb{R}^{d_{\text{model}}\times d_k}$ and $\mathbf{W}_K,\mathbf{W}_V\!\in\!\mathbb{R}^{d_{\text{fus}}\times d_k}$ are learned projections, and $d_k$ is the key/query width. The fused representation combines attention output with temporal features:
\begin{equation}
    \mathbf{F}_{\mathrm{fus}} = \mathrm{LayerNorm}\bigl(\mathbf{W}_{O}\,\mathbf{A} + \mathbf{h}_T\bigr) \in \mathbb{R}^{B\times d_{\mathrm{fus}}}.
\end{equation}

\textit{Teacher Model Prediction and Supervisory Signal Generation.} 
The teacher produces forecasts through a prediction head:
\begin{equation}
\hat{\mathbf{Y}}_T = \mathbf{W}_{\text{pred}} \mathbf{F}_{\text{fus}} + \mathbf{b}_{\text{pred}},
\end{equation}
where $\hat{\mathbf{Y}}_T \in \mathbb{R}^{B \times H \times D}$, $H$ is the forecasting horizon, and $D$ is the number of predicted variables. To ensure high-quality supervisory signals, we optimize:
\begin{equation}
\mathcal{L}_{\mathrm{recon}} = \mathcal{L}_{\mathrm{SmoothL1}}(\hat{\mathbf{Y}}_T, \mathbf{Y})
\end{equation}
where $\mathbf{Y}\in\mathbb{R}^{B\times H\times D}$ denotes ground‑truth targets. Beyond predictions, the teacher provides attention matrices $\mathbf{P}_{\text{tea}}$ capturing temporal dependencies, fused representations $\mathbf{F}_{\text{fus}}$, and soft prediction targets for distillation.

\noindent\textbf{Lightweight Student Model.} 
The student processes identical inputs but uses compact vision encoders $\mathcal{V}_{\text{student}}$ achieving substantial parameter reduction:
\begin{equation}
\mathbf{F}_{\text{vis}}^S = \mathcal{V}_{\text{student}}(I_{\text{visual}}) \in \mathbb{R}^{B \times d_{\text{vis}}^S},
\end{equation}
where $d_{\text{vis}}^S$ is the student's reduced visual feature dimension.

The student employs the same cross-attention fusion mechanism as the teacher but with reduced dimensions. The key innovation is that the student's compact encoder naturally filters out semantic noise, focusing on the minimal set of visual features essential for forecasting.

During training, the student simultaneously optimizes two objectives: forecasting accuracy and knowledge acquisition from the teacher. This design validates that strategic redundancy elimination through knowledge distillation enhances rather than compromises forecasting performance.

\subsection{Knowledge Distillation Module}

This module precisely orchestrates the selective transfer of predictive knowledge from the teacher to the student, ensuring that only beneficial visual patterns are retained while redundant semantic features are effectively filtered out.

\noindent\textbf{Pyramid-Style Feature Alignment.} The dimensional and representational differences between teacher and student models necessitate sophisticated alignment strategies. We introduce a pyramid-style feature alignment mechanism that projects student features through multiple pathways:
\begin{equation}
\mathbf{F}_{\text{aligned}}^S = \sum_{i=0}^{N_s} w_i \,\phi_i\bigl(\mathbf{F}_{\text{fus}}^S\bigr),
\end{equation}
where $\phi_i$ represents projection functions operating at scale $i$, $N_s$ is the number of scales, and $w_i$ are learnable weights normalized through softmax. This multi-scale approach enables the student to accurately match teacher representations across multiple different levels of abstraction, from fine-grained local patterns to global temporal trends.

\noindent\textbf{Selective Knowledge Transfer.} Our distillation framework employs two complementary mechanisms to transfer essential knowledge while filtering redundancy:

1) \textit{Correlation Distillation.} This component encourages the student to replicate the teacher's temporal dependency patterns. Let $P_{\mathrm{tea}}^{(i)}, P_{\mathrm{stu}}^{(i)} \in \mathbb{R}^{T'\times T'}$ denote the attention matrices for the $i$-th sample, where $T'$ is the number of temporal patches. We align these matrices via temperature-scaled KL divergence:
\begin{equation}
\mathcal{L}_{\mathrm{cd}}
= \frac{\tau^2}{B} \sum_{i=1}^{B}
\mathrm{D}_{KL}\left(\sigma\left(\frac{P_{\mathrm{tea}}^{(i)}}{\tau}\right) \bigg|\bigg| \sigma\left(\frac{P_{\mathrm{stu}}^{(i)}}{\tau}\right)\right),
\label{eq:correlation_distillation}
\end{equation}
where $\sigma(\cdot)$ denotes the softmax operator and $\tau$ is an adaptive temperature parameter that controls the smoothness of attention distributions, flexibly allowing the student to learn both sharp and smoothly distributed attention patterns.

2) \textit{Feature Distillation.} This component aligns the student's fused representations with the teacher's privileged embeddings. We employ a composite loss combining multiple perspectives:
\begin{equation}
\mathcal{L}_{\mathrm{fd}}
= \lambda_{\mathrm{MSE}} \cdot \mathcal{L}_{\mathrm{MSE}} + \lambda_{\mathrm{cos}} \cdot \mathcal{L}_{\mathrm{cosine}} + \lambda_{\mathrm{KL}} \cdot \mathcal{L}_{\mathrm{KL}},
\label{Lfd}
\end{equation}
where $\mathcal{L}_{\mathrm{MSE}}$ measures direct feature similarity, $\mathcal{L}_{\mathrm{cosine}}$ captures semantic relationships through:
\begin{equation}
\mathcal{L}_{\mathrm{cosine}} = \textbf{1} - \frac{\mathbf{F}_{\mathrm{fus}}^{T} \cdot \mathbf{F}_{\mathrm{fus}}^{S}}{||\mathbf{F}_{\mathrm{fus}}^{T}|| \cdot ||\mathbf{F}_{\mathrm{fus}}^{S}||},
\end{equation}
and $\mathcal{L}_{\mathrm{KL}}$ effectively aligns output distributions through adaptive temperature scaling while preserving sharpness.

\noindent\textbf{Training Objectives and Optimization.} The complete distillation objective combines correlation and feature components:
\begin{equation}
\mathcal{L}_{\mathrm{distill}}
= \lambda_{cd}\mathcal{L}_{\mathrm{cd}}
+ \lambda_{fd}\mathcal{L}_{\mathrm{fd}},
\end{equation}
where $\lambda_{cd}$ and $\lambda_{fd}$ are implemented as learnable parameters $\boldsymbol{\lambda} = \exp(\boldsymbol{\theta}_{\lambda})$, with $\boldsymbol{\theta}_{\lambda}$ being neural network parameters optimized through gradient descent. This adaptive weighting eliminates manual hyperparameter tuning.

\textit{Student's Total Objective.} During training, the student minimizes:
\begin{equation}
\mathcal{L}_{\mathrm{student}}
= \underbrace{\mathcal{L}_{\mathrm{fcst}}}_{\text{forecasting loss}}
\;+\;\underbrace{\lambda_{\mathrm{distill}}\;\mathcal{L}_{\mathrm{distill}}}_{\text{distillation alignment loss}},
\end{equation}
where $\mathcal{L}_{\mathrm{fcst}} = \mathcal{L}_{\mathrm{SmoothL1}}(\hat{Y}_{S}, Y)$ is the student's forecasting loss, and $\lambda_{\mathrm{distill}}$ adaptively balances between independent forecasting and cross-modal knowledge acquisition.

Through this carefully designed distillation framework, the student learns to replicate the teacher's beneficial behaviors while discarding redundant features. The adaptive weighting and temperature mechanisms ensure effective knowledge transfer throughout training, enabling the student to achieve comparable or superior performance with dramatically fewer parameters. See Appendix~\ref{appendix:adaptive_kd} for details.

\subsection{Contributions at a glance}

Overall, our methodological innovations are threefold: we design a texture-oriented cross-modal representation that converts time series into pseudo-images through frequency and periodicity enhancement and a lightweight 1D$\rightarrow$2D multi-scale transformation, followed by spatial reorganization and global pooling to robustly couple with the low-level inductive biases of pre-trained vision backbones; we propose an asymmetric teacher--student architecture that freezes the vision teacher and performs cross-attention fusion where temporal queries attend to aggregated visual keys and values to form privileged representations; and we introduce a scalable selective knowledge transfer scheme that combines pyramid-style feature alignment with two complementary objectives, namely correlation distillation on temporal attention and distillation at the representation level, so that the student preserves forecasting relevant textures while suppressing semantic redundancy and noise.
\section{Experiments}

\begin{table*}[h!]
\captionsetup{font=small}
\begin{center}
\begin{small}
\scalebox{0.65}{
\setlength\tabcolsep{4pt}
\begin{tabular}{c|cc|cc|cc|cc|cc|cc|cc|cc|cc|cc|cc|cc|cc}
\toprule

\multicolumn{1}{c|}{Methods}&\multicolumn{2}{c|}{Ours}& \multicolumn{2}{c|}{Only Teacher} & \multicolumn{2}{c|}{Only Student}&\multicolumn{2}{c|}{TimeVLM}&\multicolumn{2}{c|}{TimeMixer++}&\multicolumn{2}{c|}{TimeMixer}&\multicolumn{2}{c|}{LDM4TS}&\multicolumn{2}{c|}{TimesNet}&\multicolumn{2}{c|}{iTransformer}&\multicolumn{2}{c|}{DLinear}&\multicolumn{2}{c|}{PatchTST}&\multicolumn{2}{c|}{FEDformer}&\multicolumn{2}{c}{Autoformer} \\

\midrule

\multicolumn{1}{c|}{Metric} & MSE  & MAE & MSE & MAE & MSE & MAE & MSE & MAE & MSE  & MAE & MSE  & MAE & MSE  & MAE & MSE  & MAE & MSE  & MAE & MSE  & MAE & MSE  & MAE & MSE  & MAE & MSE  & MAE\\
\midrule

$ETTh1$ & \boldres{0.403} & \secondres{0.421} & 0.416 & 0.433 & 0.434 & 0.444  & \secondres{0.405} & \boldres{0.420} & 0.419 & 0.432 & 0.447 & 0.440 & 0.443 & 0.454 & 0.458 & 0.450 & 0.454 & 0.447 & 0.422 & 0.437 & 0.450 & 0.449 & 0.440 & 0.460 & 0.496 & 0.487 \\
\midrule
$ETTh2$ & \boldres{0.336} & \secondres{0.383} & \secondres{0.338} & 0.388 & 0.342 & 0.394 & 0.341 & 0.391 & 0.339 & \boldres{0.380} & 0.365 & 0.395 & 0.387 & 0.427 & 0.414 & 0.427 & 0.383 & 0.407 & 0.431 & 0.446 & 0.382 & 0.411 & 0.437 & 0.449 & 0.450 & 0.459 \\
\midrule
$ETTm1$ & \boldres{0.347} & \boldres{0.373} & 0.354 & \secondres{0.377} & 0.355 & 0.377 & \secondres{0.347} & 0.377 & 0.369 & 0.378 & 0.381 & 0.396 & 0.352 & 0.387 & 0.400 & 0.406 & 0.407 & 0.410 & 0.357 & 0.378 & 0.388 & 0.402 & 0.448 & 0.452 & 0.588 & 0.517 \\
\midrule
$ETTm2$ & \boldres{0.245} & \boldres{0.307} & 0.252 & 0.313 & 0.258 & 0.317 & \secondres{0.248} & \secondres{0.311} & 0.269 & 0.320 & 0.275 & 0.323 & 0.333 & 0.380 & 0.291 & 0.333 & 0.288 & 0.332 & 0.267 & 0.333 & 0.293 & 0.336 & 0.305 & 0.349 & 0.327 & 0.371 \\
\midrule
$Weather$ & \boldres{0.224} & \boldres{0.259} & 0.229 & 0.268  & 0.230 & 0.269 & \secondres{0.224} & 0.263 & 0.226 & \secondres{0.262} & 0.240 & 0.272 & 0.229 & 0.277 & 0.259 & 0.287 & 0.258 & 0.278 & 0.248 & 0.300 & 0.258 & 0.280 & 0.309 & 0.360 & 0.338 & 0.382 \\
\midrule
$ECL$ & \boldres{0.162} & \secondres{0.259} & 0.168 & 0.267 & 0.170 & 0.270 & 0.172 & 0.272 & \secondres{0.165} & \boldres{0.253} & 0.182 & 0.273 & 0.199 & 0.299 & 0.192 & 0.304 & 0.178 & 0.270 & 0.166 & 0.263 & 0.204 & 0.294 & 0.214 & 0.327 & 0.227 & 0.338 \\
\midrule
$Traffic$ & \boldres{0.407} & \secondres{0.279} & \secondres{0.415} & 0.292 & 0.419 & 0.297 & 0.419 & 0.298 & 0.416 & \boldres{0.264} & 0.485 & 0.298 & 0.550 & 0.321 & 0.620 & 0.336 & 0.428 & 0.282 & 0.433 & 0.295 & 0.482 & 0.308 & 0.610 & 0.376 & 0.628 & 0.379 \\

\bottomrule
\end{tabular}
}
\end{small}
\end{center}

\caption{Long-term forecasting results. Results are averaged over forecasting horizons $H \in $\{96, 192, 336, 720\}. Lower values indicate better performance.\boldres{Red}: best, \secondres{Blue}: second best. Full results see Appendix~\ref{appx:long-term-forecasting}.}
\label{tab:long-term-forecasting-avg}

\end{table*}

\begin{table*}[h!]
\captionsetup{font=small}
\begin{center}
\begin{small}
\scalebox{0.65}{
\setlength\tabcolsep{4pt}
\begin{tabular}{c|cc|cc|cc|cc|cc|cc|cc|cc|cc|cc|cc|cc|cc}
\toprule

\multicolumn{1}{c|}{Methods}&\multicolumn{2}{c|}{Ours}& \multicolumn{2}{c|}{Only Teacher} & \multicolumn{2}{c|}{Only Student} &\multicolumn{2}{c|}{TimeVLM}&\multicolumn{2}{c|}{TimeMixer++}&\multicolumn{2}{c|}{TimeMixer}&\multicolumn{2}{c|}{LDM4TS}&\multicolumn{2}{c|}{TimesNet}&\multicolumn{2}{c|}{iTransformer}&\multicolumn{2}{c|}{DLinear}&\multicolumn{2}{c|}{PatchTST}&\multicolumn{2}{c|}{FEDformer}&\multicolumn{2}{c}{Autoformer} \\

\midrule

\multicolumn{1}{c|}{Metric} & MSE  & MAE & MSE & MAE & MSE & MAE & MSE  & MAE & MSE  & MAE & MSE  & MAE & MSE  & MAE & MSE  & MAE & MSE  & MAE & MSE  & MAE & MSE  & MAE & MSE  & MAE & MSE  & MAE\\
\midrule

$ETTh1$ & \boldres{0.422} & \boldres{0.439} & 0.443 & 0.456 & 0.446 & 0.458 & \secondres{0.431} & \secondres{0.442} & 0.517 & 0.512 & 0.613 & 0.520 & 0.471 & 0.468 & 0.869 & 0.628 & 0.518 & 0.488 & 0.691 & 0.600 & 0.633 & 0.542 & 0.639 & 0.561 & 0.702 & 0.596 \\
\midrule
$ETTh2$ & \boldres{0.344} & \boldres{0.390} & \secondres{0.356} & 0.402 & 0.357 & 0.403 & 0.356 & 0.402 & 0.379 & \secondres{0.391} & 0.402 & 0.433 & 0.452 & 0.460 & 0.479 & 0.465 & 0.428 & 0.438 & 0.605 & 0.538 & 0.415 & 0.431 & 0.466 & 0.475 & 0.488 & 0.499 \\
\midrule
$ETTm1$ & \boldres{0.356} & \boldres{0.379} & 0.364 & 0.387 & 0.365 & 0.387 & \secondres{0.360} & \secondres{0.382} & 0.398 & 0.431 & 0.487 & 0.461 & 0.371 & 0.393 & 0.677 & 0.537 & 0.447 & 0.432 & 0.411 & 0.429 & 0.501 & 0.466 & 0.722 & 0.605 & 0.802 & 0.628 \\
\midrule
$ETTm2$ & \boldres{0.253} & \boldres{0.313} & \secondres{0.261} & \secondres{0.321} & 0.262 & 0.322  & 0.263 & 0.323 & 0.291 & 0.351 & 0.311 & 0.367 & 0.336 & 0.373 & 0.320 & 0.353 & 0.295 & 0.338 & 0.316 & 0.368 & 0.296 & 0.343 & 0.463 & 0.488 & 1.342 & 0.930 \\
\midrule
$Weather$ & \boldres{0.227} & \boldres{0.262} & 0.230 & \secondres{0.268} & 0.231 & 0.269 & 0.233 & 0.274 & 0.241 & 0.271 & 0.242 & 0.281 & \secondres{0.229} & 0.276 & 0.279 & 0.301 & 0.272 & 0.290 & 0.241 & 0.283 & 0.242 & 0.279 & 0.284 & 0.324 & 0.300 & 0.342 \\
\midrule
$ECL$   & 0.181 & 0.283 & 0.206 & 0.310 & 0.209 & 0.312 & 0.188 & 0.291 & \boldres{0.168} & \boldres{0.271} & 0.187 & 0.277 & \secondres{0.172} & 0.275 & 0.323 & 0.392 & 0.202 & 0.288 & 0.180 & 0.280 & 0.180 & \secondres{0.273} & 0.346 & 0.427 & 0.431 & 0.478 \\
\midrule
$Traffic$  & 0.460 & 0.332 & 0.531 & 0.385 & 0.536 & 0.390 & 0.484 & 0.357 & 0.483 & 0.315 & 0.536 & 0.349 & 0.621 & 0.357 & 0.951 & 0.535 & 0.470 & 0.318 & \secondres{0.447} & \secondres{0.313} & \boldres{0.430} & \boldres{0.305} & 0.663 & 0.425 & 0.749 & 0.446 \\

\bottomrule
\end{tabular}
}
\end{small}
\end{center}

\caption{Few-shot learning on 10\% training data. We use the same protocol in Table~\ref{tab:long-term-forecasting-avg}. Full results see Appendix~\ref{appx:few-shot-forecasting}.}
\label{tab:few-shot-forecasting-10per-avg}

\end{table*}

\subsection{Experimental Settings}

\noindent\textbf{Datasets \& Evaluation Metrics.} We evaluate OccamVTS on eight benchmark datasets: ETTh1, ETTh2, ETTm1, ETTm2, Weather, Electricity, Traffic~\cite{zhou2021informer, lai2018modeling}, and M4~\cite{makridakis2018m4}. Performance is measured using MAE and MSE for the first seven datasets, while M4 uses SMAPE, MASE, and OWA following competition protocols \cite{oreshkin2019n}. Dataset details and metric specifications are in Appendix~\ref{appx:dataset_details} and~\ref{appx:evaluation_metrics}.

\noindent\textbf{Baselines.} We extensively compare OccamVTS with state-of-the-art time series models and ablation variants (teacher-only and student-only configurations), including recent vision-augmented methods like TimeVLM \cite{zhong2025time}, LDM4TS \cite{ruan2025vision}, TimeMixer++ \cite{wang2024timemixer++}, TimeMixer \cite{wang2024timemixer}, and TimesNet \cite{wu2022timesnet}; transformer-based architectures like iTransformer \cite{liu2023itransformer}, PatchTST \cite{nie2022time}, FEDformer \cite{zhou2022fedformer}, ETSformer \cite{woo2022etsformer}, Non-Stationary Transformer \cite{liu2022non}, Autoformer \cite{wu2021autoformer}, and Informer \cite{zhou2021informer}; and highly competitive linear models like DLinear \cite{zeng2023transformers} and LightTS \cite{campos2023lightts}.

\noindent\textbf{Implementation Details.} We use a unified benchmarking pipeline~\cite{wu2022timesnet} with a knowledge distillation framework. 
From the teacher pool (MAE variants, CLIP, EfficientNet\mbox{-}B3, ResNet\mbox{-}101) and the student pool (EfficientNet\mbox{-}B0, MobileNet\mbox{-}V3, Tiny\mbox{-}ViT), we instantiate one teacher and one student per experiment; unless otherwise noted, we adopt MAE\mbox{-}Large as the teacher and Tiny\mbox{-}ViT as the student, which corresponds to \textbf{$\approx$1\%} of the teacher's total parameters. Models are trained using Adam optimizer with learning rate $10^{-3}$ on NVIDIA RTX A6000 GPU (48GB). See Appendix~\ref{appx:optimization_settings} for details.

\subsection{Long-term Forecasting}

We evaluate the long-term forecasting capabilities of our model across seven benchmark datasets and compare against a wide range of state-of-the-art baselines. As shown in Table~\ref{tab:long-term-forecasting-avg}, our approach consistently outperforms state-of-the-art baseline methods across all datasets. On the ETTh2 dataset, OccamVTS achieves a 12.0\% MSE reduction compared to PatchTST, demonstrating significant improvements in capturing long-term temporal dependencies. The advantages become more pronounced in high-dimensional scenarios, where we achieve 1.8\% improvement over TimeMixer++ on the Electricity dataset and 4.9\% improvement over iTransformer on the Traffic dataset. Crucially, the knowledge distillation variant consistently outperforms its non-distilled counterpart across every benchmark, with improvements ranging from 2.3\% to 8.1\%, demonstrating that knowledge distillation(KD) is a key driver of performance gains. Even without KD, our vision-enhanced architecture achieves second-best performance on multiple datasets, validating the effectiveness of cross-modal temporal modeling.

\subsection{Few-shot Forecasting}

To evaluate the data efficiency and generalization capability of our proposed method, we conducted comprehensive few-shot learning experiments using only 10\% of the training data. As shown in Table~\ref{tab:few-shot-forecasting-10per-avg}, our method achieved the best performance on five out of seven datasets (ETTh1, ETTh2, ETTm1, ETTm2, and Weather) in terms of both MSE and MAE metrics. On the ETTh1 dataset, we obtain 2.1\% improvement over TimeVLM, while on ETTm2, our approach achieves significant 4\% MSE reduction compared to the second-best performer. The ablation experiments in the few-shot setting further validate the importance of the knowledge distillation mechanism, where the complete method consistently outperforms the variant without knowledge distillation by 3.0-4.6\% across all datasets. This indicates that the knowledge distillation component is particularly valuable when training data is limited, as it enables more effective cross-modal knowledge transfer from pre-trained vision models to lightweight temporal architectures.

\subsection{Zero-shot Forecasting}
We conduct zero-shot transfer experiments across ETT datasets without any fine-tuning to evaluate cross-domain generalization capabilities. As shown in Table~\ref{tab:zero-shot-avg}, OccamVTS achieves the best performance in 5 MSE and 7 MAE metrics out of 8 scenarios, demonstrating strong cross-domain transferability. For challenging transfer tasks like \texttt{ETTh2->ETTh1} and \texttt{ETTm2->ETTm1}, our approach achieves 13.5\% and 6.7\% MSE improvements over TimeVLM respectively, significantly outperforming other baseline methods. The ablation study confirms that knowledge distillation consistently improves zero-shot performance across all transfer scenarios, with particularly notable improvements of 5.4\% on \texttt{ETTh2->ETTh1} and 10.7\% on \texttt{ETTm2->ETTm1} when comparing KD versus non-KD variants. While most baseline methods show significant performance degradation (20-40\%) in cross-dataset transfers, our method maintains consistent performance with only ±8\% variation across different transfer pairs, suggesting robust generalization capabilities through effective distillation of universal temporal patterns from vision models.
\begin{table*}[h!]
\begin{center}
\captionsetup{font=small} 
\begin{small}
\setlength\tabcolsep{4pt}
\scalebox{0.65}{
\begin{tabular}{c|cc|cc|cc|cc|cc|cc|cc|cc|cc|cc|cc|cc}
\toprule
\multicolumn{1}{c|}{Methods}&\multicolumn{2}{c|}{Ours}&\multicolumn{2}{c|}{Only Teacher} & \multicolumn{2}{c|}{Only Student}&\multicolumn{2}{c|}{TimeVLM}&\multicolumn{2}{c|}{TimeMixer++}&\multicolumn{2}{c|}{TimeMixer}&\multicolumn{2}{c|}{LDM4TS}&\multicolumn{2}{c|}{TimesNet}&\multicolumn{2}{c|}{iTransformer}&\multicolumn{2}{c|}{DLinear}&\multicolumn{2}{c|}{PatchTST}&\multicolumn{2}{c}{Autoformer}\\
\midrule
\multicolumn{1}{c|}{Metric} & MSE & MAE & MSE & MAE & MSE & MAE & MSE & MAE & MSE & MAE & MSE & MAE & MSE & MAE & MSE & MAE & MSE & MAE & MSE & MAE & MSE & MAE & MSE & MAE\\
\midrule
$ETTh1$ $\rightarrow$ $ETTh2$ & \secondres{0.342} & \boldres{0.385} & 0.351 & 0.396 & 0.350 & 0.395 & \boldres{0.338} & \secondres{0.385} & 0.367 & 0.391 & 0.427 & 0.424 & 0.458 & 0.452 & 0.421 & 0.431 & 0.384 & 0.404 & 0.493 & 0.488 & 0.380 & 0.405 & 0.582 & 0.548\\
\midrule
$ETTh1$ $\rightarrow$ $ETTm2$ & \secondres{0.295} & \boldres{0.350} & 0.300 & 0.355 & 0.301 & 0.356 & \boldres{0.293} & \secondres{0.350} & 0.301 & 0.357 & 0.361 & 0.397 & 0.369 & 0.400 & 0.327 & 0.361 & 0.337 & 0.374 & 0.415 & 0.452 & 0.314 & 0.360 & 0.457 & 0.483\\
\midrule
$ETTh2$ $\rightarrow$ $ETTh1$ & \boldres{0.429} & \boldres{0.446} & \secondres{0.453} & \secondres{0.466} & 0.532 & 0.508 & 0.496 & 0.480 & 0.511 & 0.498 & 0.679 & 0.577 & 0.723 & 0.577 & 0.865 & 0.621 & 0.657 & 0.563 & 0.703 & 0.574 & 0.565 & 0.513 & 0.757 & 0.608\\
\midrule
$ETTh2$ $\rightarrow$ $ETTm2$ & \boldres{0.285} & \boldres{0.343} & \secondres{0.288} & \secondres{0.346} & 0.295 & 0.352 & 0.297 & 0.353 & 0.329 & 0.370 & 0.342 & 0.378 & 0.432 & 0.444 & 0.342 & 0.376 & 0.336 & 0.374 & 0.328 & 0.386 & 0.325 & 0.365 & 0.366 & 0.411\\
\midrule
$ETTm1$ $\rightarrow$ $ETTh2$ & \secondres{0.357} & \secondres{0.398} & 0.359 & 0.400 & 0.359 & 0.399 & \boldres{0.354} & \boldres{0.397} & 0.417 & 0.422 & 0.452 & 0.441 & 0.452 & 0.434 & 0.457 & 0.454 & 0.443 & 0.443 & 0.464 & 0.475 & 0.439 & 0.438 & 0.470 & 0.479\\
\midrule
$ETTm1$ $\rightarrow$ $ETTm2$ & \boldres{0.259} & \boldres{0.315} & \secondres{0.262} & \secondres{0.319} & 0.263 & 0.319 & 0.264 & 0.319 & 0.291 & 0.331 & 0.329 & 0.357 & 0.354 & 0.367 & 0.322 & 0.354 & 0.301 & 0.337 & 0.335 & 0.389 & 0.296 & 0.334 & 0.469 & 0.484\\
\midrule
$ETTm2$ $\rightarrow$ $ETTh2$ & \boldres{0.357} & \boldres{0.394} & 0.366 & 0.402 & 0.364 & 0.403 & \secondres{0.359} & \secondres{0.399} & 0.432 & 0.443 & 0.413 & 0.427 & 0.494 & 0.474 & 0.435 & 0.443 & 0.457 & 0.456 & 0.455 & 0.471 & 0.409 & 0.425 & 0.423 & 0.439\\
\midrule
$ETTm2$ $\rightarrow$ $ETTm1$ & \boldres{0.403} & \boldres{0.410} & 0.451 & 0.442 & 0.437 & 0.434 & 0.432 & \secondres{0.426} & \secondres{0.427} & 0.448 & 0.554 & 0.478 & 0.588 & 0.487 & 0.769 & 0.567 & 0.719 & 0.546 & 0.649 & 0.537 & 0.568 & 0.492 & 0.755 & 0.591\\
\bottomrule
\end{tabular}}
\end{small}
\end{center}

\caption{Zero-shot learning results. We use the same protocol in Table~\ref{tab:long-term-forecasting-avg}. Full results see Appendix~\ref{appx:zero-shot}.}
\label{tab:zero-shot-avg}

\end{table*}

\begin{table*}[h!]
\renewcommand\arraystretch{1.2}
\captionsetup{font=small} 

\begin{center}
\begin{small}
\scalebox{0.65}{
\setlength\tabcolsep{3pt}
\begin{tabular}{c|cccccccccccccccc}
\toprule
Methods& Ours &Only Teacher &Only Student &TimeVLM&Timemixer++&Timemixer&TimesNet&iTransformer&DLinear&PatchTST&ETSformer&LightTS&FEDformer&Stationary&Autoformer&Informer \\
\midrule
SMAPE&12.050&12.205 & 12.222 &\boldres{11.894}&\secondres{11.905}&11.947&12.880&12.684&13.639&12.059&14.718&13.525&13.160&12.780&12.909&14.086\\
MASE&\secondres{1.611}&1.642 &1.643 &\boldres{1.592}&1.611&1.614&1.836&1.764&2.095&1.623&2.408&2.111&1.775&1.756&1.771&2.718\\
OWA&0.866&0.879 & 0.880 &\boldres{0.855}&\secondres{0.860}&0.862&0.955&0.929&1.051&0.869&1.172&1.051&0.949&0.930&0.939&1.230\\
\bottomrule
\end{tabular}
}
\end{small}
\end{center}

\caption{Short-term time series forecasting results (Average). The forecasting horizons are in [6, 48] and the results are weighted averaged from all datasets under different sampling intervals. Full results see Appendix~\ref{appx:short-term}.}
\label{tab:short-term-forecasting}

\end{table*}

\subsection{Short-term Forecasting}

To evaluate performance on short-term prediction tasks, we conduct experiments with forecasting horizons ranging from 6 to 48 time steps across multiple datasets. As shown in Table~\ref{tab:short-term-forecasting}, our method consistently demonstrates 1.3\% improvement in SMAPE compared to traditional approaches and maintains significant consistent advantages over the variant without knowledge distillation, with improvements of 1.4-2\% across all metrics. This validates the effectiveness of our knowledge distillation mechanism even in short-term forecasting scenarios, where the distilled visual features help capture subtle fine-grained temporal patterns that might be missed by pure temporal models. The results suggest that OccamVTS maintains remarkably robust temporal modeling capabilities across different prediction horizons, from short-term operational forecasting to long-term strategic planning.

\subsection{Model Analysis}

\begin{figure}[!b]
    \centering
    \includegraphics[width=\linewidth]{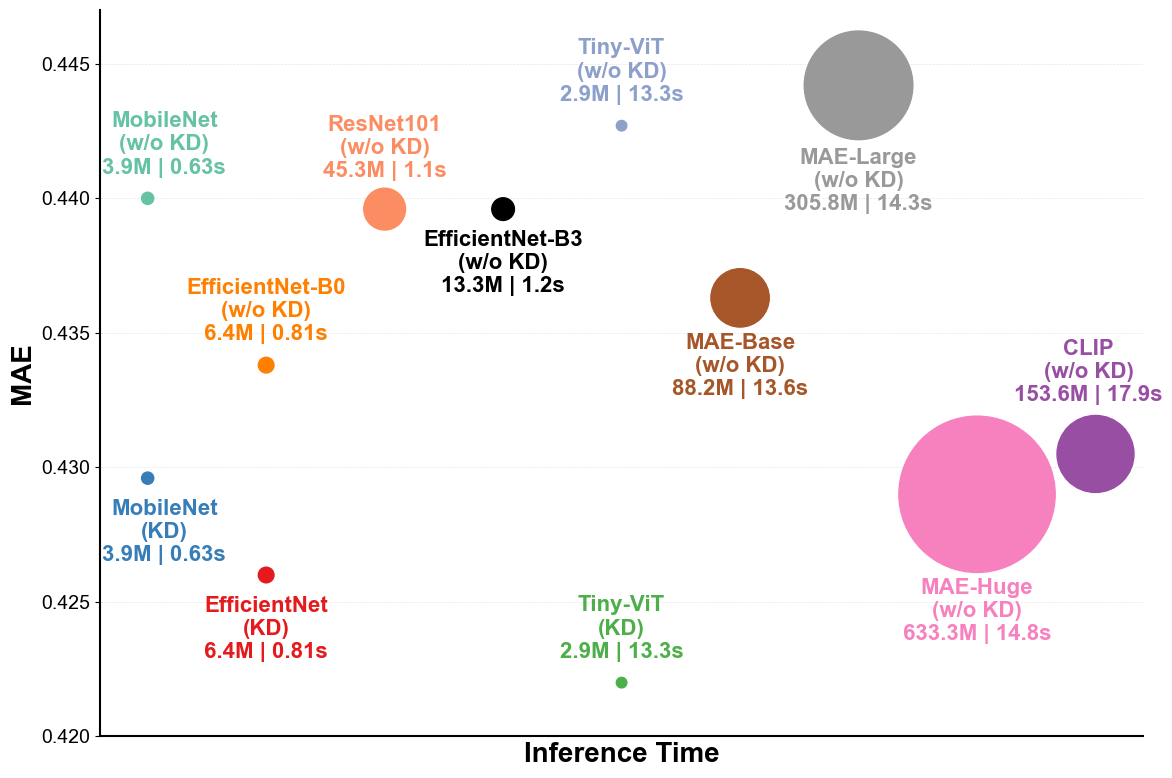}
    \caption{Model Efficiency Comparison, MAE vs Inference Time vs Parameters.}
    \label{fig:model_efficiency}

\end{figure}
\noindent\textbf{Computational Efficiency Analysis.} Figure~\ref{fig:model_efficiency} presents the average MAE and inference time across four forecasting horizons on the ETTh1 dataset, demonstrating the effectiveness of OccamVTS distillation. Student models with knowledge distillation (KD) consistently outperform their non-distilled counterparts (w/o KD) across all architectures. Remarkably, these lightweight student models, using less than 1-2\% of teacher model parameters, achieve substantially superior performance compared to massive teacher models. This 99\% parameter reduction is achieved while improving accuracy, as OccamVTS effectively extracts essential temporal patterns while filtering out redundant visual features. The results empirically validate our hypothesis that vision models contain substantial redundancy for time series tasks, and OccamVTS can achieve superior performance with minimal computational resources. For comprehensive results across all teacher-student pairings, see Appendix~\ref{appx:teacher-student-efficiency}.

\begin{table}[h!]
\centering
\renewcommand\arraystretch{1}
\captionsetup{font=small}

\begin{small}

\scalebox{1}{
\begin{tabular}{@{}c  cc  cc  cc@{}}
\toprule
& \multicolumn{2}{c}{Full} 
& \multicolumn{2}{c}{w/o Vision} 
& \multicolumn{2}{c}{w/o Temporal} \\
\cmidrule(lr){2-3} \cmidrule(lr){4-5} \cmidrule(lr){6-7}
Horizon & MSE & MAE & MSE & MAE & MSE & MAE \\
\midrule
96  & 0.150 & 0.201 & 0.172 & 0.225 & 0.269 & 0.314 \\
192 & 0.197 & 0.245 & 0.210 & 0.255 & 0.293 & 0.330 \\
336 & 0.248 & 0.285 & 0.252 & 0.287 & 0.322 & 0.348 \\
720 & 0.323 & 0.343 & 0.323 & 0.338 & 0.367 & 0.378 \\
\midrule
Avg & 0.229 & 0.268 & 0.239 & 0.276 & 0.313 & 0.343 \\
\bottomrule
\end{tabular}
}
\end{small}
\caption{Ablation Study on Multimodal Components 
on the Weather Dataset, Reporting MSE and MAE.}
\label{tab:kd_ablation}

\end{table}

\noindent\textbf{Ablation Study.}  To validate the effectiveness of each component, we first conduct ablation experiments on the undistilled teacher model. As shown in Table~\ref{tab:kd_ablation}, removing the vision component causes a 4.4\% MSE degradation on the Weather dataset, confirming that visual features meaningfully enhance temporal modeling. The more severe 36.7\% MSE increase when removing temporal components validates that both modalities are essential, with visual augmentation complementing core temporal representations.

\begin{figure}[!t]
  \centering
  \begin{subfigure}[b]{0.23\textwidth}
    \centering
    \includegraphics[width=\linewidth]{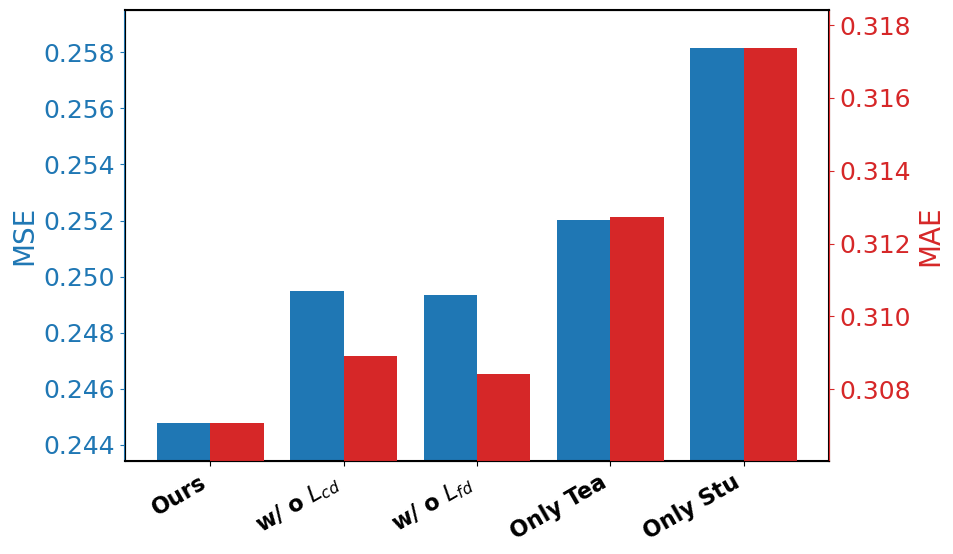}
    \caption{ETTm2.}
    \label{fig:column-ettm2}
  \end{subfigure}
  \hfill
  \begin{subfigure}[b]{0.23\textwidth}
    \centering
    \includegraphics[width=\linewidth]{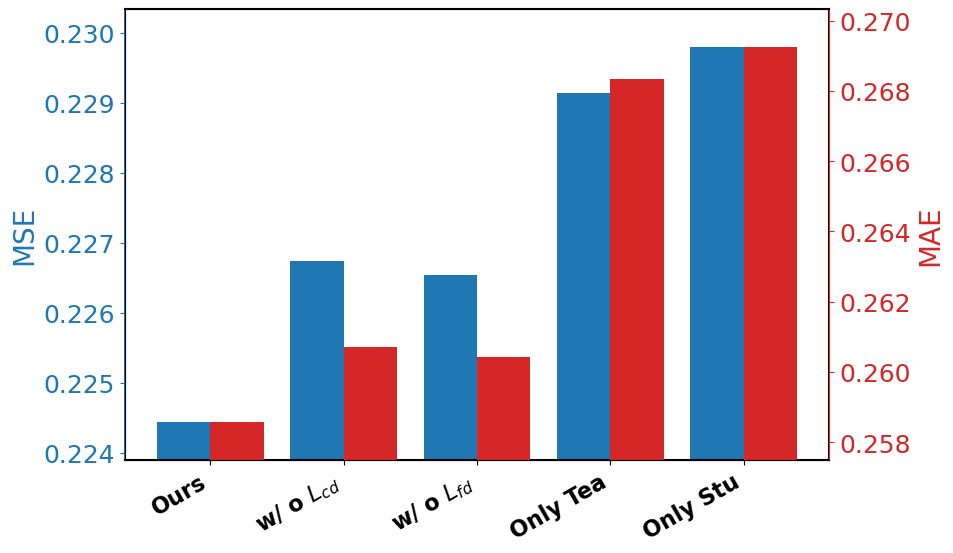}
    \caption{Weather.}
    \label{fig:column-weather}
  \end{subfigure}

  \begin{subfigure}[b]{0.23\textwidth}
    \centering
    \includegraphics[width=\linewidth]{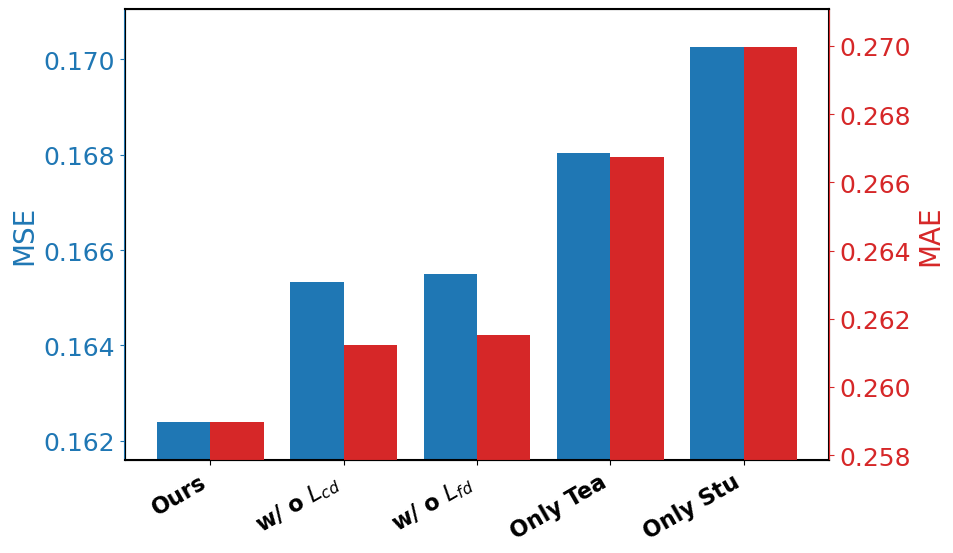}
    \caption{ECL.}
    \label{fig:column-ecl}
  \end{subfigure}
  \hfill
  \begin{subfigure}[b]{0.23\textwidth}
    \centering
    \includegraphics[width=\linewidth]{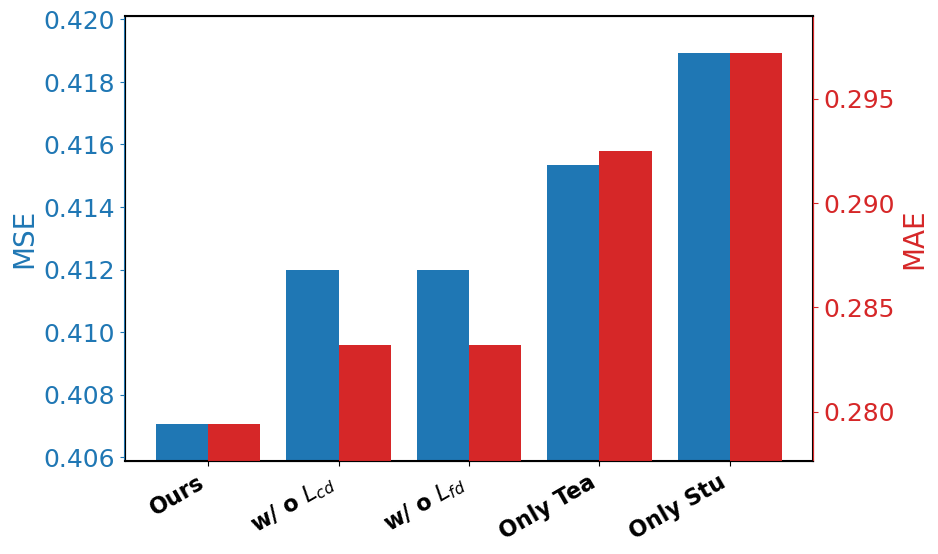}
    \caption{Traffic.}
    \label{fig:column-traffic}
  \end{subfigure}
  \caption{Ablation Experiment on Four Datasets.}
  \label{fig:ablg-dataset}

\end{figure}
Figure~\ref{fig:ablg-dataset} analyzes our knowledge distillation framework across four datasets, comparing our complete method against variants without correlation distillation(w/o $L_{cd}$), without feature distillation(w/o $L_{fd}$), teacher-only(Only Tea), and student-only(Only Stu) configurations. The results clearly show that our full framework consistently achieves the best performance. Both distillation components contribute meaningfully, while the teacher-only model suffers from interference of redundant parameters in pre-trained vision models, and the student-only model lacks essential cross-modal guidance. These ablation experiments confirm that our architecture benefits from the synergistic combination of all proposed components.
\begin{figure}[!b]
  \centering
  \begin{subfigure}[b]{0.23\textwidth}
    \centering
    \includegraphics[width=\linewidth]{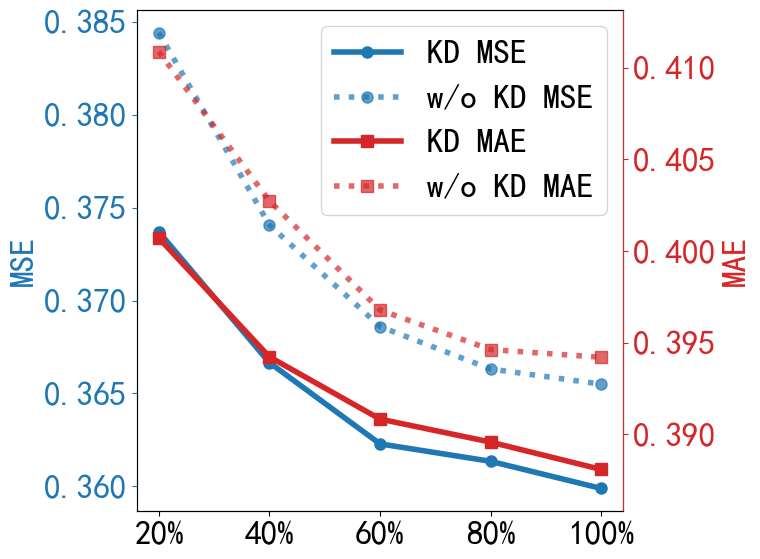}
    \caption{ETTh1.}
    \label{fig:sub-etth1}
  \end{subfigure}
  \hfill
  \begin{subfigure}[b]{0.23\textwidth}
    \centering
    \includegraphics[width=\linewidth]{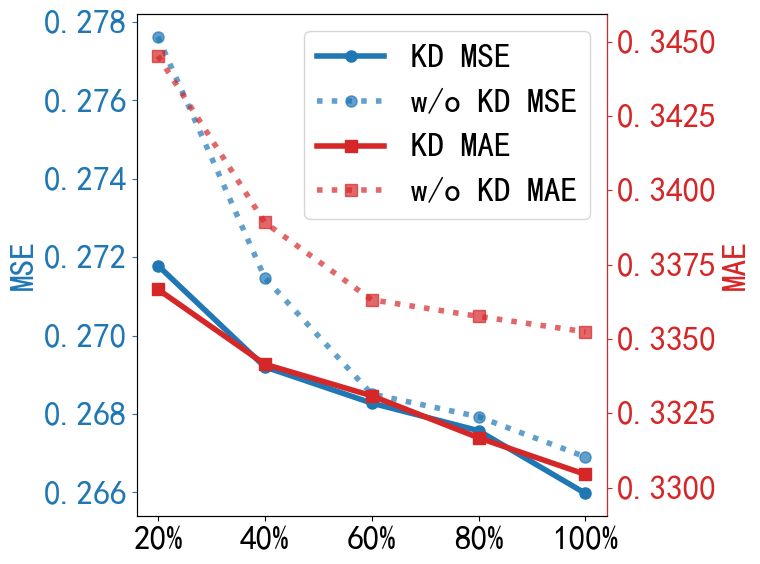}
    \caption{ETTh2.}
    \label{fig:sub-etth2}
  \end{subfigure}

  \begin{subfigure}[b]{0.23\textwidth}
    \centering
    \includegraphics[width=\linewidth]{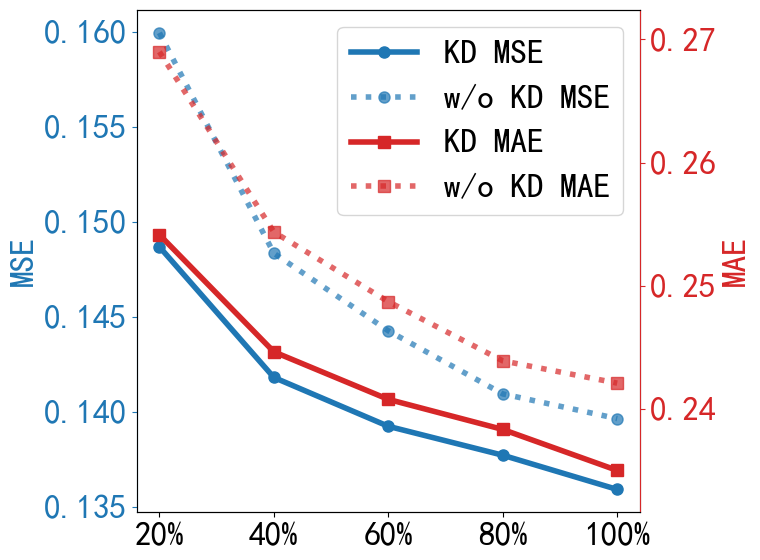}
    \caption{ECL.}
    \label{fig:sub-ecl}
  \end{subfigure}
  \hfill
  \begin{subfigure}[b]{0.23\textwidth}
    \centering
    \includegraphics[width=\linewidth]{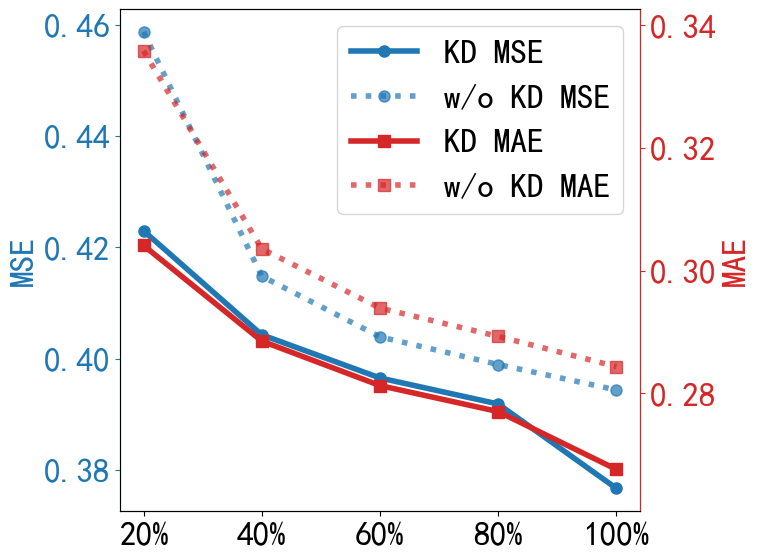}
    \caption{Traffic.}
    \label{fig:sub-traffic}
  \end{subfigure}

  \caption{Effect of Different Training Data on Four Datasets.}
  \label{fig:diff-training-data}

\end{figure}

\noindent\textbf{Scalability Study.} The scalability analysis on ETTh1, ETTh2, ECL, and Traffic datasets clearly demonstrates OccamVTS's significant advantages under varying data availability. As shown in Figure~\ref{fig:diff-training-data}, we evaluate performance with forecasting horizon $H=96$ as training data increases from 20\% to 100\%, where the KD version consistently outperforms the non-KD variant across all datasets.

The performance gap is particularly most pronounced under data scarcity (20\%-40\% training data), where knowledge distillation effectively transfers cross-modal knowledge despite limited samples. On ETTh1 and ETTh2, the MSE/MAE gap reaches 10-15\% in low-data scenarios. ECL and Traffic also show similar patterns, though with higher overall errors. As data availability increases, the performance gap narrows but persists, confirming that knowledge distillation provides substantial benefits even with sufficient data. These results validate OccamVTS's effectiveness, particularly in data-scarce scenarios common in practical applications.

\section{Conclusion}




We present OccamVTS, a novel knowledge distillation framework that eliminates visual model redundancy in time series forecasting. By distilling only the essential 1\% of predictive knowledge from vision models into lightweight student networks, OccamVTS achieves state-of-the-art performance while dramatically reducing computational overhead. Our approach challenges the paradigm of directly deploying massive pre-trained models, demonstrating that strategic knowledge distillation preserves cross-modal benefits while eliminating harmful redundancy. This proves especially valuable in data-scarce scenarios where traditional methods suffer from overfitting. By embodying Occam's razor principle, OccamVTS establishes a new direction for efficient cross-modal time series forecasting. 


Future work may explore distilling from foundation models, multi-expert ensembles, other modalities beyond vision, and domain-specific compression ratios. For comprehensive discussion of these directions, see Appendix \ref{appx:future_work}.



\section*{Acknowledgments}
This work was supported by the Guangdong Basic and Applied Basic Research Foundation (Grant Nos.\ 2025A1515011994, 2023B1515120057). It was also supported by the National Key R\&D Program of China (Grant No.\ 2023YFF0725001); the National Natural Science Foundation of China (Grant Nos.\ 62402414, 92370204); the Guangzhou Municipal Science and Technology Project (Grant No.\ 2023A03J0011); the Guangzhou Industrial Information and Intelligent Key Laboratory Project (Grant No.\ 2024A03J0628); a grant from the State Key Laboratory of Resources and Environmental Information System; the Guangdong Provincial Key Laboratory of Integrated Communication, Sensing and Computation for Ubiquitous Internet of Things (Grant No.\ 2023B1212010007); the Key‑Area Special Project of Guangdong Provincial Ordinary Universities (Grant No.\ 2024ZDZX1007); and the Education Bureau of Guangzhou, which we gratefully acknowledge.


\bibliography{main}

\makeatletter
\@ifundefined{isChecklistMainFile}{
  \newif\ifreproStandalone
  \reproStandalonetrue
}{
  \newif\ifreproStandalone
  \reproStandalonefalse
}
\makeatother

\ifreproStandalone
\documentclass[letterpaper]{article}
\usepackage[submission]{aaai2026}
\setlength{\pdfpagewidth}{8.5in}
\setlength{\pdfpageheight}{11in}
\usepackage{times}
\usepackage{helvet}
\usepackage{courier}
\usepackage{xcolor}
\frenchspacing

\begin{document}
\fi
\setlength{\leftmargini}{20pt}
\makeatletter\def\@listi{\leftmargin\leftmargini \topsep .5em \parsep .5em \itemsep .5em}
\def\@listii{\leftmargin\leftmarginii \labelwidth\leftmarginii \advance\labelwidth-\labelsep \topsep .4em \parsep .4em \itemsep .4em}
\def\@listiii{\leftmargin\leftmarginiii \labelwidth\leftmarginiii \advance\labelwidth-\labelsep \topsep .4em \parsep .4em \itemsep .4em}\makeatother

\renewcommand\thesubsection{\arabic{subsection}}
\renewcommand\labelenumi{\thesubsection.\arabic{enumi}}

\newcounter{checksubsection}
\newcounter{checkitem}[checksubsection]

\newcommand{\checksubsection}[1]{%
  \refstepcounter{checksubsection}%
  \paragraph{\arabic{checksubsection}. #1}%
  \setcounter{checkitem}{0}%
}

\newcommand{\checkitem}{%
  \refstepcounter{checkitem}%
  \item[\arabic{checksubsection}.\arabic{checkitem}.]%
}
\newcommand{\question}[2]{\normalcolor\checkitem #1 #2 \color{blue}}
\newcommand{\ifyespoints}[1]{\makebox[0pt][l]{\hspace{-15pt}\normalcolor #1}}

\ifreproStandalone
\end{document}
\fi

\newpage
\onecolumn
\appendix

\section{Dataset Details}
\label{appx:dataset_details}
We evaluate our proposed method on thirteen well-established benchmark datasets that represent a comprehensive suite of real-world time series forecasting challenges, as detailed in Table \ref{tab:dataset}. These datasets encompass diverse application domains, including electricity transformer temperature monitoring (ETTm1, ETTm2, ETTh1, ETTh2), power consumption analysis (Electricity), transportation systems (Traffic), and meteorological forecasting (Weather). Each dataset comprises multiple multivariate time series with varying dimensions and sequence lengths, systematically partitioned into training, validation, and testing sets following standard evaluation protocols. The datasets exhibit distinct temporal characteristics, with sampling frequencies ranging from fine-grained 15-minute intervals to coarse yearly observations, and demonstrate various periodic patterns that reflect real-world phenomena. For short-term forecasting evaluation, we employ the M4 competition benchmark, which encompasses six temporal granularities (yearly, quarterly, monthly, weekly, daily, and hourly) across multiple domains including demographics, finance, industry, macroeconomics, microeconomics, and other specialized fields. This comprehensive dataset collection provides a rigorous testbed for evaluating forecasting model performance across different temporal scales, dimensionalities, and domain-specific characteristics.

\begin{table}[htbp]
  \centering
  \begin{small}
    \scalebox{0.9}{
      \begin{tabular}{c|l|c|c|c|c|c|c|c}
        \toprule
        Tasks& Dataset & Dim. & Series Length & Dataset Size & Frequency & Domain & Forecastability* & Periodicity\\
        \toprule
        & ETTm1 & 7 & {\{96, 192, 336, 720\}} & (34465, 11521, 11521) & 15 min & Temperature & 0.46 & 96\\
        \cmidrule{2 - 9}
        Long-term & ETTm2 & 7 & {\{96, 192, 336, 720\}} & (34465, 11521, 11521) & 15 min & Temperature & 0.55 & 96\\
        \cmidrule{2 - 9}
        Forecasting & ETTh1 & 7 & {\{96, 192, 336, 720\}} & (8545, 2881, 2881) & 1 hour & Temperature & 0.38 & 24\\
        \cmidrule{2 - 9}
        & ETTh2 & 7 & {\{96, 192, 336, 720\}} & (8545, 2881, 2881) & 1 hour & Temperature & 0.45 & 24\\ 
        \cmidrule{2 - 9}
        & Electricity & 321 & {\{96, 192, 336, 720\}} & (18317, 2633, 5261) & 1 hour & Electricity & 0.77 & 24\\ 
        \cmidrule{2 - 9}
        & Traffic & 862 & {\{96, 192, 336, 720\}} & (12185, 1757, 3509) & 1 hour & Transportation & 0.68 & 24\\ 
        \cmidrule{2 - 9}
        & Weather & 21 & {\{96, 192, 336, 720\}} & (36792, 5271, 10540) & 10 min & Weather & 0.75 & 144\\
        \midrule
        & M4-Yearly & 1 & 6 & (23000, 0, 23000) & Yearly & Demographic & 0.43 & 1\\
        \cmidrule{2 - 9}
        & M4-Quarterly & 1 & 8 & (24000, 0, 24000) & Quarterly & Finance & 0.47 & 4\\
        \cmidrule{2 - 9}
        Short-term & M4-Monthly & 1 & 18 & (48000, 0, 48000) & Monthly & Industry & 0.44 & 12\\
        \cmidrule{2 - 9}
        Forecasting & M4-Weekly & 1 & 13 & (359, 0, 359) & Weekly & Macro & 0.43 & 52\\
        \cmidrule{2 - 9}
        & M4-Daily & 1 & 14 & (4227, 0, 4227) & Daily & Micro & 0.44 & 7\\
        \cmidrule{2 - 9}
        & M4-Hourly & 1 & 48 & (414, 0, 414) & Hourly & Other & 0.46 & 24\\
        \bottomrule
        \multicolumn{9}{l}{\scriptsize* The forecastability is calculated by one minus the entropy of Fourier decomposition of time series \cite{goerg2013forecastable}. A larger value indicates better predictability.} \\
      \end{tabular}
    }
  \end{small}
    \caption{Summary of benchmark datasets. Each dataset includes multiple time series (Dim.) with varying sequence lengths, split into training, validation, and testing sets. Data are collected at different frequencies across various domains.}
  \label{tab:dataset}
\end{table}

\noindent The datasets employed in our comprehensive evaluation are described below:

\begin{itemize}
    \item \textbf{ETT}: Four datasets (ETTh1, ETTh2, ETTm1, ETTm2) containing two years of electricity transformer temperature data from two counties in China. ETTh1/ETTh2 provide hourly measurements, while ETTm1/ETTm2 offer 15-minute resolution. Each dataset contains seven variables: six power load features and one target oil temperature variable.
    \item \textbf{Traffic}: Hourly road occupancy rates from 862 sensors across the San Francisco Bay Area freeway system, collected by the California Department of Transportation. The dataset captures traffic flow patterns and congestion dynamics across multiple road segments.
    \item \textbf{Weather}: Meteorological measurements from 21 weather stations in Germany, recorded every 10 minutes over one year. The dataset includes 21 atmospheric indicators such as air temperature, humidity, atmospheric pressure, and wind conditions.
    \item \textbf{Electricity}: Hourly electricity consumption records from 321 customers, including both residential and commercial entities. The dataset exhibits complex daily, weekly, and seasonal consumption patterns reflecting diverse user behaviors.
    \item \textbf{M4}: A comprehensive collection from the Makridakis Forecasting Competition, comprising six subsets with different temporal frequencies: Yearly (23,000 series), Quarterly (24,000 series), Monthly (48,000 series), Weekly (359 series), Daily (4,227 series), and Hourly (414 series). Each subset spans different domains including finance, industry, demographics, and economics.

\end{itemize}

\noindent The \textit{Periodicity} column in Table~\ref{tab:dataset} delineates the periodicity hyperparameter $P$ utilized in the periodicity encoding process. This parameter is empirically derived from the inherent temporal characteristics of each dataset and encapsulates the dominant cyclic patterns, encompassing daily, weekly, and seasonal periodicities. Specifically, for the ETTm1 and ETTm2 datasets, which employ a 15-minute sampling interval, the periodicity parameter $P = 96$ corresponds to a complete diurnal cycle (24 hours $\times$ 4 samples per hour). Analogously, the ETTh1 and ETTh2 datasets, characterized by hourly sampling, adopt $P = 24$ to represent the daily periodicity. The Weather dataset, with its 10-minute sampling resolution, utilizes $P = 144$, thereby capturing the full 24-hour cycle (24 hours $\times$ 6 samples per hour). Regarding the M4 benchmark datasets, the periodicity values are systematically determined according to their respective temporal granularities: $P = 1$ for yearly data (reflecting annual cycles), $P = 4$ for quarterly data, $P = 3$ for monthly data, $P = 4$ for weekly data, and $P = 24$ for hourly data (capturing diurnal patterns). These carefully calibrated values are incorporated through the following trigonometric encoding formulation:

\begin{equation}
    \text{encoding}(t) = \left[ \sin\left(\frac{2\pi t}{P}\right), \cos\left(\frac{2\pi t}{P}\right) \right],
\end{equation}

where $t$ represents the discrete time step and $P$ denotes the dataset-specific periodicity hyperparameter. The resulting sinusoidal encodings are subsequently concatenated with the original time series representations, thereby augmenting the model's capacity to capture both short-term temporal dependencies and long-term periodic patterns inherent in the data.

\section{Evaluation Metrics}
\label{appx:evaluation_metrics}
For evaluation purposes, we employ mean squared error (MSE) and mean absolute error (MAE) as metrics for long-term forecasting tasks. In contrast, short-term forecasting performance on the M4 benchmark is assessed using symmetric mean absolute percentage error (SMAPE), mean absolute scaled error (MASE), and overall weighted average (OWA), adhering to the evaluation framework established by N-BEATS \cite{oreshkin2019n}. Note that OWA represents a specialized metric designed for the M4 competition. The computation of these metrics follows the formulations below:

\begin{align*} \label{equ:metrics}
    \text{MSE} &= \frac{1}{H}\sum_{h=1}^T (\mathbf{Y}_{h} - \Hat{\mathbf{Y}}_{h})^2,
    &
    \text{MAE} &= \frac{1}{H}\sum_{h=1}^H|\mathbf{Y}_{h} - \Hat{\mathbf{Y}}_{h}|,\\
    \text{SMAPE} &= \frac{200}{H} \sum_{h=1}^H \frac{|\mathbf{Y}_{h} - \Hat{\mathbf{Y}}_{h}|}{|\mathbf{Y}_{h}| + |\Hat{\mathbf{Y}}_{h}|},
    &
    \text{MAPE} &= \frac{100}{H} \sum_{h=1}^H \frac{|\mathbf{Y}_{h} - \Hat{\mathbf{Y}}_{h}|}{|\mathbf{Y}_{h}|}, \\
    \text{MASE} &= \frac{1}{H} \sum_{h=1}^H \frac{|\mathbf{Y}_{h} - \Hat{\mathbf{Y}}_{h}|}{\frac{1}{H-s}\sum_{j=s+1}^{H}|\mathbf{Y}_j - \mathbf{Y}_{j-s}|},
    &
    \text{OWA} &= \frac{1}{2} \left[ \frac{\text{SMAPE}}{\text{SMAPE}_{\textrm{Naïve2}}}  + \frac{\text{MASE}}{\text{MASE}_{\textrm{Naïve2}}}  \right],
\end{align*}

where $s$ is the periodicity of the time series, $H$ is the prediction horizon, and $\mathbf{Y}_{h}$ and $\Hat{\mathbf{Y}}_{h}$ are the ground truth and prediction at time step $h$, respectively.

\section{Optimization Settings}
\label{appx:optimization_settings}

\subsection{Model Architecture Parameters}

The proposed method encompasses several pivotal components, each meticulously configured with specific parameter settings as detailed in Table~\ref{tab:model_arch_params}. Input image representations are established at a size of $56 \times 56$, striking an optimal balance between computational efficiency and comprehensive visual information preservation. The model backbone employs a hidden dimension of $d\_model = 128$, while the encoder-decoder architecture comprises $e\_layers = 2$ encoder layers and $d\_layers = 1$ decoder layer. A dropout rate of $0.1$ is judiciously applied to mitigate overfitting phenomena during the training regimen. To facilitate efficient data loading, the model leverages $num\_workers = 32$ to parallelize data preprocessing operations.
The fusion layer is elegantly designed with a dimension of $d\_fusion = 256$, orchestrating the seamless integration of multimodal features. The memory mechanism incorporates a memory bank with a capacity of $memory\_size = 100$, utilizing a $top\_k = 5$ selection strategy to retrieve the most pertinent historical information. These architectural choices are consistently applied across all experimental configurations to maintain fairness in model comparison.

\begin{table}[h!]
  \centering
  \begin{small}
    \scalebox{1}{
      \begin{tabular}{c|c|p{6.5cm}}
        \toprule
        \multicolumn{1}{c|}{Parameter} & \multicolumn{1}{c|}{Default Value} & \multicolumn{1}{c}{Description} \\
        \midrule
        image\_size & 56 & Input image size \\
        \cmidrule{1-3}
        d\_model & 128 & Dimension of hidden model \\
        \cmidrule{1-3}
        d\_fusion & 256 & Dimension of fusion layer \\
        \cmidrule{1-3}
        num\_workers & 32 & Number of data loader workers \\
        \cmidrule{1-3}
        e\_layers & 2 & Number of encoder layers \\
        \cmidrule{1-3}
        d\_layers & 1 & Number of decoder layers \\
        \cmidrule{1-3}
        dropout & 0.1 & Dropout rate \\
        \cmidrule{1-3}
        memory\_size & 100 & Maximum capacity of memory bank \\
        \cmidrule{1-3}
        top\_k & 5 & Top–K selection \\
        \cmidrule{1-3}
        teacher\_hidden\_size & \makecell{512 (Clip); 768 (Mae-Base) \\ 1024 (Mae-Large); 1280 (Mae-Huge)\\ 1536 (Efficientnet-B3); 2048 (Resnet101)}  & Hidden size of teacher model \\
        \cmidrule{1-3}
        student\_hidden\_size & 128 (Tiny-Vit; Efficientnet-B0; MobileNet-V3) & Hidden size of student model \\
        \bottomrule
      \end{tabular}
    }
  \end{small}
    \caption{Default Model Architecture Parameters}
  \label{tab:model_arch_params}
\end{table}

\subsection{Training Parameters} 
To ensure fair comparison across all methods, we adopt identical experimental settings: 70\%/10\%/20\% train/validation/test splits, consistent hardware infrastructure, and standardized evaluation metrics (MSE and MAE). We embrace a comprehensive training methodology encompassing both general and task-specific parameter configurations, as summarized in Table~\ref{tab:training_params}. The model undergoes training with a batch size of $32$ and an initial learning rate of $0.001$, employing the \textit{AdamW} optimizer \cite{loshchilov2017decoupled}. Early stopping mechanisms are implemented with a patience threshold of $5$ epochs to circumvent overfitting. The training process adopts Mean Squared Error (MSE) as the principal loss function and executes for a maximum of $10$ epochs. For temporal sequence processing, we utilize an input sequence length of $512$, with prediction horizons configured to $96$, $192$, $336$, or $720$ according to task specifications. The output dimension ($c\_out$) varies across datasets: $7$ for ETTh1/h2/m1/m2, $21$ for Weather, $321$ for Electricity, and $862$ for Traffic. Periodicity parameters are meticulously tailored to dataset characteristics: $24$ for ETTh1/h2, Electricity, and Traffic; $96$ for ETTm1/m2; and $144$ for Weather, ensuring harmonious alignment with dataset-specific temporal patterns. A normalization coefficient of $0.4$ is applied to stabilize training dynamics. The patch embedding module employs a patch length of $16$, stride of $8$, and padding of $8$ to process input sequences. The temporal memory mechanism harnesses $8$ learnable queries and $4$ attention heads to capture sophisticated high-level dependencies.
\begin{table}[h!]
  \centering
  \begin{small}
    \scalebox{1}{
      \begin{tabular}{c|c|p{6.5cm}}
        \toprule
        \multicolumn{1}{c|}{Parameter} & \multicolumn{1}{c|}{Default Value} & \multicolumn{1}{c}{Description} \\
        \midrule
        batch\_size & 32 & Training batch size \\
        \cmidrule{1-3}
        learning\_rate & 0.001 & Initial learning rate \\
        \cmidrule{1-3}
        train\_epochs & 10 & Number of training epochs \\
        \cmidrule{1-3}
        patience & 5 & Early stopping patience \\
        \cmidrule{1-3}
        loss & MSE & Mean square error loss \\
        \cmidrule{1-3}
        seq\_len & 512 & Input sequence length \\
        \cmidrule{1-3}
        c\_out & \makecell{7 (ETTh1/h2/m1/m2) \\ 21 (Weather) \\ 321 (Electricity) \\ 862 (Traffic)} & Output dimension (dataset-specific) \\
        \cmidrule{1-3}
        pred\_len & \makecell{96/192/336/720} & Prediction length \\
        \cmidrule{1-3}
        periodicity & \makecell{24 (ETTh1/h2/Electricity/Traffic) \\ 96 (ETTm1/m2) \\ 144 (Weather)} & Dataset periodicity (dataset-specific) \\
        \cmidrule{1-3}
        norm\_const & 0.4 & Normalization coefficient \\
        \cmidrule{1-3}
        patch\_len & 16 & Patch length \\
        \cmidrule{1-3}
        padding & 8 & Padding length \\
        \cmidrule{1-3}
        stride & 8 & Stride length \\
        \cmidrule{1-3}
        num\_queries & 8 & Number of learnable queries for temporal memory \\
        \cmidrule{1-3}
        n\_heads & 4 & Number of attention heads \\
        \bottomrule
      \end{tabular}
    }
  \end{small}
    \caption{Default Training Parameters}
  \label{tab:training_params}
\end{table}

\subsection{Distillation Parameters} 

The knowledge distillation process adopts an adaptive weight balancing strategy, wherein all loss weights are architected as learnable parameters to achieve dynamic optimization, as presented in Table~\ref{tab:distillation_params}. The initial weight for feature distillation loss is configured as $init\_feature\_w = 0.01$, while the prediction task loss maintains an initial weight of $init\_fcst\_w = 1$, the reconstruction loss begins with $init\_recon\_w = 0.5$, and the attention distillation loss commences with $init\_att\_w = 0.01$. The distillation temperature is initialized to $init\_temperature = 4$, serving to soften the teacher model's output distributions. Distillation parameters undergo optimization at $0.1$ times the primary model's learning rate, specifically $distill\_lr\_ratio = 0.1$.
The multi-scale feature alignment mechanism is configured with $num\_alignment\_scales = 3$ hierarchical scales, incorporating a dropout rate of $feature\_alignment\_dropout = 0.1$ within feature alignment layers to enhance generalization capabilities. The loss balancer employs a momentum parameter of $loss\_momentum = 0.9$ to smoothly modulate the update process of loss weights. To prevent weight degradation, a weight regularization coefficient of $weight\_regularization = 0.001$ is applied. This comprehensive distillation strategy ensures the efficacious transfer of teacher model knowledge while preserving the compactness and inference efficiency of the student model. 

\begin{table}[h!]
  \centering
  \begin{small}
    \scalebox{1}{
      \begin{tabular}{c|c|p{6.5cm}}
        \toprule
        \multicolumn{1}{c|}{Parameter} & \multicolumn{1}{c|}{Default Value} & \multicolumn{1}{c}{Description} \\
        \midrule
        init\_feature\_w & 0.01 & Initial feature distillation loss weight (learnable) \\
        \cmidrule{1-3}
        init\_fcst\_w & 1 & Initial prediction task loss weight (learnable) \\
        \cmidrule{1-3}
        init\_recon\_w & 0.5 & Initial reconstruction loss weight (learnable) \\
        \cmidrule{1-3}
        init\_att\_w & 0.01 & Initial attention distillation loss weight (learnable) \\
        \cmidrule{1-3}
        init\_temperature & 4 & Initial distillation temperature (learnable) \\
        \cmidrule{1-3}
        distill\_lr\_ratio & 0.1 & Learning rate ratio for distillation parameters \\
        \cmidrule{1-3}
        num\_alignment\_scales & 3 & Number of scales for multi-scale feature alignment \\
        \cmidrule{1-3}
        feature\_alignment\_dropout & 0.1 & Dropout rate for feature alignment layers \\
        \cmidrule{1-3}
        loss\_momentum & 0.9 & Momentum parameter for loss balancer \\
        \cmidrule{1-3}
        weight\_regularization & 0.001 & Weight regularization coefficient \\
        \bottomrule
      \end{tabular}
    }
  \end{small}
    \caption{Default Distillation Parameters}
  \label{tab:distillation_params}
\end{table}

\subsection{Teacher and Student Model}

The knowledge distillation architecture supports six heterogeneous teacher models—CLIP (512-dimensional), MAE-Base (768-dimensional), MAE-Large (1024-dimensional), MAE-Huge (1280-dimensional)~\cite{he2022masked}, EfficientNet-B3 (1536-dimensional)~\cite{tan2019efficientnet}, and ResNet101 (2048-dimensional)~\cite{he2016deep}—while all student models share a 128-dimensional hidden size and adopt lightweight backbones such as Tiny-ViT, EfficientNet-B0, and MobileNet-V3~\cite{howard2017mobilenets}. All visual baselines (teachers and students alike) are reproduced from their official publicly released pre-trained weights to ensure fair comparisons.

\noindent\textbf{Why these six teacher models?}

We deliberately combine teachers that differ along several orthogonal axes to thoroughly assess how visual knowledge transfers to time series forecasting:
\begin{itemize}
    \item \textbf{Vision Transformers (MAE variants).} MAE-Base (768-dimensional), MAE-Large (1024-dimensional), and MAE-Huge (1280-dimensional) are self-supervised ViTs trained with masked autoencoding. They emphasize texture/frequency cues over high-level semantics, aligning well with our texture-like sequence renderings.
    \item \textbf{Multi-modal foundation model (CLIP).} CLIP (512-dimensional) introduces vision–language aligned features, enabling us to test whether multi-modal pre-training provides advantages beyond purely visual objectives for temporal pattern discovery.
    \item \textbf{Convolutional architectures.} EfficientNet-B3 (1536-dimensional) and ResNet101 (2048-dimensional) exemplify two CNN philosophies—compound-scaled efficiency vs.\ very deep residual learning—letting us contrast hierarchical convolutional features with transformer token representations. These CNNs' lower layers capture edge and gradient patterns that naturally align with temporal transitions, while their multi-scale receptive fields match time series' varying periodicities.
\end{itemize}
This ensemble spans feature dimensionalities from 512 to 2048 and covers contrastive, masked-reconstruction, and supervised pre-training regimes, as well as CNN and Transformer inductive biases. Such breadth allows a systematic study of how capacity, architecture, and objective choice influence distillation efficacy.

\noindent\textbf{Why these three student models?}

All students use a fixed 128-dimensional hidden size to ensure that gains stem from \emph{selective distillation}, not from increased capacity. We nonetheless choose architecturally diverse, deployment-friendly backbones:
\begin{itemize}
    \item \textbf{Tiny-ViT.} This compact Vision Transformer retains self-attention mechanisms while operating under strict parameter constraints. It serves as an ideal probe for examining whether transformer teachers most effectively transfer knowledge to transformer students when computational budgets are tight.
    
    \item \textbf{EfficientNet-B0.} As the smallest member of the EfficientNet family, this model was produced via Neural Architecture Search (NAS) to achieve optimal accuracy-efficiency trade-offs. It represents the state-of-the-art in modern efficient CNN design principles.
    
    \item \textbf{MobileNet-V3.} This mobile-first CNN incorporates hardware-aware components such as squeeze-and-excitation blocks and h-swish activations. It demonstrates that our distillation framework maintains effectiveness even for deployment scenarios with stringent on-device constraints.
\end{itemize}
Across these students we achieve a 52–99\% parameter reduction relative to their teachers while covering distinct architectural priors, providing a rigorous test bed for our core hypothesis that massive parameter redundancy exists in vision models for time series forecasting.

\section{Long-term Forecasting}
\label{appx:long-term-forecasting}
Table~\ref{tab:long-term-forecasting} provides comprehensive performance evaluation across different prediction horizons (96, 192, 336, 720) for long-term forecasting tasks. Our OccamVTS framework demonstrates consistent superiority, achieving the best MSE performance across all seven datasets while obtaining either the best or second-best MAE results. Particularly noteworthy is the performance on the ETT dataset series, where we achieve MSE values of 0.403 on ETTh1 (vs. TimeVLM's 0.405) and 0.336 on ETTh2 (representing a substantial 12.0\% improvement over PatchTST's 0.382). The effectiveness becomes more pronounced in high-dimensional scenarios, with excellent results on both the Electricity dataset (MSE: 0.162 vs. TimeMixer++'s 0.165) and Traffic dataset (MSE: 0.407 vs. iTransformer's 0.428), demonstrating our architecture's capability to handle complex temporal patterns effectively.

\begin{table}[h!]
\captionsetup{font=small}
\begin{center}
\begin{small}
\scalebox{0.65}{
\setlength\tabcolsep{4pt}
\begin{tabular}{c|c|cc|cc|cc|cc|cc|cc|cc|cc|cc|cc|cc|cc|cc}
\toprule

\multicolumn{2}{c|}{Methods}&\multicolumn{2}{c|}{Ours}& \multicolumn{2}{c|}{Only Teacher} & \multicolumn{2}{c|}{Only Student}&\multicolumn{2}{c|}{TimeVLM}&\multicolumn{2}{c|}{TimeMixer++}&\multicolumn{2}{c|}{TimeMixer}&\multicolumn{2}{c|}{LDM4TS}&\multicolumn{2}{c|}{TimesNet}&\multicolumn{2}{c|}{iTransformer}&\multicolumn{2}{c|}{DLinear}&\multicolumn{2}{c|}{PatchTST}&\multicolumn{2}{c|}{FEDformer}&\multicolumn{2}{c}{Autoformer} \\

\midrule

\multicolumn{2}{c|}{Metric} & MSE  & MAE & MSE & MAE & MSE & MAE & MSE & MAE & MSE  & MAE & MSE  & MAE & MSE  & MAE & MSE  & MAE & MSE  & MAE & MSE  & MAE & MSE  & MAE & MSE  & MAE & MSE  & MAE\\
\midrule

\multirow{5}{*}{\rotatebox{90}{$ETTh1$}}
& 96 & \boldres{0.360} & \secondres{0.388} & 0.365 & 0.394 & 0.368 & 0.397 & 0.361 & \boldres{0.386} & \secondres{0.361} & 0.403 & 0.375 & 0.400 & 0.388 & 0.411 & 0.384 & 0.402 & 0.386 & 0.405 & 0.375 & 0.399 & 0.378 & 0.398 & 0.376 & 0.419 & 0.449 & 0.459 \\
& 192 & \boldres{0.396} & \boldres{0.410} & 0.407 & 0.424 & 0.399 & 0.415 & \secondres{0.397} & \secondres{0.415} & 0.416 & 0.441 & 0.429 & 0.421 & 0.412 & 0.430 & 0.436 & 0.429 & 0.441 & 0.436 & 0.405 & 0.416 & 0.425 & 0.432 & 0.420 & 0.448 & 0.500 & 0.482 \\
& 336 & \boldres{0.416} & \secondres{0.424} & 0.421 & 0.431 & 0.453 & 0.455 & \secondres{0.420} & \boldres{0.421} & 0.430 & 0.434 & 0.484 & 0.458 & 0.471 & 0.473 & 0.491 & 0.469 & 0.487 & 0.458 & 0.439 & 0.443 & 0.470 & 0.458 & 0.459 & 0.465 & 0.521 & 0.496 \\
& 720 & \boldres{0.440} & 0.462 & 0.470 & 0.483 & 0.515 & 0.510 & \secondres{0.441} & \secondres{0.458} & 0.467 & \boldres{0.451} & 0.498 & 0.482 & 0.501 & 0.502 & 0.521 & 0.500 & 0.503 & 0.491 & 0.472 & 0.490 & 0.525 & 0.507 & 0.506 & 0.507 & 0.514 & 0.512 \\
& Avg & \boldres{0.403} & \secondres{0.421} & 0.416 & 0.433 & 0.434 & 0.444 & \secondres{0.405} & \boldres{0.420} & 0.419 & 0.432 & 0.447 & 0.440 & 0.443 & 0.454 & 0.458 & 0.450 & 0.454 & 0.447 & 0.422 & 0.437 & 0.450 & 0.449 & 0.440 & 0.460 & 0.496 & 0.487 \\
\midrule

\multirow{5}{*}{\rotatebox{90}{$ETTh2$}}
& 96 & \boldres{0.266} & \secondres{0.330} & \secondres{0.267} & 0.335 & 0.268 & 0.336 & 0.267 & 0.335 & 0.276 & \boldres{0.328} & 0.289 & 0.341 & 0.316 & 0.378 & 0.340 & 0.374 & 0.297 & 0.349 & 0.289 & 0.353 & 0.291 & 0.346 & 0.358 & 0.397 & 0.346 & 0.388 \\
& 192 & 0.330 & \boldres{0.372} & \secondres{0.329} & 0.377 & 0.335 & 0.384 & \boldres{0.326} & \secondres{0.373} & 0.342 & 0.379 & 0.372 & 0.392 & 0.356 & 0.404 & 0.402 & 0.414 & 0.380 & 0.400 & 0.383 & 0.418 & 0.378 & 0.404 & 0.429 & 0.439 & 0.456 & 0.452 \\
& 336 & 0.357 & \secondres{0.400} & \secondres{0.355} & 0.400 & 0.360 & 0.409 & 0.357 & 0.406 & \boldres{0.346} & \boldres{0.398} & 0.386 & 0.414 & 0.438 & 0.461 & 0.452 & 0.452 & 0.428 & 0.432 & 0.448 & 0.465 & 0.425 & 0.440 & 0.496 & 0.487 & 0.482 & 0.486 \\
& 720 & \boldres{0.389} & \secondres{0.431} & 0.400 & 0.440 & 0.404 & 0.446 & 0.412 & 0.449 & \secondres{0.392} & \boldres{0.415} & 0.412 & 0.434 & 0.436 & 0.465 & 0.462 & 0.468 & 0.427 & 0.445 & 0.605 & 0.551 & 0.436 & 0.454 & 0.463 & 0.474 & 0.515 & 0.511 \\
& Avg & \boldres{0.336} & \secondres{0.383} & \secondres{0.338} & 0.388 & 0.342 & 0.394 & 0.341 & 0.391 & 0.339 & \boldres{0.380} & 0.365 & 0.395 & 0.387 & 0.427 & 0.414 & 0.427 & 0.383 & 0.407 & 0.431 & 0.446 & 0.382 & 0.411 & 0.437 & 0.449 & 0.450 & 0.459 \\
\midrule

\multirow{5}{*}{\rotatebox{90}{$ETTm1$}}
& 96 & \boldres{0.292} & \secondres{0.343} & 0.303 & 0.347 & 0.303 & 0.347 & \secondres{0.294} & 0.346 & 0.310 & \boldres{0.334} & 0.320 & 0.357 & 0.331 & 0.373 & 0.338 & 0.375 & 0.334 & 0.368 & 0.299 & 0.343 & 0.324 & 0.364 & 0.379 & 0.419 & 0.505 & 0.475 \\
& 192 & \boldres{0.324} & \boldres{0.362} & 0.332 & 0.363 & 0.332 & 0.363 & \secondres{0.330} & 0.366 & 0.348 & \secondres{0.362} & 0.361 & 0.381 & 0.346 & 0.382 & 0.374 & 0.387 & 0.377 & 0.391 & 0.335 & 0.365 & 0.372 & 0.392 & 0.426 & 0.441 & 0.553 & 0.496 \\
& 336 & \boldres{0.360} & \boldres{0.378} & 0.363 & \secondres{0.382} & 0.363 & 0.381 & \secondres{0.361} & 0.383 & 0.376 & 0.391 & 0.390 & 0.404 & 0.371 & 0.394 & 0.410 & 0.411 & 0.426 & 0.420 & 0.369 & 0.386 & 0.399 & 0.408 & 0.445 & 0.459 & 0.621 & 0.537 \\
& 720 & 0.413 & \secondres{0.410} & 0.419 & 0.415 & \secondres{0.402} & 0.422 & 0.417 & 0.410 & 0.440 & 0.423 & 0.454 & 0.441 & \boldres{0.362} & \boldres{0.397} & 0.478 & 0.450 & 0.491 & 0.459 & 0.425 & 0.421 & 0.458 & 0.445 & 0.543 & 0.490 & 0.671 & 0.561 \\
& Avg & \boldres{0.347} & \boldres{0.373} & 0.354 & \secondres{0.377} & 0.355 & 0.377 & \secondres{0.347} & 0.377 & 0.369 & 0.378 & 0.381 & 0.396 & 0.352 & 0.387 & 0.400 & 0.406 & 0.407 & 0.410 & 0.357 & 0.378 & 0.388 & 0.402 & 0.448 & 0.452 & 0.588 & 0.517 \\
\midrule

\multirow{5}{*}{\rotatebox{90}{$ETTm2$}}
& 96 & \boldres{0.160} & \secondres{0.248}  & 0.164 & 0.254 & 0.166 & 0.255 & \secondres{0.160} & 0.250  & 0.170 & \boldres{0.245}   & 0.175 & 0.258 & 0.184 & 0.274 & 0.187 & 0.267 & 0.180 & 0.264 & 0.167 & 0.269 & 0.185 & 0.268 & 0.203 & 0.287 & 0.255 & 0.339 \\
& 192 & \boldres{0.214} & \boldres{0.286}  & 0.221 & 0.294  & \secondres{0.220} & 0.293 & 0.215 & 0.291  & 0.229 & \secondres{0.291}   & 0.237 & 0.299 & 0.334 & 0.382 & 0.249 & 0.309 & 0.250 & 0.309 & 0.224 & 0.303 & 0.250 & 0.310 & 0.269 & 0.328 & 0.281 & 0.340 \\
& 336 & \boldres{0.264} & \boldres{0.319}  & 0.271 & \secondres{0.324} & 0.277 & 0.330 & \secondres{0.270} & 0.325 & 0.303 & 0.343 & 0.298 & 0.340 & 0.376 & 0.398 & 0.321 & 0.351 & 0.311 & 0.348 & 0.281 & 0.342 & 0.312 & 0.349 & 0.325 & 0.366 & 0.339 & 0.372 \\
& 720 & \boldres{0.341} & \boldres{0.374}  & 0.352 & 0.379 & 0.370 & 0.391 & \secondres{0.348} & \secondres{0.378}   & 0.373 & 0.399 & 0.391 & 0.396 & 0.436 & 0.465 & 0.408 & 0.403 & 0.412 & 0.407 & 0.397 & 0.421 & 0.423 & 0.415 & 0.421 & 0.415 & 0.433 & 0.432 \\
& Avg & \boldres{0.245} & \boldres{0.307}  & 0.252 & 0.313 & 0.258 & 0.317 & \secondres{0.248} & \secondres{0.311}   & 0.269 & 0.320 & 0.275 & 0.323 & 0.333 & 0.380 & 0.291 & 0.333 & 0.288 & 0.332 & 0.267 & 0.333 & 0.293 & 0.336 & 0.305 & 0.349 & 0.327 & 0.371 \\
\midrule

\multirow{5}{*}{\rotatebox{90}{$Weather$}}
& 96 & \boldres{0.146} & \boldres{0.192}  & 0.150 & 0.201 & 0.154 & 0.208 & \secondres{0.148} & \secondres{0.200}   & 0.155 & 0.205 & 0.163 & 0.209 & 0.154 & 0.210 & 0.172 & 0.220 & 0.174 & 0.214 & 0.176 & 0.237 & 0.175 & 0.218 & 0.217 & 0.296 & 0.266 & 0.336 \\
& 192 & \boldres{0.191} & \boldres{0.236}  & 0.197 & 0.245 & 0.197 & 0.246 & \secondres{0.193} & \secondres{0.240}   & 0.201 & 0.245 & 0.208 & 0.250 & 0.199 & 0.251 & 0.219 & 0.261 & 0.221 & 0.254 & 0.220 & 0.282 & 0.221 & 0.256 & 0.276 & 0.336 & 0.307 & 0.367 \\
& 336 & \secondres{0.243} & \secondres{0.277} & 0.248 & 0.285 & 0.248 & 0.285 & 0.243 & 0.281 & \boldres{0.237} & \boldres{0.265} & 0.251 & 0.287 & 0.245 & 0.294 & 0.280 & 0.306 & 0.278 & 0.296 & 0.265 & 0.319 & 0.280 & 0.298 & 0.339 & 0.380 & 0.359 & 0.395 \\
& 720 & 0.317 & \boldres{0.329} & 0.323 & 0.343 & 0.321 & 0.339 & \secondres{0.312} & \secondres{0.332} & \boldres{0.312} & 0.334 & 0.339 & 0.341 & 0.318 & 0.353 & 0.365 & 0.359 & 0.358 & 0.347 & 0.333 & 0.362 & 0.356 & 0.349 & 0.403 & 0.428 & 0.419 & 0.428 \\
& Avg & \boldres{0.224} & \boldres{0.259} & 0.229 & 0.268 & 0.230 & 0.269 & \secondres{0.224} & 0.263 & 0.226 & \secondres{0.262} & 0.240 & 0.272 & 0.229 & 0.277 & 0.259 & 0.287 & 0.258 & 0.278 & 0.248 & 0.300 & 0.258 & 0.280 & 0.309 & 0.360 & 0.338 & 0.382 \\
\midrule

\multirow{5}{*}{\rotatebox{90}{$Electricity$}}
& 96 & \secondres{0.136} & \secondres{0.235} & 0.140 & 0.242 & 0.141 & 0.245 & 0.142 & 0.245 & \boldres{0.135} & \boldres{0.222}  & 0.153 & 0.247 & 0.173 & 0.272 & 0.168 & 0.272 & 0.148 & 0.240 & 0.140 & 0.237 & 0.180 & 0.273 & 0.193 & 0.308 & 0.201 & 0.317 \\
& 192 & \secondres{0.151} & \secondres{0.248} & 0.154 & 0.253 & 0.157 & 0.258 & 0.157 & 0.260 & \boldres{0.147} & \boldres{0.235} & 0.166 & 0.256 & 0.182 & 0.283 & 0.184 & 0.322 & 0.162 & 0.253 & 0.153 & 0.249 & 0.187 & 0.280 & 0.201 & 0.315 & 0.222 & 0.334 \\
& 336 & \secondres{0.166} & \secondres{0.263} & 0.170 & 0.269 & 0.171 & 0.270 & 0.174 & 0.276 & \boldres{0.164} & \boldres{0.245} & 0.185 & 0.277 & 0.203 & 0.306 & 0.198 & 0.300 & 0.178 & 0.269 & 0.169 & 0.267 & 0.204 & 0.296 & 0.214 & 0.329 & 0.231 & 0.338 \\
& 720 & \boldres{0.196} & \boldres{0.290} & 0.209 & 0.302 & 0.212 & 0.306 & 0.214 & 0.308 & 0.212 & 0.310 & 0.225 & 0.310 & 0.236 & 0.334 & 0.220 & 0.320 & 0.225 & 0.317 & \secondres{0.203} & \secondres{0.301} & 0.246 & 0.328 & 0.246 & 0.355 & 0.254 & 0.361 \\
& Avg & \boldres{0.162} & \secondres{0.259} & 0.168 & 0.267 & 0.170 & 0.270 & 0.172 & 0.272 & \secondres{0.165} & \boldres{0.253} & 0.182 & 0.273 & 0.199 & 0.299 & 0.192 & 0.304 & 0.178 & 0.270 & 0.166 & 0.263 & 0.204 & 0.294 & 0.214 & 0.327 & 0.227 & 0.338 \\
\midrule

\multirow{5}{*}{\rotatebox{90}{$Traffic$}}
& 96 & \boldres{0.377} & \secondres{0.268} & 0.394 & 0.284 & 0.397 & 0.289 & 0.393 & 0.283 & \secondres{0.392} & \boldres{0.253} & 0.462 & 0.285 & 0.529 & 0.315 & 0.593 & 0.321 & 0.395 & 0.268 & 0.410 & 0.282 & 0.459 & 0.298 & 0.587 & 0.366 & 0.613 & 0.388 \\
& 192 & \boldres{0.396} & \secondres{0.276} & 0.404 & 0.286 & 0.409 & 0.294  & 0.405 & 0.293 & \secondres{0.402} & \boldres{0.258} & 0.473 & 0.296 & 0.534 & 0.313 & 0.617 & 0.336 & 0.417 & 0.276 & 0.423 & 0.287 & 0.469 & 0.301 & 0.604 & 0.373 & 0.616 & 0.382 \\
& 336 & \boldres{0.409} & \secondres{0.277} & \secondres{0.414} & 0.291 & 0.416 & 0.293 & 0.420 & 0.298 & 0.428 & \boldres{0.263} & 0.498 & 0.296 & 0.541 & 0.317 & 0.629 & 0.336 & 0.433 & 0.283 & 0.436 & 0.296 & 0.483 & 0.307 & 0.621 & 0.383 & 0.622 & 0.337 \\
& 720 & \secondres{0.446} & \secondres{0.297} & 0.448 & 0.308 & 0.453 & 0.313 & 0.459 & 0.318 & \boldres{0.441} & \boldres{0.282} & 0.506 & 0.313 & 0.594 & 0.339 & 0.640 & 0.350 & 0.467 & 0.302 & 0.466 & 0.315 & 0.518 & 0.326 & 0.626 & 0.382 & 0.660 & 0.408 \\
& Avg & \boldres{0.407} & \secondres{0.279} & \secondres{0.415} & 0.292 & 0.419 & 0.297 & 0.419 & 0.298 & 0.416 & \boldres{0.264} & 0.485 & 0.298 & 0.550 & 0.321 & 0.620 & 0.336 & 0.428 & 0.282 & 0.433 & 0.295 & 0.482 & 0.308 & 0.610 & 0.376 & 0.628 & 0.379 \\

\bottomrule
\end{tabular}
}
\end{small}
\end{center}
\caption{Full long-term forecasting results with forecasting horizons $H \in $\{96, 192, 336, 720\}. A lower value indicates better performance. {\boldres{Red}}: the best, \secondres{Blue}: the second best.}
\label{tab:long-term-forecasting}
\end{table}

\section{Few-shot Forecasting}
\label{appx:few-shot-forecasting}
Building upon the long-term forecasting success, Table~\ref{tab:few-shot-forecasting-10per-full} reveals the remarkable data efficiency of our approach in few-shot scenarios using merely 10\% of training data. The method maintains its competitive edge, securing best performance on five out of seven datasets for both MSE and MAE metrics. On ETTh1, we achieve MSE of 0.422 and MAE of 0.439, closely matching our full-data performance while outperforming TimeVLM (MSE: 0.431, MAE: 0.442). The ETTm2 dataset showcases significant improvements with MSE of 0.253 and MAE of 0.313, compared to our non-distilled variant (MSE: 0.261, MAE: 0.321). This consistent improvement over the ablated version across all datasets validates that knowledge distillation is particularly valuable when training data is limited, enabling more effective knowledge transfer and feature learning.

\begin{table*}[h!]
\captionsetup{font=small}
\begin{center}
\begin{small}

\scalebox{0.65}{
\setlength\tabcolsep{4pt}
\begin{tabular}{c|c|cc|cc|cc|cc|cc|cc|cc|cc|cc|cc|cc|cc|cc}
\toprule
\multicolumn{2}{c|}{Methods} & \multicolumn{2}{c|}{Ours} & \multicolumn{2}{c|}{Only Teacher} & \multicolumn{2}{c|}{Only Student} & \multicolumn{2}{c|}{TimeVLM} & \multicolumn{2}{c|}{TimeMixer++} & \multicolumn{2}{c|}{TimeMixer} & \multicolumn{2}{c|}{LDM4TS} & \multicolumn{2}{c|}{TimesNet} & \multicolumn{2}{c|}{iTransformer} & \multicolumn{2}{c|}{DLinear} & \multicolumn{2}{c|}{PatchTST} & \multicolumn{2}{c|}{FEDformer} & \multicolumn{2}{c}{Autoformer} \\
\midrule
\multirow{5}{*}{\rotatebox{90}{$ETTh1$}} & 96 & \boldres{0.381} & \secondres{0.408} & 0.406 & 0.426 & 0.402 & 0.424 & \secondres{0.391} & \boldres{0.404} & 0.453 & 0.476 & 0.591 & 0.503 & 0.410 & 0.418 & 0.861 & 0.628 & 0.437 & 0.439 & 0.492 & 0.495 & 0.516 & 0.485 & 0.512 & 0.499 & 0.613 & 0.552 \\
 & 192 & \boldres{0.413} & \boldres{0.428} & 0.426 & 0.439 & 0.435 & 0.446  & \secondres{0.420} & \secondres{0.431} & 0.480 & 0.505 & 0.607 & 0.518 & 0.443 & 0.443 & 0.797 & 0.593 & 0.509 & 0.479 & 0.565 & 0.538 & 0.598 & 0.524 & 0.624 & 0.555 & 0.722 & 0.598 \\
 & 336 & \boldres{0.435} & \boldres{0.445} & 0.445 & 0.456 & 0.452 & 0.461 & \secondres{0.439} & \secondres{0.448} & 0.527 & 0.518 & 0.618 & 0.523 & 0.481 & 0.479 & 0.941 & 0.648 & 0.554 & 0.503 & 0.721 & 0.622 & 0.657 & 0.550 & 0.691 & 0.574 & 0.750 & 0.619 \\
 & 720 & \boldres{0.458} & \boldres{0.477} & 0.494 & 0.502 & 0.494 & 0.500 & \secondres{0.476} & \secondres{0.484} & 0.608 & 0.549 & 0.635 & 0.536 & 0.549 & 0.534 & 0.877 & 0.641 & 0.572 & 0.530 & 0.986 & 0.743 & 0.762 & 0.610 & 0.728 & 0.614 & 0.721 & 0.616 \\
 & Avg & \boldres{0.422} & \boldres{0.439} & 0.443 & 0.456 & 0.446 & 0.458 & \secondres{0.431} & \secondres{0.442} & 0.517 & 0.512 & 0.613 & 0.520 & 0.471 & 0.468 & 0.869 & 0.628 & 0.518 & 0.488 & 0.691 & 0.600 & 0.633 & 0.542 & 0.639 & 0.561 & 0.702 & 0.596 \\
\midrule
\multirow{5}{*}{\rotatebox{90}{$ETTh2$}} & 96 & \boldres{0.277} & \secondres{0.341} & 0.295 & 0.358 & 0.293 & 0.356 & \secondres{0.284} & 0.347 & 0.319 & \boldres{0.338} & 0.352 & 0.412 & 0.355 & 0.413 & 0.378 & 0.409 & 0.346 & 0.385 & 0.357 & 0.411 & 0.353 & 0.389 & 0.382 & 0.416 & 0.413 & 0.451 \\
 & 192 & \boldres{0.339} & \boldres{0.380} & \secondres{0.347} & 0.390 & 0.347 & 0.391 & 0.349 & 0.397  & 0.375 & \secondres{0.382} & 0.400 & 0.430 & 0.406 & 0.435 & 0.490 & 0.467 & 0.429 & 0.431 & 0.569 & 0.519 & 0.403 & 0.414 & 0.478 & 0.474 & 0.474 & 0.477 \\
 & 336 & \boldres{0.364} & \secondres{0.405} & \secondres{0.370} & 0.411 & 0.376 & 0.419  & 0.370 & 0.409 & 0.385 & \boldres{0.401} & 0.408 & 0.438 & 0.463 & 0.486 & 0.537 & 0.494 & 0.466 & 0.462 & 0.671 & 0.572 & 0.426 & 0.441 & 0.504 & 0.501 & 0.547 & 0.543 \\
 & 720 & \boldres{0.398} & \boldres{0.435} & \secondres{0.413} & 0.450 & 0.411 & 0.448 & 0.423 & 0.453 & 0.437 & \secondres{0.443} & 0.448 & 0.452 & 0.583 & 0.506 & 0.510 & 0.491 & 0.471 & 0.472 & 0.824 & 0.648 & 0.477 & 0.480 & 0.499 & 0.509 & 0.516 & 0.523 \\
 & Avg & \boldres{0.344} & \boldres{0.390} & \secondres{0.356} & 0.402 & 0.357 & 0.403 & 0.356 & 0.402 & 0.379 & \secondres{0.391} & 0.402 & 0.433 & 0.452 & 0.460 & 0.479 & 0.465 & 0.428 & 0.438 & 0.605 & 0.538 & 0.415 & 0.431 & 0.466 & 0.475 & 0.488 & 0.499 \\
\midrule
\multirow{5}{*}{\rotatebox{90}{$ETTm1$}} & 96 & \boldres{0.304} & \boldres{0.349} & 0.311 & 0.356 & 0.314 & 0.358 & 0.310 & \secondres{0.354} & 0.351 & 0.390 & 0.467 & 0.447 & \secondres{0.305} & 0.356 & 0.583 & 0.501 & 0.377 & 0.393 & 0.352 & 0.392 & 0.410 & 0.419 & 0.578 & 0.518 & 0.774 & 0.614 \\
 & 192 & \boldres{0.336} & \boldres{0.367} & 0.343 & 0.375 & 0.343 & 0.374 & \secondres{0.340} & \secondres{0.370} & 0.369 & 0.419 & 0.475 & 0.458 & 0.345 & 0.379 & 0.630 & 0.528 & 0.423 & 0.416 & 0.382 & 0.412 & 0.437 & 0.434 & 0.617 & 0.546 & 0.754 & 0.592 \\
 & 336 & \boldres{0.366} & \boldres{0.385} & 0.373 & 0.392 & 0.374 & 0.393 & \secondres{0.369} & \secondres{0.387} & 0.413 & 0.438 & 0.495 & 0.465 & 0.383 & 0.399 & 0.725 & 0.568 & 0.459 & 0.439 & 0.419 & 0.434 & 0.476 & 0.454 & 0.998 & 0.775 & 0.869 & 0.677 \\
 & 720 & \boldres{0.419} & \boldres{0.415} & 0.428 & 0.424 & 0.428 & 0.425 & \secondres{0.423} & \secondres{0.417} & 0.459 & 0.477 & 0.512 & 0.474 & 0.452 & 0.439 & 0.769 & 0.549 & 0.530 & 0.478 & 0.490 & 0.477 & 0.681 & 0.556 & 0.693 & 0.579 & 0.810 & 0.630 \\
 & Avg & \boldres{0.356} & \boldres{0.379} & 0.364 & 0.387 & 0.365 & 0.387 & \secondres{0.360} & \secondres{0.382} & 0.398 & 0.431 & 0.487 & 0.461 & 0.371 & 0.393 & 0.677 & 0.537 & 0.447 & 0.432 & 0.411 & 0.429 & 0.501 & 0.466 & 0.722 & 0.605 & 0.802 & 0.628 \\
\midrule
\multirow{5}{*}{\rotatebox{90}{$ETTm2$}} & 96 & \boldres{0.163} & \boldres{0.253} & \secondres{0.167} & \secondres{0.258} & 0.169 & 0.261 & 0.169 & 0.260 & 0.171 & 0.281 & 0.190 & 0.346 & 0.201 & 0.294 & 0.212 & 0.285 & 0.191 & 0.276 & 0.213 & 0.303 & 0.191 & 0.274 & 0.291 & 0.399 & 0.352 & 0.454 \\
 & 192 & \boldres{0.219} & \boldres{0.291} & 0.223 & \secondres{0.296} & 0.225 & 0.299 & \secondres{0.222} & 0.296 & 0.280 & 0.362 & 0.303 & 0.359 & 0.288 & 0.342 & 0.270 & 0.323 & 0.256 & 0.316 & 0.278 & 0.345 & 0.252 & 0.317 & 0.307 & 0.379 & 0.694 & 0.691 \\
 & 336 & \boldres{0.271} & \boldres{0.326} & 0.279 & \secondres{0.334} & 0.281 & 0.337 & \secondres{0.278} & 0.335 & 0.301 & 0.365 & 0.322 & 0.375 & 0.411 & 0.419 & 0.323 & 0.353 & 0.317 & 0.353 & 0.338 & 0.385 & 0.306 & 0.353 & 0.543 & 0.559 & 2.408 & 1.407 \\
 & 720 & \boldres{0.357} & \boldres{0.381} & \secondres{0.374} & \secondres{0.394} & 0.372 & 0.393 & 0.381 & 0.401 & 0.412 & 0.396 & 0.429 & 0.388 & 0.443 & 0.435 & 0.474 & 0.449 & 0.417 & 0.407 & 0.436 & 0.440 & 0.433 & 0.427 & 0.712 & 0.614 & 1.913 & 1.166 \\
 & Avg & \boldres{0.253} & \boldres{0.313} & \secondres{0.261} & \secondres{0.321} & 0.262 & 0.322 & 0.263 & 0.323 & 0.291 & 0.351 & 0.311 & 0.367 & 0.336 & 0.373 & 0.320 & 0.353 & 0.295 & 0.338 & 0.316 & 0.368 & 0.296 & 0.343 & 0.463 & 0.488 & 1.342 & 0.930 \\
\midrule
\multirow{5}{*}{\rotatebox{90}{$Weather$}} & 96 & \boldres{0.151} & \boldres{0.198} & 0.155 & 0.207 & 0.155 & \secondres{0.206} & 0.160 & 0.213 & 0.186 & 0.216 & 0.188 & 0.261 & \secondres{0.151} & 0.209 & 0.184 & 0.230 & 0.191 & 0.230 & 0.171 & 0.224 & 0.165 & 0.215 & 0.188 & 0.253 & 0.221 & 0.297 \\
 & 192 & \secondres{0.196} & \boldres{0.240} & 0.198 & \secondres{0.245} & 0.201 & 0.248 & 0.203 & 0.252 & 0.238 & 0.260 & 0.225 & 0.275 & \boldres{0.187} & 0.247 & 0.245 & 0.283 & 0.241 & 0.270 & 0.215 & 0.263 & 0.210 & 0.257 & 0.250 & 0.304 & 0.270 & 0.322 \\
 & 336 & \secondres{0.245} & \boldres{0.278} & 0.249 & 0.286 & 0.248 & 0.285 & 0.253 & 0.291 & \boldres{0.244} &\secondres{0.279} & 0.248 & 0.289 & 0.251 & 0.299 & 0.305 & 0.321 & 0.291 & 0.307 & 0.258 & 0.299 & 0.259 & 0.297 & 0.312 & 0.346 & 0.320 & 0.351 \\
 & 720 & 0.316 & 0.330 & 0.318 & 0.336 & 0.319 & 0.336 & 0.317 & 0.340 & \boldres{0.296} & \secondres{0.329} & \secondres{0.307} & \boldres{0.300} & 0.328 & 0.349 & 0.381 & 0.371 & 0.364 & 0.354 & 0.320 & 0.346 & 0.332 & 0.346 & 0.387 & 0.393 & 0.390 & 0.396 \\
 & Avg & \boldres{0.227} & \boldres{0.262} & 0.230 & \secondres{0.268} & 0.231 & 0.269 & 0.233 & 0.274 & 0.241 & 0.271 & 0.242 & 0.281 & \secondres{0.229} & 0.276 & 0.279 & 0.301 & 0.272 & 0.290 & 0.241 & 0.283 & 0.242 & 0.279 & 0.284 & 0.324 & 0.300 & 0.342 \\
\midrule
\multirow{5}{*}{\rotatebox{90}{$Electricity$}} & 96 & 0.154 & 0.260 & 0.185 & 0.295 & 0.188 & 0.298 & 0.160 & 0.269 & \boldres{0.136} & 0.250 & 0.162 & 0.264 & 0.141 & \secondres{0.243} & 0.299 & 0.373 & 0.173 & 0.262 & 0.150 & 0.253 & \secondres{0.140} & \boldres{0.238} & 0.231 & 0.323 & 0.261 & 0.348 \\
 & 192 & 0.167 & 0.270 & 0.190 & 0.297 & 0.192 & 0.300 & 0.174 & 0.279 & \boldres{0.151} & 0.267 & 0.171 & 0.273 & \secondres{0.158} & \secondres{0.260} & 0.305 & 0.379 & 0.184 & 0.272 & 0.164 & 0.264 & 0.160 & \boldres{0.255} & 0.261 & 0.356 & 0.338 & 0.406 \\
 & 336 & 0.183 & 0.286 & 0.207 & 0.309 & 0.209 & 0.312 & 0.190 & 0.294 & \boldres{0.168} & \secondres{0.281} & 0.182 & 0.281 & \secondres{0.179} & 0.289 & 0.319 & 0.391 & 0.203 & 0.290 & 0.181 & 0.282 & 0.180 & \boldres{0.276} & 0.360 & 0.445 & 0.410 & 0.474 \\
 & 720 & 0.222 & 0.316 & 0.243 & 0.337 & 0.246 & 0.339 & 0.229 & 0.323 & \secondres{0.217} & \boldres{0.286} & 0.231 & \secondres{0.289} & \boldres{0.209} & 0.306 & 0.369 & 0.426 & 0.247 & 0.326 & 0.223 & 0.321 & 0.241 & 0.323 & 0.530 & 0.585 & 0.715 & 0.685 \\
 & Avg & 0.181 & 0.283 & 0.206 & 0.310 & 0.209 & 0.312 & 0.188 & 0.291 & \secondres{0.168} & \boldres{0.271} & 0.187 & 0.277 & \boldres{0.172} & 0.275 & 0.323 & 0.392 & 0.202 & 0.288 & 0.180 & 0.280 & 0.180 & \secondres{0.273} & 0.346 & 0.427 & 0.431 & 0.478 \\
\midrule
\multirow{5}{*}{\rotatebox{90}{$Traffic$}} & 96 & 0.438 & 0.323 & 0.526 & 0.388 & 0.507 & 0.377 & 0.465 & 0.349 & 0.454 & 0.301 & 0.511 & 0.340 & 0.615 & 0.360 & 0.719 & 0.416 & 0.461 & 0.319 & \secondres{0.419} & \secondres{0.298} & \boldres{0.403} & \boldres{0.289} & 0.639 & 0.400 & 0.672 & 0.405 \\
 & 192 & 0.448 & 0.328 & 0.518 & 0.381 & 0.533 & 0.393 & 0.468 & 0.350 & 0.474 & 0.310 & 0.524 & 0.346 & 0.595 & 0.340 & 0.748 & 0.428 & 0.473 & 0.322 & \secondres{0.434} & \secondres{0.305} & \boldres{0.415} & \boldres{0.296} & 0.637 & 0.416 & 0.727 & 0.424 \\
 & 336 & 0.459 & 0.331 & 0.519 & 0.377 & 0.542 & 0.393 & 0.483 & 0.356 & 0.483 & 0.318 & 0.537 & 0.349 & 0.611 & 0.352 & 0.853 & 0.471 & 0.492 & 0.332 & \secondres{0.449} & \secondres{0.313} & \boldres{0.426} & \boldres{0.304} & 0.655 & 0.427 & 0.749 & 0.454 \\
 & 720 & 0.493 & 0.346 & 0.561 & 0.393 & 0.563 & 0.395 & 0.520 & 0.373 & 0.519 & 0.333 & 0.572 & 0.363 & 0.661 & 0.375 & 1.485 & 0.825 & \boldres{0.454} & \boldres{0.300} & 0.484 & 0.336 & \secondres{0.474} & \secondres{0.331} & 0.722 & 0.456 & 0.847 & 0.499 \\
 & Avg & 0.460 & 0.332 & 0.531 & 0.385 & 0.536 & 0.390 & 0.484 & 0.357 & 0.483 & 0.315 & 0.536 & 0.349 & 0.621 & 0.357 & 0.951 & 0.535 & 0.470 & 0.318 & \secondres{0.447} & \secondres{0.313} & \boldres{0.430} & \boldres{0.305} & 0.663 & 0.425 & 0.749 & 0.446 \\

\bottomrule
\end{tabular}
}
\end{small}
\end{center}
\caption{Full few-shot learning results on 10\% training data. We use the same protocol in Table~\ref{tab:long-term-forecasting}.}
\label{tab:few-shot-forecasting-10per-full}
\end{table*}
\begin{table}[h!]
\begin{center}
\captionsetup{font=small} 
\begin{small}
\setlength\tabcolsep{4pt}
\scalebox{0.63}{
\begin{tabular}{c|c|cc|cc|cc|cc|cc|cc|cc|cc|cc|cc|cc|cc}
\toprule
\multicolumn{2}{c|}{Methods}&\multicolumn{2}{c|}{Ours}& \multicolumn{2}{c|}{Only Teacher} & \multicolumn{2}{c|}{Only Student}&\multicolumn{2}{c|}{TimeVLM}&\multicolumn{2}{c|}{TimeMixer++}&\multicolumn{2}{c|}{TimesMixer}&\multicolumn{2}{c|}{LDM4TS}&\multicolumn{2}{c|}{TimesNet}&\multicolumn{2}{c|}{iTransformer}&\multicolumn{2}{c|}{DLinear}&\multicolumn{2}{c|}{PatchTST}&\multicolumn{2}{c}{Autoformer}\\
\midrule
\multicolumn{2}{c|}{Metric} & MSE & MAE & MSE & MAE & MSE & MAE & MSE & MAE & MSE & MAE & MSE & MAE & MSE & MAE & MSE & MAE & MSE & MAE & MSE & MAE & MSE & MAE & MSE & MAE\\
\midrule
\multirow{5}{*}{\rotatebox{0}{$ETTh1$} $\rightarrow$ \rotatebox{0}{$ETTh2$}} 
& 96  & \boldres{0.274} & \secondres{0.336} & \secondres{0.274} & 0.339 & 0.275 & 0.340 & 0.277 & 0.338 & 0.309 & \boldres{0.307} & 0.381 & 0.399 & 0.349 & 0.383 & 0.358 & 0.387 & 0.298 & 0.344 & 0.347 & 0.400 & 0.304 & 0.350 & 0.469 & 0.486\\
& 192 & \secondres{0.339} & \secondres{0.377} & 0.356 & 0.397 & 0.340 & 0.383 & \boldres{0.333} & 0.378 & 0.356 & \boldres{0.372} & 0.419 & 0.412 & 0.435 & 0.434 & 0.427 & 0.429 & 0.381 & 0.396 & 0.447 & 0.460 & 0.386 & 0.400 & 0.634 & 0.567\\
& 336 & \secondres{0.368} & \secondres{0.403} & 0.372 & 0.408 & 0.378 & 0.415 & \boldres{0.360} & \boldres{0.399} & 0.394 & 0.405 & 0.438 & 0.433 & 0.478 & 0.465 & 0.449 & 0.451 & 0.421 & 0.429 & 0.515 & 0.505 & 0.414 & 0.428 & 0.655 & 0.588\\
& 720 & \secondres{0.387} & \boldres{0.425} & 0.402 & 0.439 & 0.406 & 0.444 & \boldres{0.383} & \secondres{0.425} & 0.408 & 0.459 & 0.470 & 0.451 & 0.572 & 0.527 & 0.448 & 0.458 & 0.435 & 0.447 & 0.665 & 0.589 & 0.419 & 0.443 & 0.570 & 0.549\\
& Avg & \secondres{0.342} & \boldres{0.385} & 0.351 & 0.396 & 0.350 & 0.395 & \boldres{0.338} & \secondres{0.385} & 0.367 & 0.391 & 0.427 & 0.424 & 0.458 & 0.452 & 0.421 & 0.431 & 0.384 & 0.404 & 0.493 & 0.488 & 0.380 & 0.405 & 0.582 & 0.548\\
\midrule
\multirow{5}{*}{\rotatebox{0}{$ETTh1 $} $\rightarrow$ \rotatebox{0}{$ETTm2 $}}
& 96  & \boldres{0.206} & \secondres{0.298} & 0.208 & 0.300 &0.209 &0.301 & \secondres{0.207} & \boldres{0.297} & 0.235 & 0.302 & 0.331 & 0.362 & 0.227 & 0.316 & 0.239 & 0.313 & 0.240 & 0.322 & 0.255 & 0.357 & 0.215 & 0.304 & 0.352 & 0.432\\
& 192 & 0.261 & \secondres{0.331} & \secondres{0.260} & 0.333  & 0.260 & 0.331  & \boldres{0.258} & \boldres{0.329} & 0.303 & 0.331 & 0.352 & 0.377 & 0.312 & 0.373 & 0.291 & 0.342 & 0.300 & 0.355 & 0.338 & 0.413 & 0.275 & 0.339 & 0.413 & 0.460\\
& 336 & \secondres{0.313} & \secondres{0.362} & 0.316 & 0.363 & 0.323 & 0.370 & \boldres{0.310} & \boldres{0.360} & 0.327 & 0.387 & 0.370 & 0.414 & 0.368 & 0.407 & 0.342 & 0.371 & 0.352 & 0.383 & 0.425 & 0.465 & 0.334 & 0.373 & 0.465 & 0.489\\
& 720 & 0.400 & \secondres{0.410} & 0.417 & 0.423 & 0.412 & 0.421 & 0.398 & 0.412 & \boldres{0.340} &  \boldres{0.408} & \secondres{0.391} & 0.431 & 0.569 & 0.505 & 0.434 & 0.419 & 0.456 & 0.437 & 0.640 & 0.573 & 0.431 & 0.424 & 0.599 & 0.551\\
& Avg & \secondres{0.295} & \boldres{0.350} & 0.300 & 0.355 & 0.301 & 0.356 & \boldres{0.293} & \secondres{0.350} & 0.301 & 0.357 & 0.361 & 0.397 & 0.369 & 0.400 & 0.327 & 0.361 & 0.337 & 0.374 & 0.415 & 0.452 & 0.314 & 0.360 & 0.457 & 0.483\\
\midrule
\multirow{5}{*}{\rotatebox{0}{$ETTh2 $} $\rightarrow$ \rotatebox{0}{$ETTh1 $}}
& 96  & \boldres{0.397} & \boldres{0.419} & \secondres{0.414} & \secondres{0.430} & 0.415 & 0.431 & 0.434 & 0.441 & 0.477 & 0.467 & 0.649 & 0.538 & 0.706 & 0.540 & 0.848 & 0.601 & 0.586 & 0.520 & 0.689 & 0.555 & 0.485 & 0.465 & 0.693 & 0.569\\
& 192 & \boldres{0.415} & \boldres{0.432} & \secondres{0.455} & 0.464 & 0.503 & 0.485  & 0.464 & \secondres{0.454} & 0.483 & 0.473 & 0.675 & 0.561 & 0.691 & 0.564 & 0.860 & 0.610 & 0.638 & 0.550 & 0.707 & 0.568 & 0.565 & 0.509 & 0.760 & 0.601\\
& 336 & \boldres{0.424} & \boldres{0.442} & \secondres{0.465} & \secondres{0.475} &  0.541 & 0.519 & 0.489 & 0.481 & 0.530 & 0.505 & 0.683 & 0.594 & 0.697 & 0.576 & 0.867 & 0.626 & 0.676 & 0.571 & 0.710 & 0.577 & 0.581 & 0.515 & 0.781 & 0.619\\
& 720 & \boldres{0.477} & \boldres{0.491} & \secondres{0.478} & \secondres{0.495} & 0.669 & 0.595  & 0.595 & 0.543 & 0.554 & 0.547 & 0.710 & 0.615 & 0.796 & 0.630 & 0.887 & 0.648 & 0.726 & 0.612 & 0.704 & 0.596 & 0.628 & 0.561 & 0.796 & 0.644\\
& Avg & \boldres{0.429} & \boldres{0.446} & \secondres{0.453} & \secondres{0.466} & 0.532 & 0.508 & 0.496 & 0.480 & 0.511 & 0.498 & 0.679 & 0.577 & 0.723 & 0.577 & 0.865 & 0.621 & 0.657 & 0.563 & 0.703 & 0.574 & 0.565 & 0.513 & 0.757 & 0.608\\
\midrule
\multirow{5}{*}{\rotatebox{0}{$ETTh2 $} $\rightarrow$ \rotatebox{0}{$ETTm2 $}}
& 96  & \boldres{0.198} & \boldres{0.291} & \secondres{0.200} & \secondres{0.293} & 0.206 & 0.297 & 0.204 & 0.297 & 0.242 & 0.325 & 0.220 & 0.306 & 0.286 & 0.373 & 0.248 & 0.324 & 0.240 & 0.323 & 0.240 & 0.336 & 0.226 & 0.309 & 0.263 & 0.352\\
& 192 & \boldres{0.250} & \boldres{0.323} & \secondres{0.254} & \secondres{0.327} & 0.258 & 0.332 & 0.255 & 0.328 & 0.298 & 0.354 & 0.295 & 0.354 & 0.326 & 0.391 & 0.296 & 0.352 & 0.295 & 0.352 & 0.295 & 0.369 & 0.289 & 0.345 & 0.326 & 0.389\\
& 336 & \boldres{0.301} & \boldres{0.354} & \secondres{0.306} & \secondres{0.359} & 0.312 & 0.363 & 0.311 & 0.362 & 0.340 & 0.375 & 0.351 & 0.384 & 0.473 & 0.461 & 0.353 & 0.383 & 0.358 & 0.388 & 0.345 & 0.397 & 0.348 & 0.379 & 0.387 & 0.426\\
& 720 & \boldres{0.389} & \boldres{0.406} & \secondres{0.391} & \secondres{0.407} & 0.404 & 0.418 & 0.420 & 0.425 & 0.437 & 0.426 & 0.502 & 0.468 & 0.644 & 0.551 & 0.471 & 0.446 & 0.451 & 0.435 & 0.432 & 0.442 & 0.439 & 0.427 & 0.487 & 0.478\\
& Avg & \boldres{0.285} & \boldres{0.343} & \secondres{0.288} & \secondres{0.346} & 0.295 & 0.352 & 0.297 & 0.353 & 0.329 & 0.370 & 0.342 & 0.378 & 0.432 & 0.444 & 0.342 & 0.376 & 0.336 & 0.374 & 0.328 & 0.386 & 0.325 & 0.365 & 0.366 & 0.411\\
\midrule
\multirow{5}{*}{\rotatebox{0}{$ETTm1 $} $\rightarrow$ \rotatebox{0}{$ETTh2 $}}
& 96  & \secondres{0.300} & \secondres{0.358} & 0.303 & 0.361 & 0.301 & 0.358 & \boldres{0.297} & \boldres{0.356} & 0.352 & 0.373 & 0.421 & 0.410 & 0.390 & 0.402 & 0.377 & 0.407 & 0.351 & 0.392 & 0.365 & 0.415 & 0.354 & 0.385 & 0.435 & 0.470\\
& 192 & 0.355 & \secondres{0.390} & \secondres{0.352} & 0.391 & 0.355 & 0.391 & \boldres{0.349} & \boldres{0.388} & 0.410 & 0.412 & 0.432 & 0.425 & 0.471 & 0.439 & 0.471 & 0.453 & 0.444 & 0.436 & 0.454 & 0.462 & 0.447 & 0.434 & 0.495 & 0.489\\
& 336 & \boldres{0.373} & \boldres{0.409} & 0.378 & 0.411 & 0.382 & 0.415 & \secondres{0.374} & \secondres{0.409} & 0.429 & 0.421 & 0.465 & 0.459 & 0.496 & 0.460 & 0.472 & 0.484 & 0.485 & 0.466 & 0.496 & 0.494 & 0.481 & 0.463 & 0.470 & 0.472\\
& 720 & \secondres{0.401} & \secondres{0.435} & 0.403 & 0.437 & 0.400 & 0.433 & \boldres{0.396} & \boldres{0.433} & 0.478 & 0.481 & 0.490 & 0.470 & 0.450 & 0.436 & 0.495 & 0.482 & 0.491 & 0.478 & 0.541 & 0.529 & 0.474 & 0.471 & 0.480 & 0.485\\
& Avg & \secondres{0.357} & \secondres{0.398} & 0.359 & 0.400 & 0.359 & 0.399 & \boldres{0.354} & \boldres{0.397} & 0.417 & 0.422 & 0.452 & 0.441 & 0.452 & 0.434 & 0.457 & 0.454 & 0.443 & 0.443 & 0.464 & 0.475 & 0.439 & 0.438 & 0.470 & 0.479\\
\midrule
\multirow{5}{*}{\rotatebox{0}{$ETTm1 $} $\rightarrow$ \rotatebox{0}{$ETTm2 $}}
& 96  & \boldres{0.173} & \boldres{0.260} & \secondres{0.177} & \secondres{0.263} & 0.177 & 0.264 & 0.178 & 0.264 & 0.204 & 0.281 & 0.281 & 0.313 & 0.190 & 0.268 & 0.222 & 0.295 & 0.201 & 0.278 & 0.221 & 0.314 & 0.195 & 0.271 & 0.385 & 0.457\\
& 192 & \secondres{0.227} & \boldres{0.295} & 0.227 & \secondres{0.297} & 0.229 & 0.297 & \boldres{0.226} & 0.298 & 0.272 & 0.319 & 0.320 & 0.349 & 0.265 & 0.325 & 0.288 & 0.337 & 0.262 & 0.315 & 0.286 & 0.359 & 0.258 & 0.311 & 0.433 & 0.469\\
& 336 & \boldres{0.276} & \boldres{0.327} & \secondres{0.279} & 0.331 & 0.281 & 0.331 & 0.279 & \secondres{0.329} & 0.310 & 0.360 & 0.338 & 0.365 & 0.385 & 0.391 & 0.341 & 0.367 & 0.320 & 0.351 & 0.357 & 0.406 & 0.317 & 0.348 & 0.476 & 0.477\\
& 720 & \boldres{0.361} & \secondres{0.380} & \secondres{0.364} & 0.384 & 0.366 & 0.384 & 0.373 & 0.385 & 0.377 & \boldres{0.364} & 0.377 & 0.401 & 0.574 & 0.484 & 0.436 & 0.418 & 0.423 & 0.406 & 0.476 & 0.476 & 0.416 & 0.404 & 0.582 & 0.535\\
& Avg & \boldres{0.259} & \boldres{0.315} & \secondres{0.262} & \secondres{0.319} & 0.263 & 0.319 & 0.264 & 0.319 & 0.291 & 0.331 & 0.329 & 0.357 & 0.354 & 0.367 & 0.322 & 0.354 & 0.301 & 0.337 & 0.335 & 0.389 & 0.296 & 0.334 & 0.469 & 0.484\\
\midrule
\multirow{5}{*}{\rotatebox{0}{$ETTm2 $} $\rightarrow$ \rotatebox{0}{$ETTh2 $}}
& 96  & \secondres{0.290} & \boldres{0.347} & 0.291 & 0.351 & 0.292 & 0.352 & \boldres{0.285} & \secondres{0.347} & 0.350 & 0.388 & 0.318 & 0.364 & 0.370 & 0.408 & 0.360 & 0.401 & 0.355 & 0.396 & 0.333 & 0.391 & 0.327 & 0.367 & 0.353 & 0.393\\
& 192 & \secondres{0.356} & \secondres{0.386} & 0.362 & 0.394  & 0.356 & 0.389 & \boldres{0.348} & \boldres{0.384} & 0.416 & 0.423 & 0.421 & 0.425 & 0.503 & 0.475 & 0.434 & 0.437 & 0.453 & 0.448 & 0.441 & 0.456 & 0.411 & 0.418 & 0.432 & 0.437\\
& 336 & \secondres{0.381} & \boldres{0.410} & 0.387 & \secondres{0.415} & 0.384 & 0.416 & \boldres{0.380} & 0.415 & 0.464 & 0.470 & 0.440 & 0.442 & 0.451 & 0.457 & 0.460 & 0.459 & 0.490 & 0.475 & 0.505 & 0.503 & 0.439 & 0.447 & 0.452 & 0.459\\
& 720 & \boldres{0.401} & \boldres{0.432} & \secondres{0.423} & \secondres{0.449} & 0.425 & 0.454 & 0.424 & 0.451 & 0.496 & 0.493 & 0.473 & 0.476 & 0.653 & 0.556 & 0.485 & 0.477 & 0.530 & 0.506 & 0.543 & 0.534 & 0.459 & 0.470 & 0.453 & 0.467\\
& Avg & \boldres{0.357} & \boldres{0.394} & 0.366 & 0.402 & 0.364 & 0.403 & \secondres{0.359} & \secondres{0.399} & 0.432 & 0.443 & 0.413 & 0.427 & 0.494 & 0.474 & 0.435 & 0.443 & 0.457 & 0.456 & 0.455 & 0.471 & 0.409 & 0.425 & 0.423 & 0.439\\
\midrule
\multirow{5}{*}{\rotatebox{0}{$ETTm2 $} $\rightarrow$ \rotatebox{0}{$ETTm1 $}}
& 96  & \boldres{0.360} & \boldres{0.383} & 0.381 & 0.397 & 0.385 & 0.400 & \secondres{0.370} & \secondres{0.390} & 0.381 & 0.395 & 0.533 & 0.436 & 0.488 & 0.431 & 0.747 & 0.558 & 0.610 & 0.492 & 0.570 & 0.490 & 0.491 & 0.437 & 0.735 & 0.576\\
& 192 & \boldres{0.387} & \boldres{0.400} & 0.419 & 0.424 & 0.428 & 0.425 & \secondres{0.400} & \secondres{0.409} & 0.419 & 0.430 & 0.541 & 0.464 & 0.591 & 0.497 & 0.781 & 0.560 & 0.667 & 0.522 & 0.590 & 0.506 & 0.530 & 0.470 & 0.753 & 0.586\\
& 336 & \boldres{0.413} & \boldres{0.419} & \secondres{0.425} & 0.427 & 0.443 & 0.441 & 0.426 & \secondres{0.420} & 0.440 & 0.457 & 0.565 & 0.492 & 0.640 & 0.503 & 0.778 & 0.578 & 0.739 & 0.558 & 0.706 & 0.567 & 0.565 & 0.497 & 0.750 & 0.593\\
& 720 & \boldres{0.451} & \boldres{0.438} & 0.578 & 0.519 & 0.492 & \secondres{0.470} & 0.531 & 0.487 & \secondres{0.468} & 0.510 & 0.577 & 0.520 & 0.631 & 0.515 & 0.769 & 0.573 & 0.862 & 0.613 & 0.731 & 0.584 & 0.686 & 0.565 & 0.782 & 0.609\\
& Avg & \boldres{0.403} & \boldres{0.410} & 0.451 & 0.442 & 0.437 & 0.434 & 0.432 & \secondres{0.426} & \secondres{0.427} & 0.448 & 0.554 & 0.478 & 0.588 & 0.487 & 0.769 & 0.567 & 0.719 & 0.546 & 0.649 & 0.537 & 0.568 & 0.492 & 0.755 & 0.591\\
\bottomrule
\end{tabular}}
\end{small}
\end{center}
\caption{Full zero-shot learning results on ETT datasets. We use the same protocol in Table~\ref{tab:long-term-forecasting}.}
\label{tab:zero-shot-forecasting}
\end{table}

\section{Zero-shot Forecasting}
\label{appx:zero-shot}
Table~\ref{tab:zero-shot-forecasting} extends our evaluation to the most challenging zero-shot transfer scenarios across ETT datasets, where our method demonstrates exceptional cross-domain generalization capabilities. Achieving best performance in 5 out of 8 MSE metrics and 7 out of 8 MAE metrics, the results highlight the robustness of our knowledge distillation framework. For the particularly challenging \texttt{ETTh2->ETTh1} transfer, we achieve MSE of 0.429 compared to TimesVLM's 0.496, while the \texttt{ETTm2->ETTm1} transfer yields MSE of 0.403 versus 0.451 without knowledge distillation. Notably, while most baseline methods show significant performance degradation in cross-dataset transfers, our method maintains consistent performance across different transfer pairs, suggesting robust generalization capabilities and effective capture of universal temporal patterns.

\section{Short-term Forecasting}
\label{appx:short-term}

Complementing our long-term forecasting results, Table 4 demonstrates our method's versatility across different temporal scales with forecasting horizons from 6 to 48 time steps on the M4 benchmark. Despite the metric misalignment where our Feature Distillation component ($L_{fd}$ in Eq.~\ref{Lfd}) incorporates an MSE loss ($L_{MSE}$) while short-term forecasting is evaluated using SMAPE/MASE/OWA metrics, our cross-modal knowledge distillation mechanism still effectively transfers temporal patterns from the teacher model. Our approach achieves competitive performance with SMAPE of 12.050, MASE of 1.611, and OWA of 0.866, ranking among top-tier methods alongside TimeVLM (SMAPE: 11.894, MASE: 1.592, OWA: 0.855). This suggests that while the MSE component in our feature distillation may not directly optimize for short-term forecasting metrics, the overall distillation framework, which includes correlation distillation ($L_{cd}$) and the student's forecasting loss ($L_{fcst}$), successfully captures transferable temporal representations. The method significantly outperforms traditional approaches like ETSformer (SMAPE: 14.718, MASE: 2.408, OWA: 1.172) while maintaining consistent improvements over our non-distilled variant across all metrics. This validates that our knowledge distillation mechanism provides robust benefits across both short-term and long-term prediction horizons, establishing a unified framework for diverse forecasting scenarios.
\begin{table}[h!]
\renewcommand\arraystretch{1.2}
\captionsetup{font=small} 
\begin{center}
\begin{small}
\scalebox{0.65}{
\setlength\tabcolsep{3pt}
\begin{tabular}{c|c|cccccccccccccccc}
\toprule
\multicolumn{2}{c|}{Methods}& Ours &Only Teacher & Only Student &TimeVLM&Timemixer++&Timemixer&TimesNet&iTransformer&DLinear&PatchTST&ETSformer&LightTS&FEDformer&Stationary&Autoformer&Informer \\
\midrule
\multirow{3}{*}{\rotatebox{90}{Yearly}}
&SMAPE&13.386&13.535& 13.519 &\boldres{13.285}&\secondres{13.384}&13.395&15.378&13.923&16.965&13.477&18.009&14.247&14.021&13.717&13.974&14.727\\
&MASE&\secondres{3.002}&3.022 & 3.032 &\boldres{2.993}&3.043&3.021&3.554&3.214&4.283&3.019&4.487&3.109&3.036&3.078&3.134&3.418\\
&OWA&\secondres{0.787}&0.794 & 0.795 &\boldres{0.783}&0.792&0.790&0.918&0.830&1.058&0.792&1.115&0.827&0.811&0.807&0.822&0.881\\
\midrule
\multirow{3}{*}{\rotatebox{90}{Quarterly}}
&SMAPE&10.358&10.482 &10.445 &10.218&\secondres{10.184}&\boldres{10.114}&10.465&10.757&12.145&10.380&13.376&11.364&11.100&10.958&11.338&11.360\\
&MASE&1.230&1.258&1.240 &1.203 &\secondres{1.199}&\boldres{1.186}&1.227&1.283&1.520&1.233&1.906&1.328&1.350&1.325&1.365&1.401\\
&OWA&0.919&0.934&0.926 &0.903&\secondres{0.900}&\boldres{0.892}&0.923&0.956&1.106&0.921&1.302&1.000&0.996&0.981&1.012&1.027\\
\midrule
\multirow{3}{*}{\rotatebox{90}{Monthly}}
&SMAPE&13.007&13.184 &13.221 &\secondres{12.788}&\boldres{12.771}&12.862&13.513&13.796&13.514&12.959&14.588&14.014&14.403&13.917&13.958&14.062\\
&MASE&0.964&1.004 &1.003 &\secondres{0.942}&\boldres{0.942}&0.958&1.039&1.083&1.037&0.970&1.368&1.053&1.147&1.097&1.103&1.141\\
&OWA&0.904&0.929 &0.930 &\secondres{0.886}&\boldres{0.886}&0.896&0.957&0.987&0.956&0.905&1.149&0.981&1.038&0.998&1.002&1.024\\
\midrule
\multirow{3}{*}{\rotatebox{90}{Others}}
&SMAPE
&\boldres{4.848}&4.953 &5.194 &\secondres{4.945}&5.045&5.290&6.913&5.569&6.709&4.952&7.267&15.880&7.148&6.302&5.485&24.460\\
&MASE
&\boldres{3.254}&3.271 &3.326 &\secondres{3.257}&3.423&3.489&4.507&3.940&4.953&3.347&5.240&11.434&4.041&4.064&3.865&20.960\\
&OWA
&\boldres{1.023}&1.037 &1.071 &\secondres{1.034}&1.071&1.107&1.438&1.207&1.487&1.049&1.591&3.474&1.389&1.304&1.187&5.879\\
\midrule
\multirow{3}{*}{\rotatebox{90}{Average}}
&SMAPE
&12.050&12.205 & 12.222&\boldres{11.894}&\secondres{11.905}&11.947&12.880&12.684&13.639&12.059&14.718&13.525&13.160&12.780&12.909&14.086\\
&MASE
&\secondres{1.611}&1.642 &1.643 &\boldres{1.592}&1.611&1.614&1.836&1.764&2.095&1.623&2.408&2.111&1.775&1.756&1.771&2.718\\
&OWA
&0.866&0.879 &0.880 &\boldres{0.855}&\secondres{0.860}&0.862&0.955&0.929&1.051&0.869&1.172&1.051&0.949&0.930&0.939&1.230\\

\bottomrule
\end{tabular}
}
\end{small}
\end{center}
\caption{Full short-term time series forecasting results. The forecasting horizons are in [6, 48] and the last three rows are weighted averaged from all datasets under different sampling intervals. We use the same protocol in Table~\ref{tab:long-term-forecasting}.}
\label{tab:short-term-forecasting-full}
\end{table}

\begin{table}[h!]
\centering
\scriptsize
\begin{tabular}{c|c|c|c|c|c|c|c|c|c}
\hline
\multicolumn{2}{c|}{\textbf{Methods}} &\textbf{Metrics} & \textbf{ETTh1} & \textbf{ETTh2} & \textbf{ETTm1} & \textbf{ETTm2} & \textbf{Weather} & \textbf{ECL} & \textbf{Traffic} \\
\hline
\multirow[c]{24}{*}{\rotatebox{90}{\textbf{Teacher}}} 
& \multirow{4}{*}{EfficientNet-B3} & Vision Encoder Param.(M) & \multicolumn{7}{c}{12M} \\
\cline{3-10}
& & Total Param.(M) & \multicolumn{7}{c}{13.3M} \\
\cline{3-10}
& & Mem. (MiB) & 2370 & 2370 & 2370 & 2370 & 2412 & 7800 & 18188 \\
\cline{3-10}
& & Speed (s/iter) & 0.039 & 0.041 & 0.04 & 0.04 & 0.042 & 0.148 & 0.334 \\
\cline{2-10}
& \multirow{4}{*}{ResNet101} & Vision Encoder Param.(M) & \multicolumn{7}{c}{44.55M} \\
\cline{3-10}
& & Total Param.(M) & \multicolumn{7}{c}{45.3M} \\
\cline{3-10}
& & Mem. (MiB) & 2146 & 2146 & 2146 & 2146 & 2156 & 7916 & 18308 \\
\cline{3-10}
& & Speed (s/iter) & 0.031 & 0.032 & 0.031 & 0.03 & 0.034 & 0.141 & 0.333 \\
\cline{2-10}
& \multirow{4}{*}{CLIP} & Vision Encoder Param.(M) & \multicolumn{7}{c}{87.8M} \\
\cline{3-10}
& & Total Param.(M) & \multicolumn{7}{c}{153.6M} \\
\cline{3-10}
& & Mem. (MiB) & 2330 & 2330 & 2330 & 2330 & 2662 & 8344 & 18732 \\
\cline{3-10}
& & Speed (s/iter) & 0.087 & 0.086 & 0.088 & 0.088 & 0.089 & 0.194 & 0.374 \\
\cline{2-10}
& \multirow{4}{*}{MAE-Base} & Vision Encoder Param.(M) & \multicolumn{7}{c}{86M} \\
\cline{3-10}
& & Total Param.(M) & \multicolumn{7}{c}{88.2M} \\
\cline{3-10}
& & Mem. (MiB) & 2088 & 2088 & 2088 & 2088 & 2404 & 8098 & 18484 \\
\cline{3-10}
& & Speed (s/iter) & 0.08 & 0.077 & 0.081 & 0.073 & 0.076 & 0.182 & 0.358 \\
\cline{2-10}

& \multirow{4}{*}{MAE-Large} & Vision Encoder Param.(M) & \multicolumn{7}{c}{304.4M} \\
\cline{3-10}
& & Total Param.(M) & \multicolumn{7}{c}{305.8M} \\
\cline{3-10}
& & Mem. (MiB) & 2972 & 2972 & 2972 & 2972 & 3332 & 8886 & 19276 \\
\cline{3-10}
& & Speed (s/iter) & 0.086 & 0.087 & 0.089 & 0.089 & 0.092 & 0.195 & 0.375 \\
\cline{2-10}
& \multirow{4}{*}{MAE-Huge} & Vision Encoder Param.(M) & \multicolumn{7}{c}{632M} \\
\cline{3-10}
& & Total Param.(M) & \multicolumn{7}{c}{633.3M} \\
\cline{3-10}
& & Mem. (MiB) & 4610 & 4610 & 4610 & 4610 & 4728 & 10262 & 21840 \\
\cline{3-10}
& & Speed (s/iter) & 0.117 & 0.115 & 0.119 & 0.113 & 0.117 & 0.227 & 0.402 \\
\hline
\multirow[c]{12}{*}{\rotatebox{90}{\textbf{Student}}} 
& \multirow{4}{*}{Tiny-ViT} & Vision Encoder Param.(M) & \multicolumn{7}{c}{0.65M} \\
\cline{3-10}
& & Total Param.(M) & \multicolumn{7}{c}{2.87M} \\
\cline{3-10}
& & Mem. (MiB) & 1714 & 1714 & 1714 & 1714 & 2012 & 7736 & 18114 \\
\cline{3-10}
& & Speed (s/iter) & 0.077 & 0.075 & 0.068 & 0.07 & 0.072 & 0.176 & 0.365 \\
\cline{2-10}
& \multirow{4}{*}{EfficientNet-B0} & Vision Encoder Param.(M) & \multicolumn{7}{c}{5.3M} \\
\cline{3-10}
& & Total Param.(M) & \multicolumn{7}{c}{6.37M} \\
\cline{3-10}
& & Mem. (MiB) & 1960 & 1960 & 1960 & 1960 & 1978 & 7772 & 18162 \\
\cline{3-10}
& & Speed (s/iter) & 0.02 & 0.021 & 0.023 & 0.022 & 0.025 & 0.13 & 0.319 \\
\cline{2-10}
& \multirow{4}{*}{MobileNet} & Vision Encoder Param.(M) & \multicolumn{7}{c}{2.54M} \\
\cline{3-10}
& & Total Param.(M) & \multicolumn{7}{c}{3.85M} \\
\cline{3-10}
& & Mem. (MiB) & 1888 & 1888 & 1888 & 1888 & 2106 & 7764 & 18152 \\
\cline{3-10}
& & Speed (s/iter) & 0.017 & 0.017 & 0.018 & 0.019 & 0.02 & 0.125 & 0.307 \\
\hline
\end{tabular}
\caption{Computational efficiency comparison between Teacher Model and Student Model across datasets. The forecasting horizon is 96.}
\label{tab:model_comparison}
\end{table}

\begin{table}[h!]
\captionsetup{font=small}
\begin{center}
\begin{small}
\scalebox{0.5}{
\setlength\tabcolsep{3pt}
\begin{tabular}{c|c|cc|cc|cc|cc|cc|cc|cc|cc|cc|cc|cc|cc|cc|cc|cc|cc|cc|cc}
\toprule

\multicolumn{2}{c|}{Teacher Model}&\multicolumn{6}{c|}{EfficientNet-B3}&\multicolumn{6}{c|}{ResNet101}&\multicolumn{6}{c|}{CLIP}&\multicolumn{6}{c|}{MAE-Base}&\multicolumn{6}{c|}{MAE-Large}&\multicolumn{6}{c}{MAE-Huge} \\

\midrule

\multicolumn{2}{c|}{Student Model} & \multicolumn{2}{c|}{Tiny-ViT} & \multicolumn{2}{c|}{EfficientNet} & \multicolumn{2}{c|}{MoblieNet} & \multicolumn{2}{c|}{Tiny-ViT} & \multicolumn{2}{c|}{EfficientNet} & \multicolumn{2}{c|}{MoblieNet} & \multicolumn{2}{c|}{Tiny-ViT} & \multicolumn{2}{c|}{EfficientNet} & \multicolumn{2}{c|}{MoblieNet} & \multicolumn{2}{c|}{Tiny-ViT} & \multicolumn{2}{c|}{EfficientNet} & \multicolumn{2}{c|}{MoblieNet} & \multicolumn{2}{c|}{Tiny-ViT} & \multicolumn{2}{c|}{EfficientNet} & \multicolumn{2}{c|}{MoblieNet} & \multicolumn{2}{c|}{Tiny-ViT} & \multicolumn{2}{c|}{EfficientNet} & \multicolumn{2}{c}{MoblieNet} \\

\midrule

\multicolumn{2}{c|}{Metric} & MSE  & MAE & MSE & MAE & MSE & MAE & MSE  & MAE & MSE  & MAE & MSE  & MAE & MSE  & MAE & MSE  & MAE & MSE  & MAE & MSE  & MAE & MSE  & MAE & MSE  & MAE & MSE  & MAE & MSE  & MAE & MSE  & MAE & MSE  & MAE & MSE  & MAE & MSE  & MAE\\
\midrule

\multirow{5}{*}{\rotatebox{90}{$ETTh1$}}
& 96  & 0.361 & 0.389 & 0.364 & 0.393 & 0.360 & 0.388 & 0.361 & 0.389 & 0.361 & 0.389 & 0.360 & 0.388 & 0.362 & 0.389 & 0.360 & 0.388 & 0.360 & 0.389 & 0.363 & 0.391 & 0.360 & 0.388 & 0.360 & 0.388 & 0.361 & 0.389 & 0.361 & 0.389 & 0.368 & 0.395 & 0.361 & 0.388 & 0.361 & 0.389 & 0.360 & 0.388 \\
& 192 & 0.397 & 0.412 & 0.396 & 0.411 & 0.397 & 0.409 & 0.430 & 0.439 & 0.397 & 0.411 & 0.405 & 0.421 & 0.396 & 0.410 & 0.396 & 0.410 & 0.397 & 0.412 & 0.396 & 0.411 & 0.397 & 0.411 & 0.396 & 0.411 & 0.398 & 0.412 & 0.396 & 0.411 & 0.400 & 0.414 & 0.397 & 0.411 & 0.396 & 0.411 & 0.396 & 0.411 \\
& 336 & 0.422 & 0.428 & 0.423 & 0.430 & 0.416 & 0.424 & 0.421 & 0.427 & 0.421 & 0.426 & 0.421 & 0.427 & 0.419 & 0.424 & 0.424 & 0.429 & 0.423 & 0.427 & 0.422 & 0.428 & 0.420 & 0.426 & 0.423 & 0.428 & 0.419 & 0.424 & 0.422 & 0.427 & 0.421 & 0.427 & 0.424 & 0.429 & 0.424 & 0.429 & 0.417 & 0.424 \\
& 720 & 0.457 & 0.475 & 0.522 & 0.510 & 0.513 & 0.508 & 0.462 & 0.477 & 0.557 & 0.529 & 0.559 & 0.535 & 0.498 & 0.499 & 0.451 & 0.469 & 0.473 & 0.485 & 0.452 & 0.470 & 0.567 & 0.535 & 0.530 & 0.515 & 0.440 & 0.462 & 0.514 & 0.507 & 0.518 & 0.511 & 0.443 & 0.464 & 0.494 & 0.496 & 0.572 & 0.539 \\
& Avg & 0.409 & 0.426 & 0.426 & 0.436 & 0.421 & 0.433 & 0.419 & 0.433 & 0.434 & 0.439 & 0.436 & 0.443 & 0.419 & 0.431 & 0.408 & 0.424 & 0.413 & 0.428 & 0.408 & 0.425 & 0.436 & 0.440 & 0.427 & 0.435 & 0.405 & 0.422 & 0.423 & 0.434 & 0.427 & 0.437 & 0.406 & 0.423 & 0.419 & 0.431 & 0.436 & 0.441 \\
\midrule

\multirow{5}{*}{\rotatebox{90}{$ETTh2$}}
& 96  & 0.266 & 0.331 & 0.282 & 0.340 & 0.277 & 0.340 & 0.270 & 0.338 & 0.266 & 0.331 & 0.266 & 0.333 & 0.267 & 0.332 & 0.267 & 0.332 & 0.266 & 0.331 & 0.267 & 0.332 & 0.266 & 0.331 & 0.280 & 0.342 & 0.272 & 0.338 & 0.266 & 0.331 & 0.276 & 0.340 & 0.266 & 0.331 & 0.274 & 0.336 & 0.291 & 0.345 \\
& 192 & 0.338 & 0.378 & 0.348 & 0.382 & 0.344 & 0.377 & 0.348 & 0.393 & 0.331 & 0.377 & 0.329 & 0.377 & 0.330 & 0.372 & 0.330 & 0.372 & 0.330 & 0.374 & 0.330 & 0.372 & 0.337 & 0.379 & 0.331 & 0.373 & 0.330 & 0.372 & 0.339 & 0.378 & 0.341 & 0.377 & 0.332 & 0.376 & 0.341 & 0.385 & 0.353 & 0.385 \\
& 336 & 0.363 & 0.401 & 0.375 & 0.409 & 0.364 & 0.399 & 0.362 & 0.405 & 0.364 & 0.402 & 0.407 & 0.442 & 0.361 & 0.402 & 0.362 & 0.403 & 0.359 & 0.399 & 0.363 & 0.406 & 0.375 & 0.407 & 0.362 & 0.401 & 0.360 & 0.400 & 0.377 & 0.407 & 0.370 & 0.404 & 0.365 & 0.401 & 0.367 & 0.401 & 0.366 & 0.400 \\
& 720 & 0.393 & 0.432 & 0.472 & 0.472 & 0.452 & 0.471 & 0.405 & 0.443 & 0.437 & 0.464 & 0.403 & 0.440 & 0.405 & 0.445 & 0.412 & 0.444 & 0.402 & 0.442 & 0.390 & 0.428 & 0.426 & 0.451 & 0.416 & 0.443 & 0.410 & 0.448 & 0.413 & 0.442 & 0.420 & 0.446 & 0.394 & 0.431 & 0.421 & 0.452 & 0.430 & 0.453 \\
& Avg & 0.340 & 0.386 & 0.369 & 0.401 & 0.359 & 0.397 & 0.346 & 0.395 & 0.349 & 0.394 & 0.351 & 0.398 & 0.341 & 0.388 & 0.343 & 0.388 & 0.340 & 0.386 & 0.338 & 0.384 & 0.351 & 0.392 & 0.347 & 0.390 & 0.343 & 0.389 & 0.349 & 0.389 & 0.351 & 0.392 & 0.339 & 0.385 & 0.351 & 0.393 & 0.360 & 0.396 \\
\midrule

\multirow{5}{*}{\rotatebox{90}{$ETTm1$}}
& 96  & 0.301 & 0.343 & 0.304 & 0.347 & 0.304 & 0.348 & 0.306 & 0.348 & 0.302 & 0.344 & 0.299 & 0.341 & 0.300 & 0.342 & 0.299 & 0.341 & 0.299 & 0.341 & 0.300 & 0.342 & 0.300 & 0.343 & 0.303 & 0.348 & 0.300 & 0.342 & 0.301 & 0.344 & 0.301 & 0.344 & 0.300 & 0.341 & 0.301 & 0.343 & 0.303 & 0.348 \\
& 192 & 0.333 & 0.363 & 0.338 & 0.369 & 0.333 & 0.363 & 0.592 & 0.518 & 0.330 & 0.360 & 0.333 & 0.363 & 0.330 & 0.360 & 0.330 & 0.359 & 0.333 & 0.363 & 0.330 & 0.360 & 0.333 & 0.364 & 0.552 & 0.502 & 0.330 & 0.360 & 0.335 & 0.366 & 0.337 & 0.370 & 0.331 & 0.360 & 0.333 & 0.364 & 0.333 & 0.363 \\
& 336 & 0.360 & 0.378 & 0.363 & 0.381 & 0.363 & 0.380 & 0.361 & 0.379 & 0.363 & 0.380 & 0.408 & 0.418 & 0.364 & 0.381 & 0.363 & 0.380 & 0.360 & 0.378 & 0.361 & 0.378 & 0.363 & 0.381 & 0.364 & 0.382 & 0.363 & 0.381 & 0.363 & 0.381 & 0.364 & 0.382 & 0.361 & 0.378 & 0.368 & 0.387 & 0.363 & 0.381 \\
& 720 & 0.418 & 0.412 & 0.445 & 0.432 & 0.445 & 0.432 & 0.417 & 0.410 & 0.449 & 0.435 & 0.428 & 0.423 & 0.415 & 0.409 & 0.416 & 0.410 & 0.418 & 0.412 & 0.416 & 0.410 & 0.444 & 0.433 & 0.447 & 0.434 & 0.416 & 0.410 & 0.435 & 0.426 & 0.435 & 0.426 & 0.416 & 0.410 & 0.444 & 0.432 & 0.436 & 0.426 \\
& Avg & 0.353 & 0.374 & 0.362 & 0.382 & 0.361 & 0.381 & 0.419 & 0.414 & 0.361 & 0.380 & 0.367 & 0.386 & 0.352 & 0.373 & 0.352 & 0.373 & 0.353 & 0.373 & 0.352 & 0.372 & 0.360 & 0.380 & 0.417 & 0.416 & 0.352 & 0.373 & 0.359 & 0.379 & 0.359 & 0.381 & 0.352 & 0.372 & 0.362 & 0.382 & 0.359 & 0.379 \\
\midrule

\multirow{5}{*}{\rotatebox{90}{$ETTm2$}}
& 96  & 0.161 & 0.248 & 0.163 & 0.251 & 0.163 & 0.252 & 0.204 & 0.294 & 0.163 & 0.251 & 0.162 & 0.250 & 0.161 & 0.248 & 0.161 & 0.249 & 0.161 & 0.248 & 0.161 & 0.248 & 0.161 & 0.249 & 0.161 & 0.249 & 0.161 & 0.248 & 0.163 & 0.251 & 0.161 & 0.248 & 0.161 & 0.248 & 0.161 & 0.249 & 0.162 & 0.249 \\
& 192 & 0.223 & 0.293 & 0.222 & 0.292 & 0.222 & 0.294 & 0.230 & 0.303 & 0.224 & 0.296 & 0.217 & 0.288 & 0.218 & 0.288 & 0.217 & 0.287 & 0.217 & 0.287 & 0.218 & 0.288 & 0.219 & 0.291 & 0.225 & 0.298 & 0.217 & 0.288 & 0.219 & 0.293 & 0.230 & 0.302 & 0.217 & 0.287 & 0.216 & 0.289 & 0.216 & 0.286 \\
& 336 & 0.271 & 0.324 & 0.279 & 0.333 & 0.280 & 0.330 & 0.269 & 0.321 & 0.271 & 0.322 & 0.270 & 0.324 & 0.269 & 0.321 & 0.270 & 0.322 & 0.269 & 0.321 & 0.271 & 0.322 & 0.269 & 0.321 & 0.270 & 0.322 & 0.269 & 0.322 & 0.286 & 0.339 & 0.269 & 0.322 & 0.269 & 0.321 & 0.285 & 0.336 & 0.275 & 0.328 \\
& 720 & 0.361 & 0.383 & 0.362 & 0.383 & 0.355 & 0.385 & 0.354 & 0.375 & 0.341 & 0.374 & 0.354 & 0.374 & 0.348 & 0.375 & 0.348 & 0.376 & 0.357 & 0.376 & 0.355 & 0.377 & 0.341 & 0.375 & 0.352 & 0.376 & 0.346 & 0.375 & 0.396 & 0.409 & 0.350 & 0.376 & 0.358 & 0.379 & 0.347 & 0.375 & 0.351 & 0.376 \\
& Avg & 0.254 & 0.312 & 0.256 & 0.315 & 0.255 & 0.315 & 0.264 & 0.323 & 0.250 & 0.311 & 0.251 & 0.309 & 0.249 & 0.308 & 0.249 & 0.309 & 0.251 & 0.308 & 0.251 & 0.309 & 0.247 & 0.309 & 0.252 & 0.311 & 0.248 & 0.308 & 0.266 & 0.323 & 0.253 & 0.312 & 0.251 & 0.309 & 0.252 & 0.312 & 0.251 & 0.310 \\
\midrule

\multirow{5}{*}{\rotatebox{90}{$Weather$}}
& 96  & 0.148 & 0.194 & 0.148 & 0.194 & 0.149 & 0.194 & 0.148 & 0.194 & 0.147 & 0.194 & 0.148 & 0.193 & 0.148 & 0.194 & 0.147 & 0.193 & 0.147 & 0.192 & 0.147 & 0.193 & 0.148 & 0.195 & 0.147 & 0.193 & 0.147 & 0.193 & 0.146 & 0.192 & 0.147 & 0.193 & 0.147 & 0.194 & 0.146 & 0.193 & 0.146 & 0.192 \\
& 192 & 0.192 & 0.236 & 0.192 & 0.237 & 0.193 & 0.237 & 0.192 & 0.236 & 0.193 & 0.239 & 0.193 & 0.239 & 0.192 & 0.236 & 0.192 & 0.236 & 0.193 & 0.238 & 0.192 & 0.235 & 0.193 & 0.238 & 0.193 & 0.236 & 0.193 & 0.237 & 0.191 & 0.236 & 0.193 & 0.237 & 0.192 & 0.236 & 0.192 & 0.237 & 0.193 & 0.237 \\
& 336 & 0.246 & 0.281 & 0.252 & 0.285 & 0.258 & 0.288 & 0.247 & 0.279 & 0.253 & 0.286 & 0.259 & 0.289 & 0.245 & 0.279 & 0.247 & 0.279 & 0.244 & 0.278 & 0.246 & 0.278 & 0.244 & 0.277 & 0.246 & 0.281 & 0.246 & 0.279 & 0.247 & 0.279 & 0.245 & 0.278 & 0.245 & 0.278 & 0.247 & 0.278 & 0.244 & 0.276 \\
& 720 & 0.320 & 0.331 & 0.318 & 0.330 & 0.317 & 0.332 & 0.321 & 0.330 & 0.319 & 0.331 & 0.319 & 0.334 & 0.321 & 0.333 & 0.317 & 0.329 & 0.319 & 0.332 & 0.321 & 0.332 & 0.323 & 0.333 & 0.320 & 0.332 & 0.319 & 0.333 & 0.317 & 0.329 & 0.319 & 0.332 & 0.318 & 0.329 & 0.318 & 0.332 & 0.321 & 0.332 \\
& Avg & 0.226 & 0.261 & 0.227 & 0.261 & 0.229 & 0.263 & 0.227 & 0.260 & 0.228 & 0.262 & 0.230 & 0.264 & 0.227 & 0.260 & 0.225 & 0.259 & 0.226 & 0.260 & 0.226 & 0.260 & 0.227 & 0.260 & 0.227 & 0.261 & 0.226 & 0.260 & 0.225 & 0.259 & 0.226 & 0.260 & 0.225 & 0.259 & 0.226 & 0.260 & 0.226 & 0.259 \\
\midrule

\multirow{5}{*}{\rotatebox{90}{$Electricity$}}
& 96  & 0.136 & 0.236 & 0.136 & 0.235 & 0.136 & 0.235 & 0.136 & 0.236 & 0.136 & 0.235 & 0.136 & 0.235 & 0.136 & 0.236 & 0.136 & 0.236 & 0.137 & 0.236 & 0.137 & 0.236 & 0.137 & 0.236 & 0.138 & 0.238 & 0.136 & 0.236 & 0.136 & 0.235 & 0.137 & 0.237 & 0.136 & 0.236 & 0.136 & 0.236 & 0.138 & 0.237 \\
& 192 & 0.152 & 0.249 & 0.151 & 0.249 & 0.152 & 0.250 & 0.152 & 0.249 & 0.152 & 0.249 & 0.152 & 0.251 & 0.152 & 0.249 & 0.152 & 0.250 & 0.153 & 0.252 & 0.152 & 0.249 & 0.152 & 0.249 & 0.153 & 0.252 & 0.152 & 0.249 & 0.153 & 0.251 & 0.154 & 0.253 & 0.152 & 0.248 & 0.152 & 0.249 & 0.153 & 0.251 \\
& 336 & 0.167 & 0.264 & 0.167 & 0.264 & 0.167 & 0.264 & 0.167 & 0.264 & 0.166 & 0.264 & 0.167 & 0.264 & 0.167 & 0.264 & 0.167 & 0.264 & 0.167 & 0.263 & 0.167 & 0.264 & 0.167 & 0.264 & 0.167 & 0.264 & 0.167 & 0.264 & 0.168 & 0.264 & 0.167 & 0.264 & 0.167 & 0.264 & 0.167 & 0.264 & 0.167 & 0.263 \\
& 720 & 0.208 & 0.298 & 0.202 & 0.293 & 0.197 & 0.291 & 0.210 & 0.302 & 0.203 & 0.293 & 0.196 & 0.290 & 0.207 & 0.298 & 0.207 & 0.297 & 0.207 & 0.297 & 0.207 & 0.298 & 0.207 & 0.297 & 0.207 & 0.298 & 0.207 & 0.297 & 0.207 & 0.298 & 0.207 & 0.297 & 0.207 & 0.298 & 0.208 & 0.298 & 0.207 & 0.298 \\
& Avg & 0.166 & 0.262 & 0.164 & 0.260 & 0.163 & 0.260 & 0.166 & 0.263 & 0.164 & 0.260 & 0.163 & 0.260 & 0.165 & 0.262 & 0.166 & 0.262 & 0.166 & 0.262 & 0.166 & 0.262 & 0.166 & 0.262 & 0.166 & 0.263 & 0.166 & 0.261 & 0.166 & 0.262 & 0.166 & 0.263 & 0.165 & 0.261 & 0.166 & 0.262 & 0.166 & 0.262 \\
\midrule

\multirow{5}{*}{\rotatebox{90}{$Traffic$}}
& 96  & 0.387 & 0.273 & 0.378 & 0.269 & 0.379 & 0.268 & 0.387 & 0.272 & 0.379 & 0.269 & 0.411 & 0.298 & 0.387 & 0.273 & 0.384 & 0.271 & 0.378 & 0.268 & 0.387 & 0.273 & 0.378 & 0.269 & 0.378 & 0.268 & 0.388 & 0.273 & 0.378 & 0.269 & 0.377 & 0.268 & 0.387 & 0.273 & 0.382 & 0.271 & 0.378 & 0.268 \\
& 192 & 0.402 & 0.278 & 0.396 & 0.276 & 0.402 & 0.278 & 0.402 & 0.278 & 0.397 & 0.276 & 0.402 & 0.278 & 0.402 & 0.278 & 0.400 & 0.277 & 0.402 & 0.278 & 0.402 & 0.278 & 0.401 & 0.278 & 0.408 & 0.288 & 0.402 & 0.278 & 0.396 & 0.276 & 0.402 & 0.278 & 0.402 & 0.278 & 0.397 & 0.276 & 0.402 & 0.278 \\
& 336 & 0.411 & 0.281 & 0.410 & 0.281 & 0.409 & 0.281 & 0.411 & 0.282 & 0.412 & 0.282 & 0.411 & 0.282 & 0.411 & 0.282 & 0.411 & 0.282 & 0.412 & 0.282 & 0.411 & 0.282 & 0.411 & 0.282 & 0.409 & 0.281 & 0.411 & 0.282 & 0.411 & 0.282 & 0.411 & 0.282 & 0.411 & 0.281 & 0.412 & 0.282 & 0.409 & 0.281 \\
& 720 & 0.447 & 0.300 & 0.447 & 0.300 & 0.445 & 0.299 & 0.556 & 0.412 & 0.447 & 0.300 & 0.447 & 0.300 & 0.448 & 0.300 & 0.447 & 0.300 & 0.447 & 0.301 & 0.448 & 0.301 & 0.447 & 0.300 & 0.448 & 0.301 & 0.447 & 0.300 & 0.447 & 0.300 & 0.448 & 0.301 & 0.448 & 0.301 & 0.447 & 0.300 & 0.447 & 0.301 \\
& Avg & 0.412 & 0.283 & 0.408 & 0.281 & 0.409 & 0.281 & 0.439 & 0.311 & 0.409 & 0.282 & 0.418 & 0.289 & 0.412 & 0.283 & 0.411 & 0.283 & 0.410 & 0.282 & 0.412 & 0.283 & 0.410 & 0.282 & 0.411 & 0.284 & 0.412 & 0.283 & 0.408 & 0.282 & 0.409 & 0.282 & 0.412 & 0.283 & 0.409 & 0.282 & 0.409 & 0.282 \\

\bottomrule
\end{tabular}
}
\end{small}
\end{center}
\caption{Performance comparison of different Teacher-Student model combinations on long-term forecasting tasks.}
\label{tab:teacher-student-forecasting-full-result}
\end{table}
\section{Teacher Student Long Term Forecasting and Efficiency}
\label{appx:teacher-student-efficiency}
Table~\ref{tab:model_comparison} demonstrates the computational efficiency gains achieved through our knowledge distillation approach. Parameter reduction varies significantly depending on the teacher-student pairing. When comparing the smallest teacher model (EfficientNet-B3, 13.3M total parameters) with the largest student model (EfficientNet-B0, 6.37M), the reduction is approximately 52\%. However, more aggressive compressions are achieved with other pairings: the Tiny-ViT student model (2.87M) represents a 99.5\% reduction compared to MAE-Huge (633.3M), and even a 78\% reduction compared to the smallest teacher. For vision encoder parameters specifically, the reduction ranges from 56\% (12M to 5.3M) to 99.9\% (632M to 0.65M).

The asymmetry between parameter reduction and runtime efficiency is particularly revealing. Memory consumption decreases by only 17-60\% across datasets, while inference speed improves by 1.5-3.4×. This discrepancy stems from our targeted distillation approach: we only compress the vision component while preserving the original temporal forecasting module. The temporal module, which processes sequential data through transformer layers and prediction heads, dominates runtime resource consumption. For instance, on ETTh1, even Tiny-ViT with its minimal 0.65M vision encoder still requires 1714 MiB memory—only 18\% less than ResNet101 (2146 MiB) despite having 98.5\% fewer vision parameters. This analysis confirms that vision models contain substantial redundancy for time series tasks, while also revealing that the temporal processing pipeline represents the primary computational bottleneck during inference.

Table~\ref{tab:teacher-student-forecasting-full-result} investigates the impact of different teacher-student model combinations on our knowledge distillation framework through comprehensive experiments with fixed hyperparameters ($\lambda_{cd} = \lambda_{fd} = 0.1$
), where the learnable distillation parameter module was disabled to ensure fair comparison across six teacher models and three student architectures. The results reveal several important insights regarding optimal model pairing strategies. MAE-based teacher models (mae-large and mae-huge) consistently demonstrate superior performance across most datasets compared to other vision models, achieving the best results on multiple benchmarks. The CLIP teacher model also shows strong performance, particularly when paired with lightweight student architectures. Among student models, tiny-vit and EfficientNet exhibit the most consistent performance across different teacher-student combinations, while MobileNet demonstrates competitive results with lower computational overhead. The experimental results indicate that even with fixed distillation weights, the choice of teacher model has a more significant impact on final performance than the student architecture, with MAE-based teachers providing more effective knowledge transfer for time series forecasting tasks. Notably, certain teacher-student combinations show dataset-specific advantages, suggesting that the effectiveness of knowledge distillation depends on the compatibility between the teacher's learned representations and the target dataset characteristics. Overall, the results validate that our knowledge distillation framework benefits from carefully selected teacher-student pairs, with MAE-based teachers generally providing the most robust performance improvements even under controlled hyperparameter settings.

\section{Visualization of prediction results}
\label{appx:visualization-of-prediction-results}

\begin{figure}[!hb]
  \centering
  \begin{subfigure}[b]{0.25\textwidth}
    \centering
    \includegraphics[width=\linewidth]{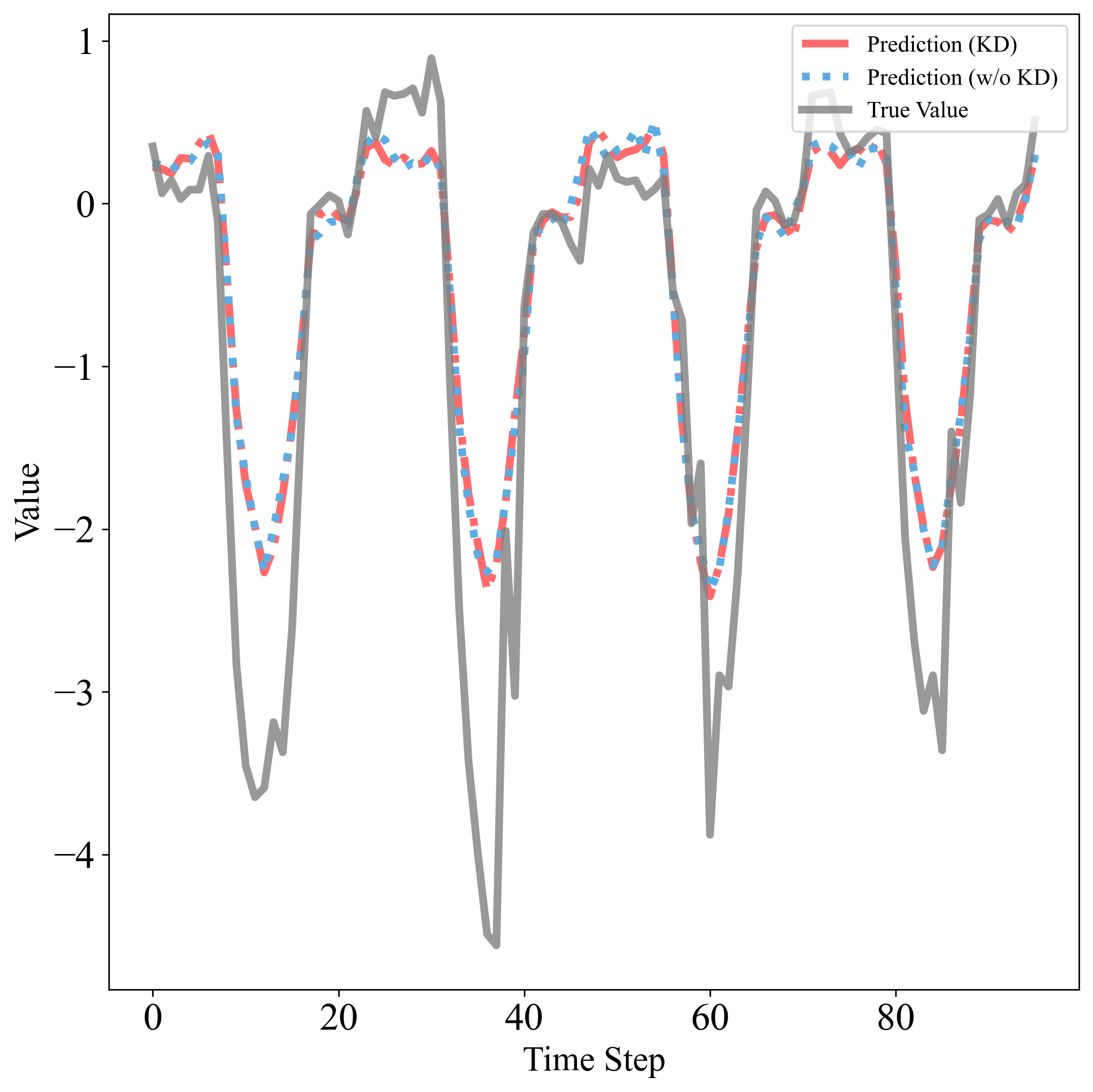}
    \caption{ETTh1}
    \label{fig:etth1_96}
  \end{subfigure}%
  \hfill
  \begin{subfigure}[b]{0.25\textwidth}
    \centering
    \includegraphics[width=\linewidth]{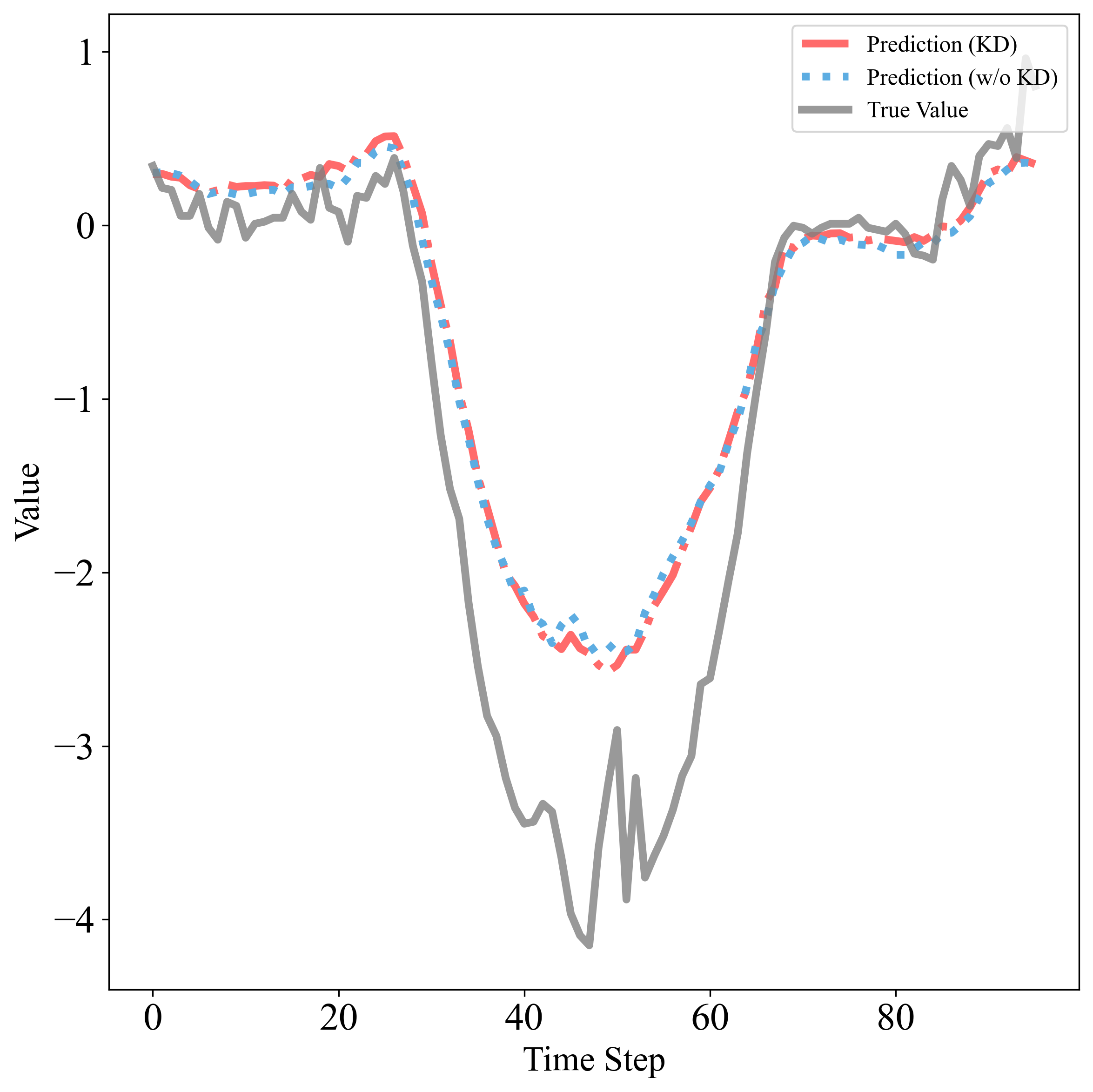}
    \caption{ETTm1}
    \label{fig:ettm1_96}
  \end{subfigure}%
  \hfill
  \begin{subfigure}[b]{0.25\textwidth}
    \centering
    \includegraphics[width=\linewidth]{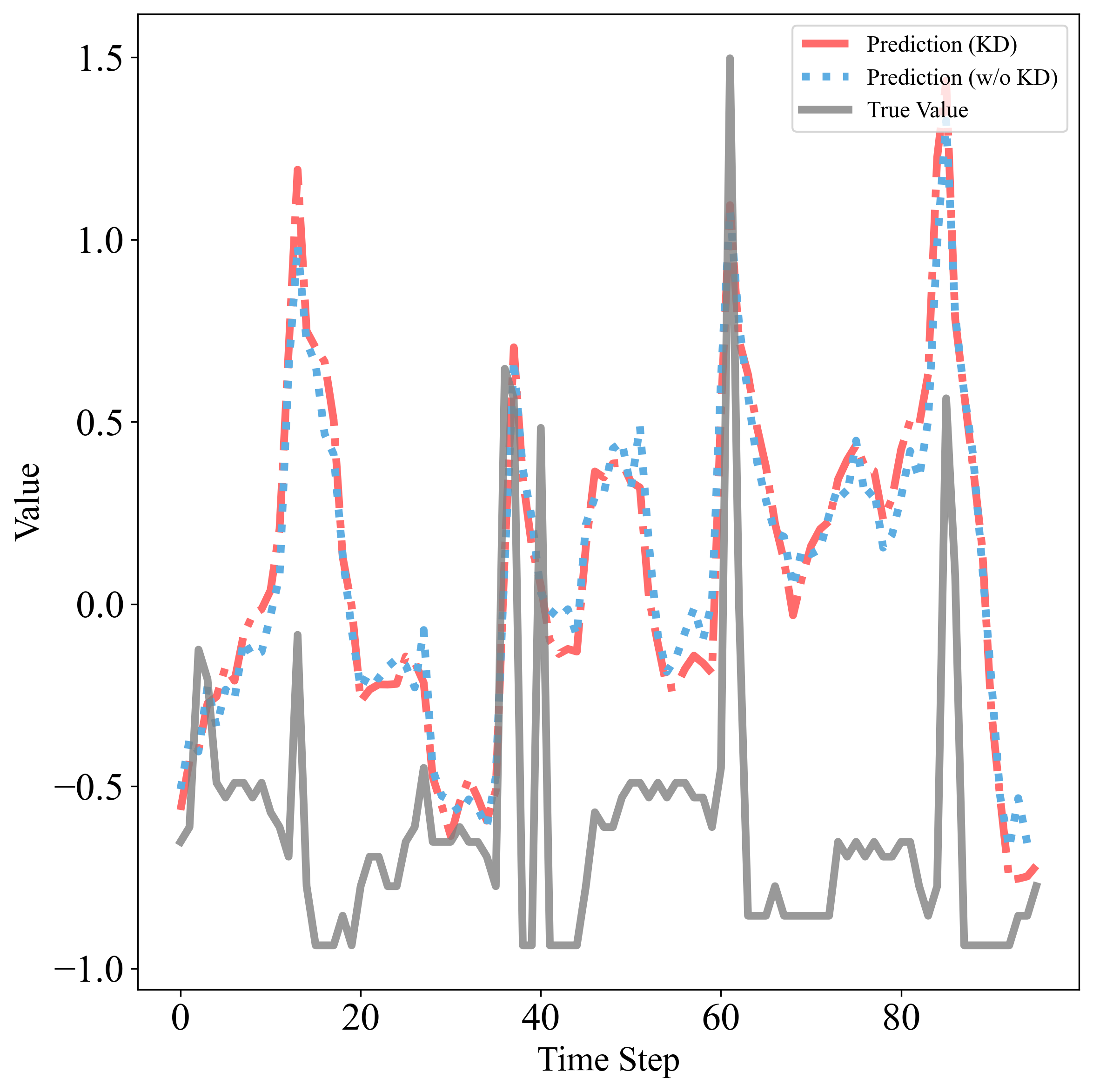}
    \caption{ECL}
    \label{fig:ecl_96}
  \end{subfigure}%
  \hfill
  \begin{subfigure}[b]{0.25\textwidth}
    \centering
    \includegraphics[width=\linewidth]{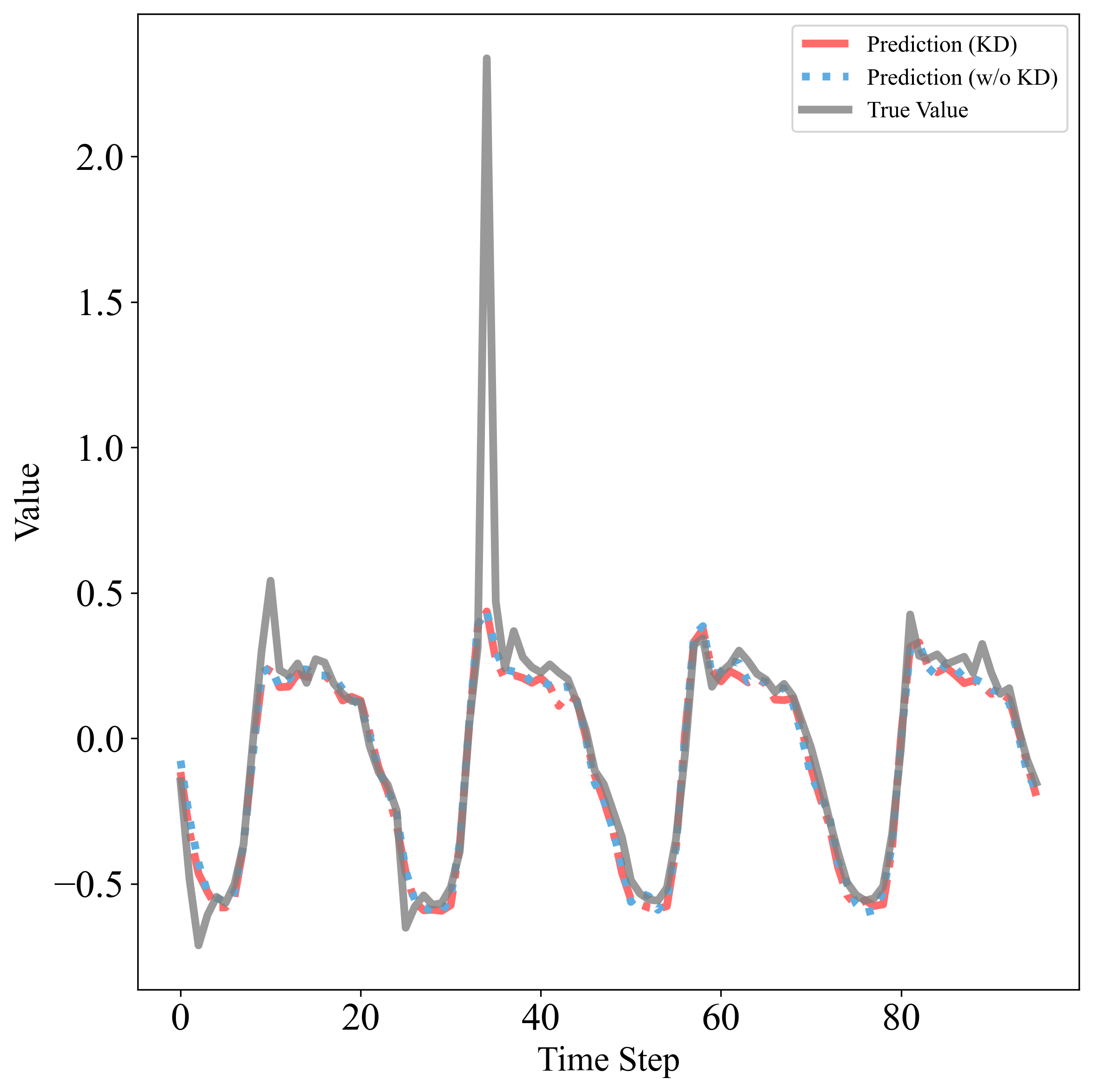}
    \caption{Traffic}
    \label{fig:traffic_96}
  \end{subfigure}
  \caption{Prediction results visualization for ETTh1, ETTm1, ECL, and Traffic datasets at 96 prediction lengths. The solid gray line shows the true values, the dashed red line shows the model’s predictions with knowledge distillation (KD), and the dotted blue line shows the teacher model's predictions without distillation (w/o KD).}
  \label{fig:visualization-96}
\end{figure}

\begin{figure}[!h]
  \centering
  \begin{subfigure}[b]{0.25\textwidth}
    \centering
    \includegraphics[width=\linewidth]{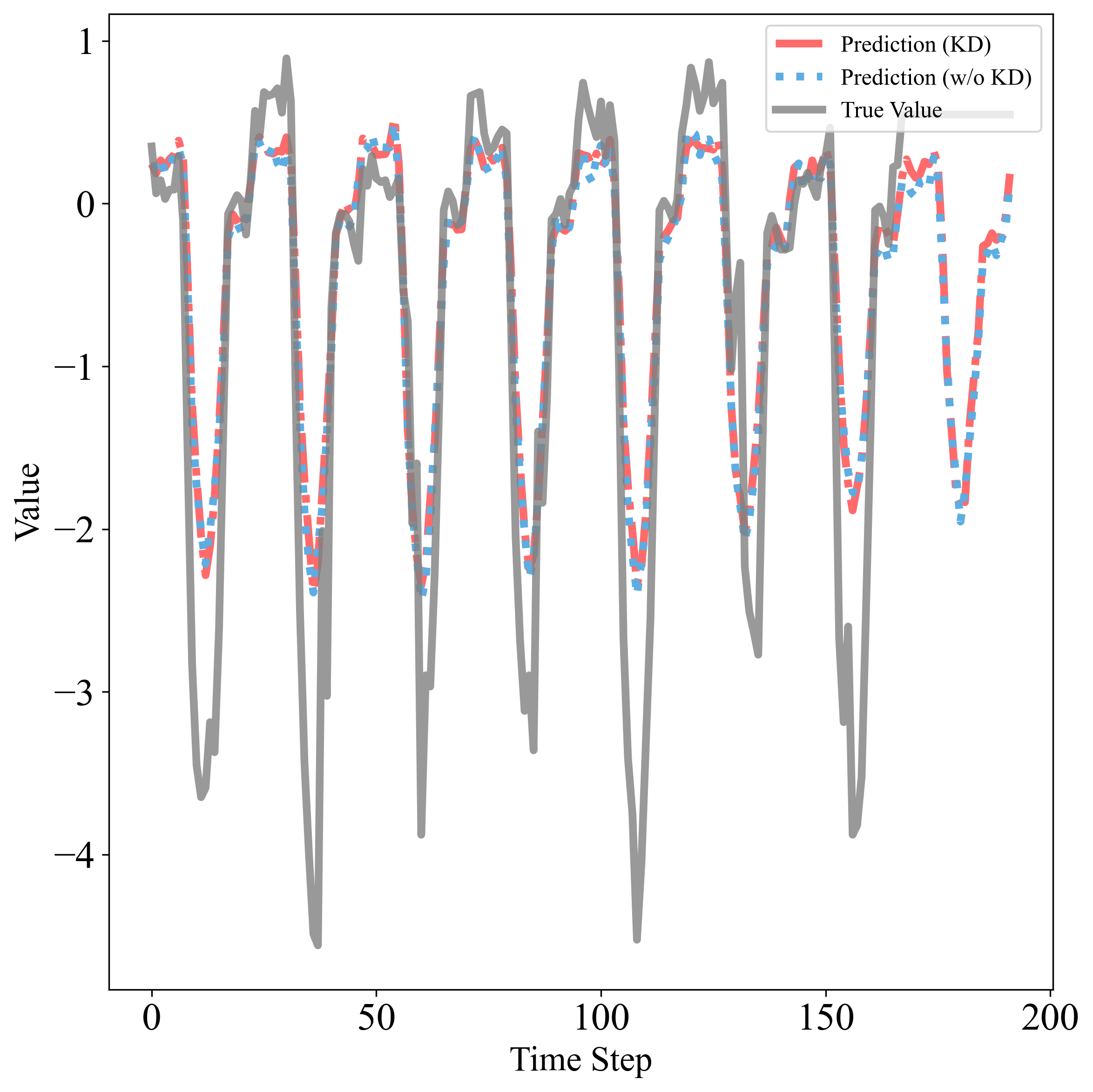}
    \caption{ETTh1}
    \label{fig:etth1_192}
  \end{subfigure}%
  \hfill
  \begin{subfigure}[b]{0.25\textwidth}
    \centering
    \includegraphics[width=\linewidth]{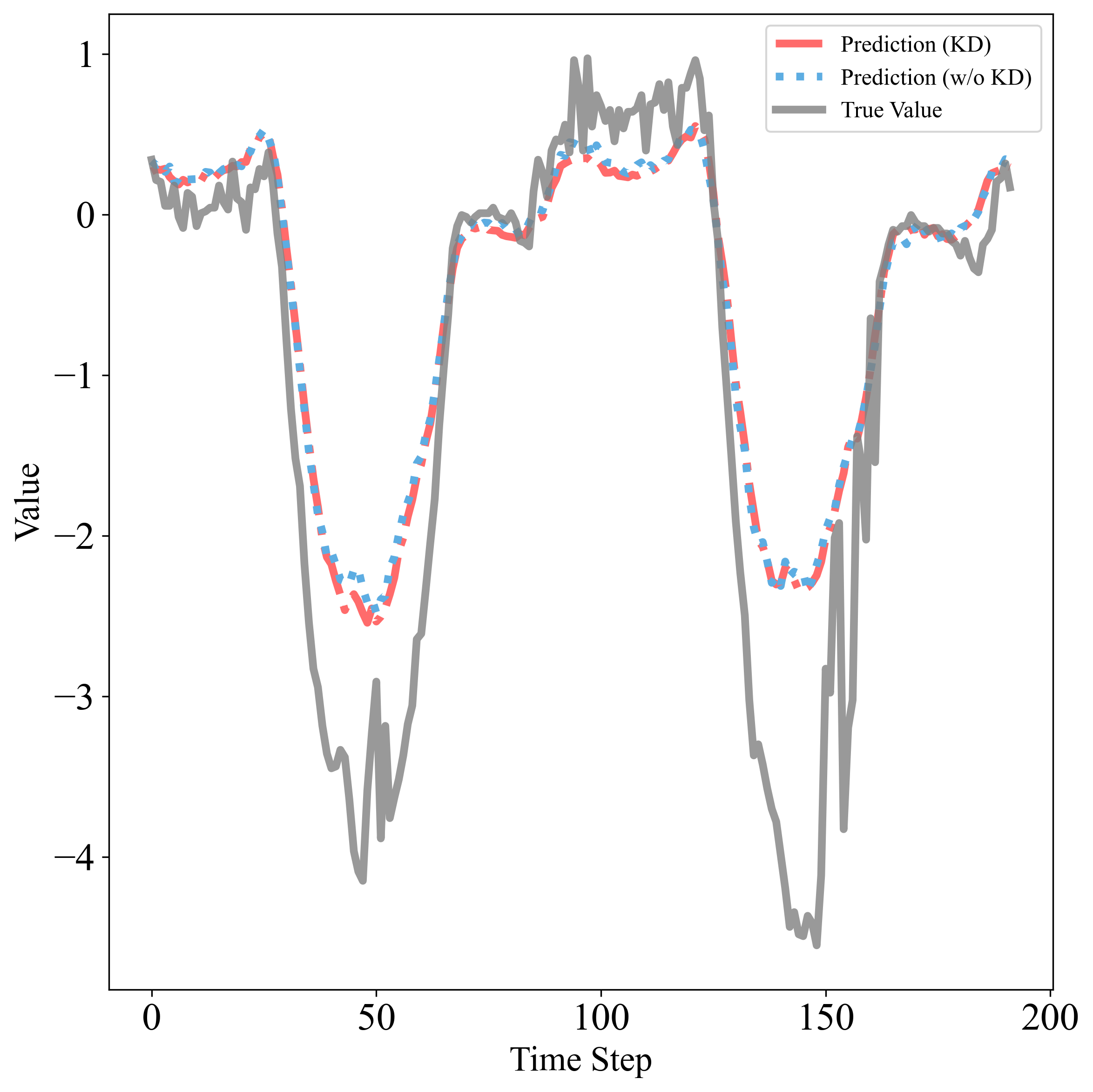}
    \caption{ETTm1}
    \label{fig:ettm1_192}
  \end{subfigure}%
  \hfill
  \begin{subfigure}[b]{0.25\textwidth}
    \centering
    \includegraphics[width=\linewidth]{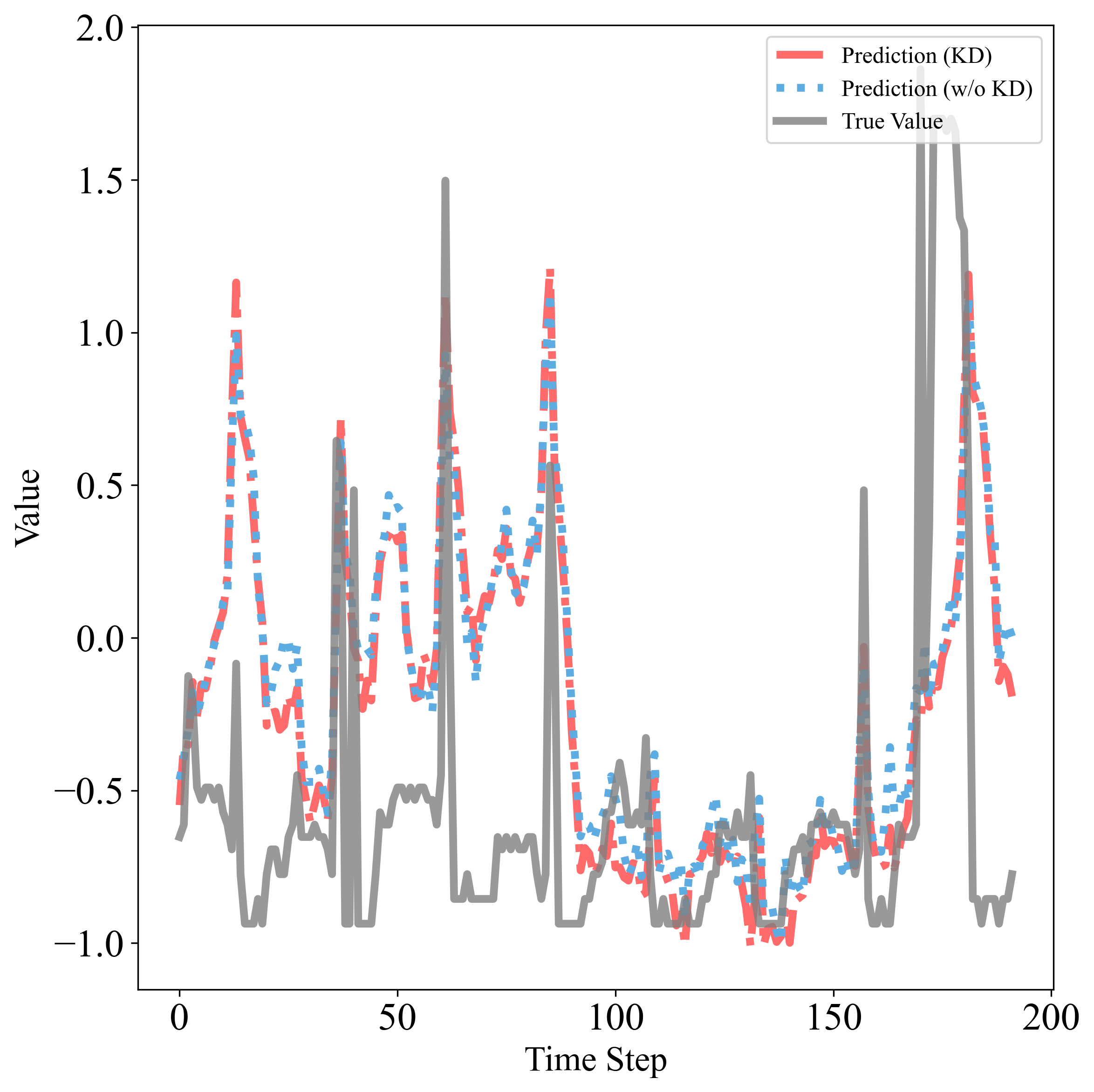}
    \caption{ECL}
    \label{fig:ecl_192}
  \end{subfigure}%
  \hfill
  \begin{subfigure}[b]{0.25\textwidth}
    \centering
    \includegraphics[width=\linewidth]{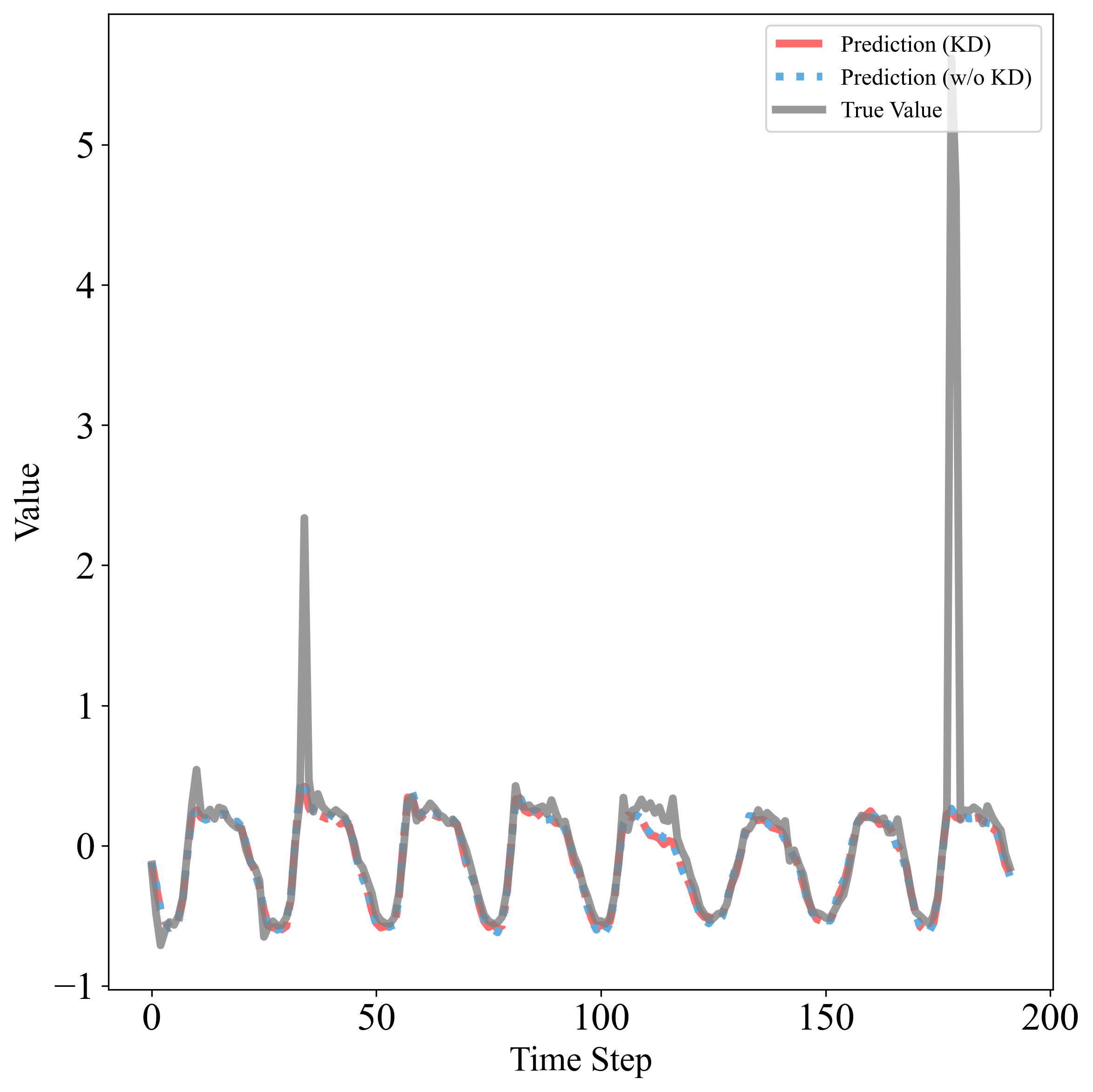}
    \caption{Traffic}
    \label{fig:traffic_192}
  \end{subfigure}
  \caption{Prediction results visualization for ETTh1, ETTm1, ECL, and Traffic datasets at 192 prediction lengths. The solid gray line shows the true values, the dashed red line shows the model’s predictions with knowledge distillation (KD), and the dotted blue line shows the teacher model's predictions without distillation (w/o KD).}
  \label{fig:visualization-192}
\end{figure}

\begin{figure}[!h]
  \centering
  \begin{subfigure}[b]{0.25\textwidth}
    \centering
    \includegraphics[width=\linewidth]{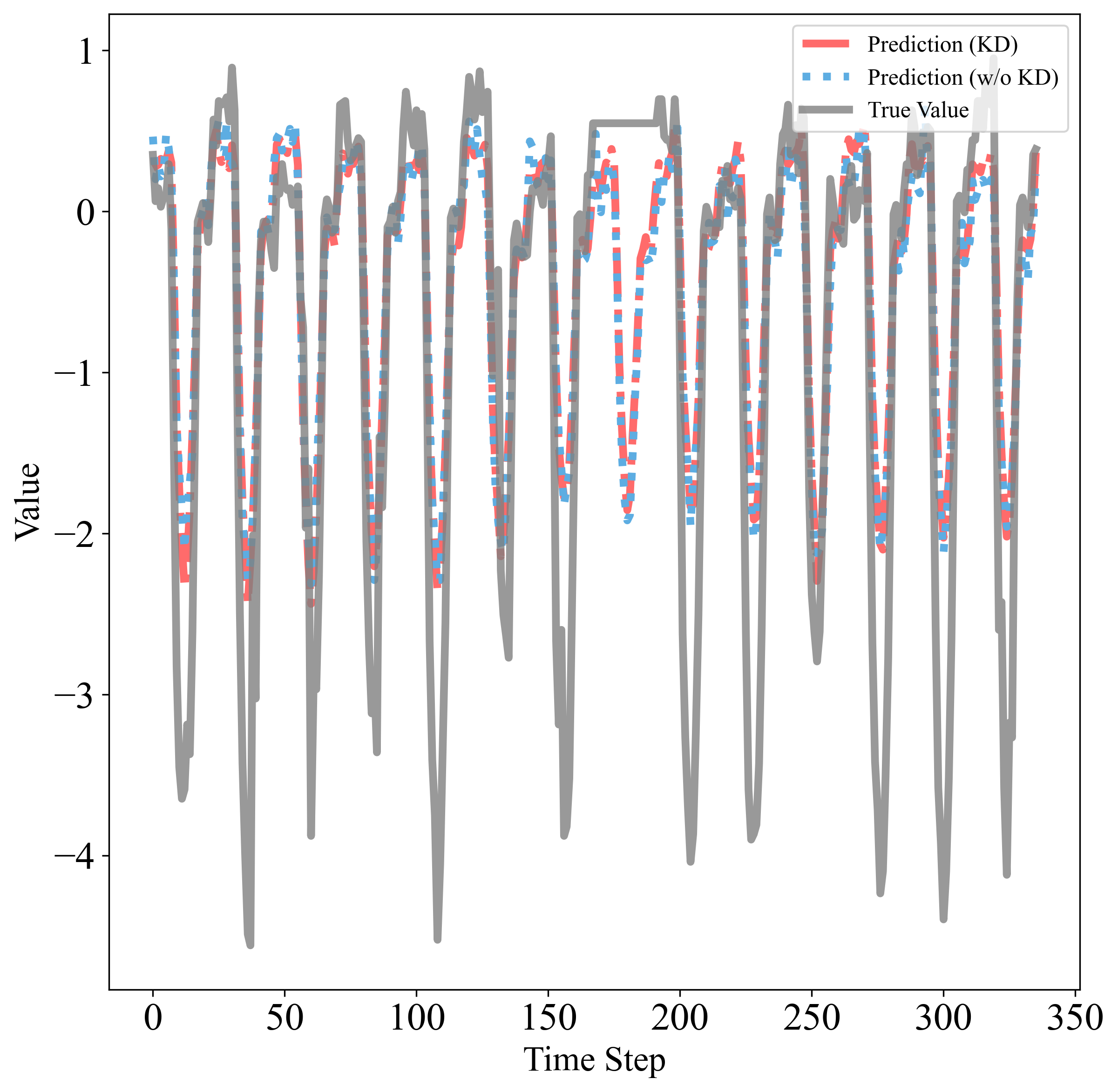}
    \caption{ETTh1}
    \label{fig:etth1_336}
  \end{subfigure}%
  \hfill
  \begin{subfigure}[b]{0.25\textwidth}
    \centering
    \includegraphics[width=\linewidth]{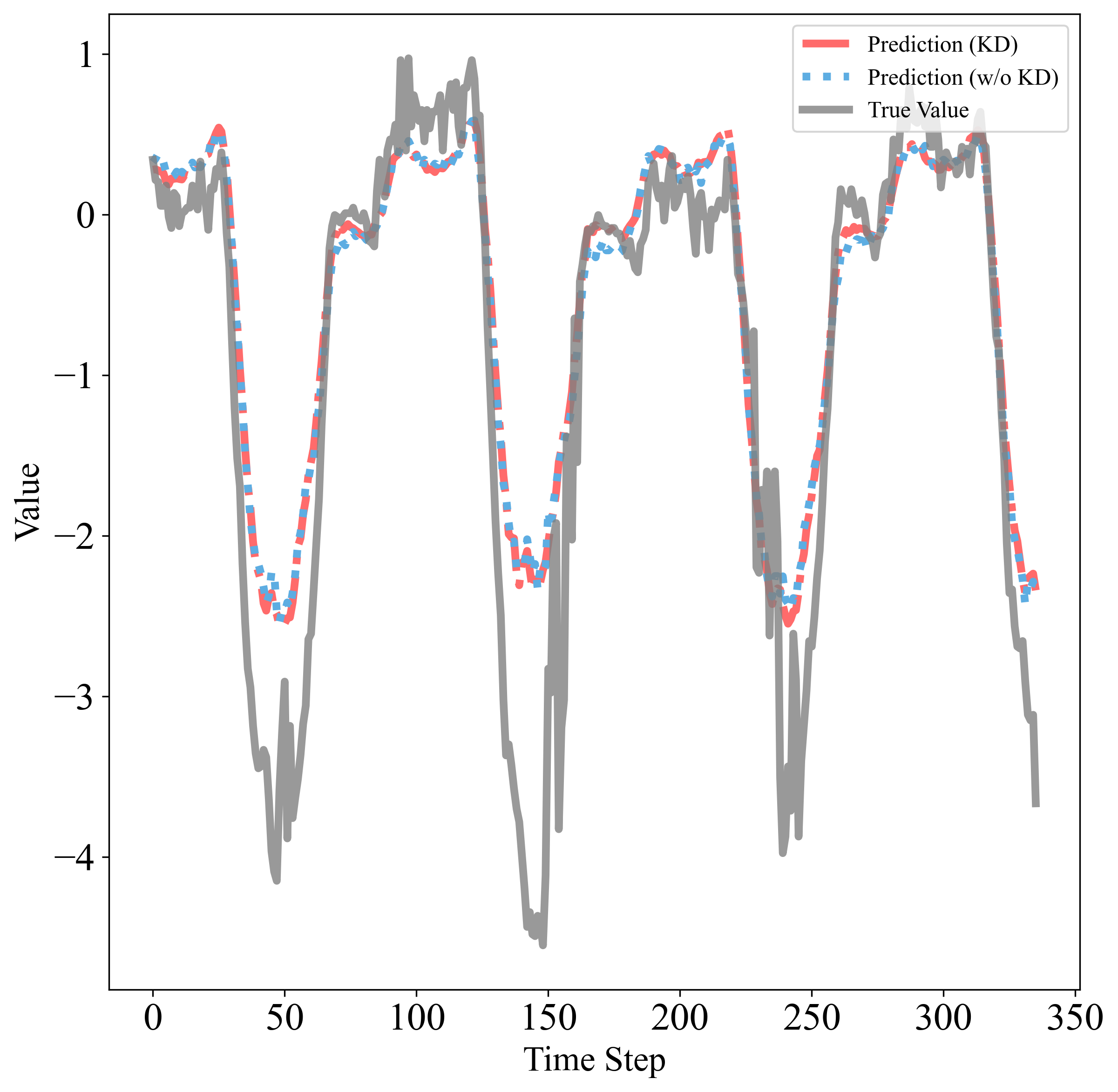}
    \caption{ETTm1}
    \label{fig:ettm1_336}
  \end{subfigure}%
  \hfill
  \begin{subfigure}[b]{0.25\textwidth}
    \centering
    \includegraphics[width=\linewidth]{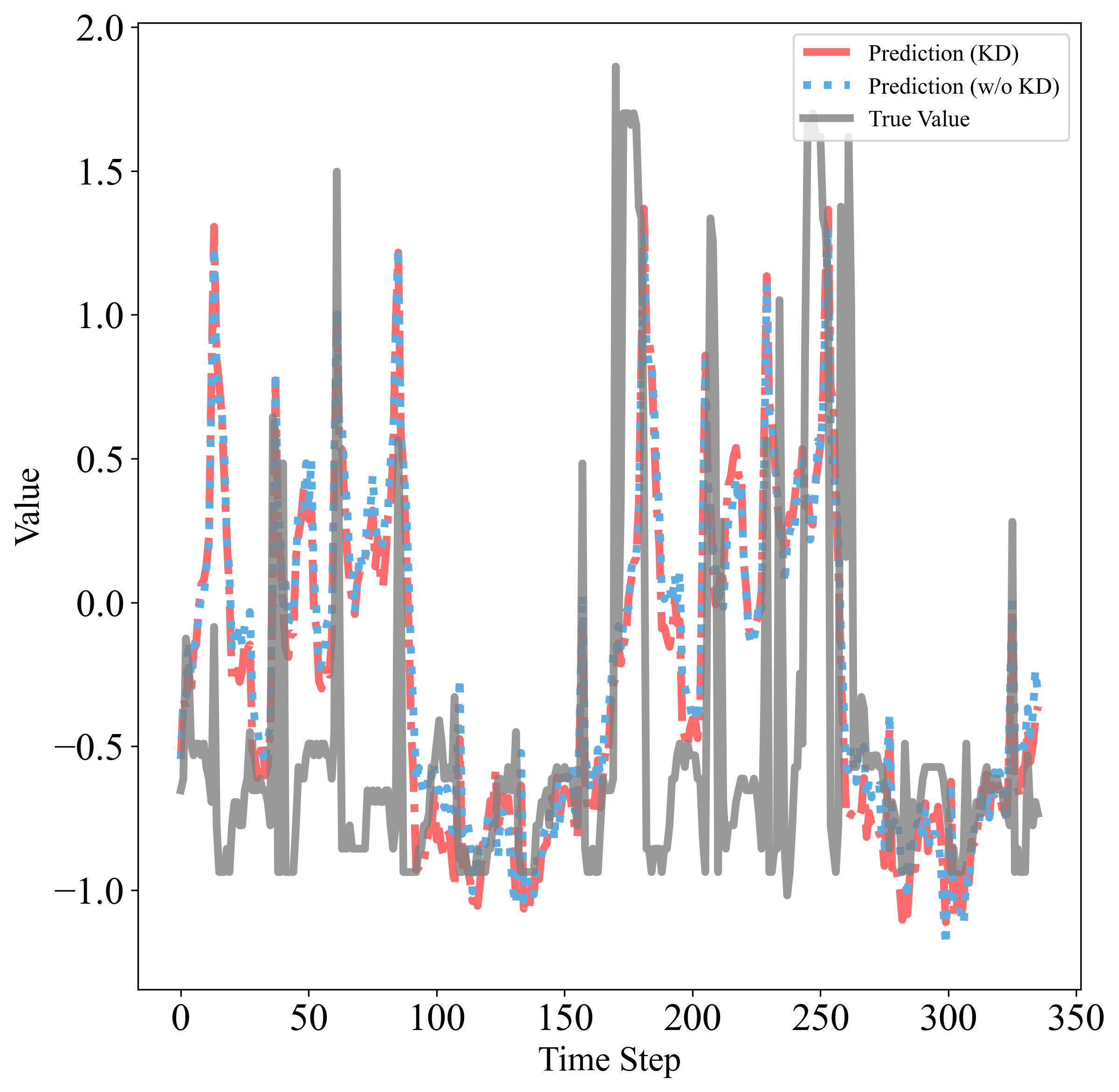}
    \caption{ECL}
    \label{fig:ecl_336}
  \end{subfigure}%
  \hfill
  \begin{subfigure}[b]{0.25\textwidth}
    \centering
    \includegraphics[width=\linewidth]{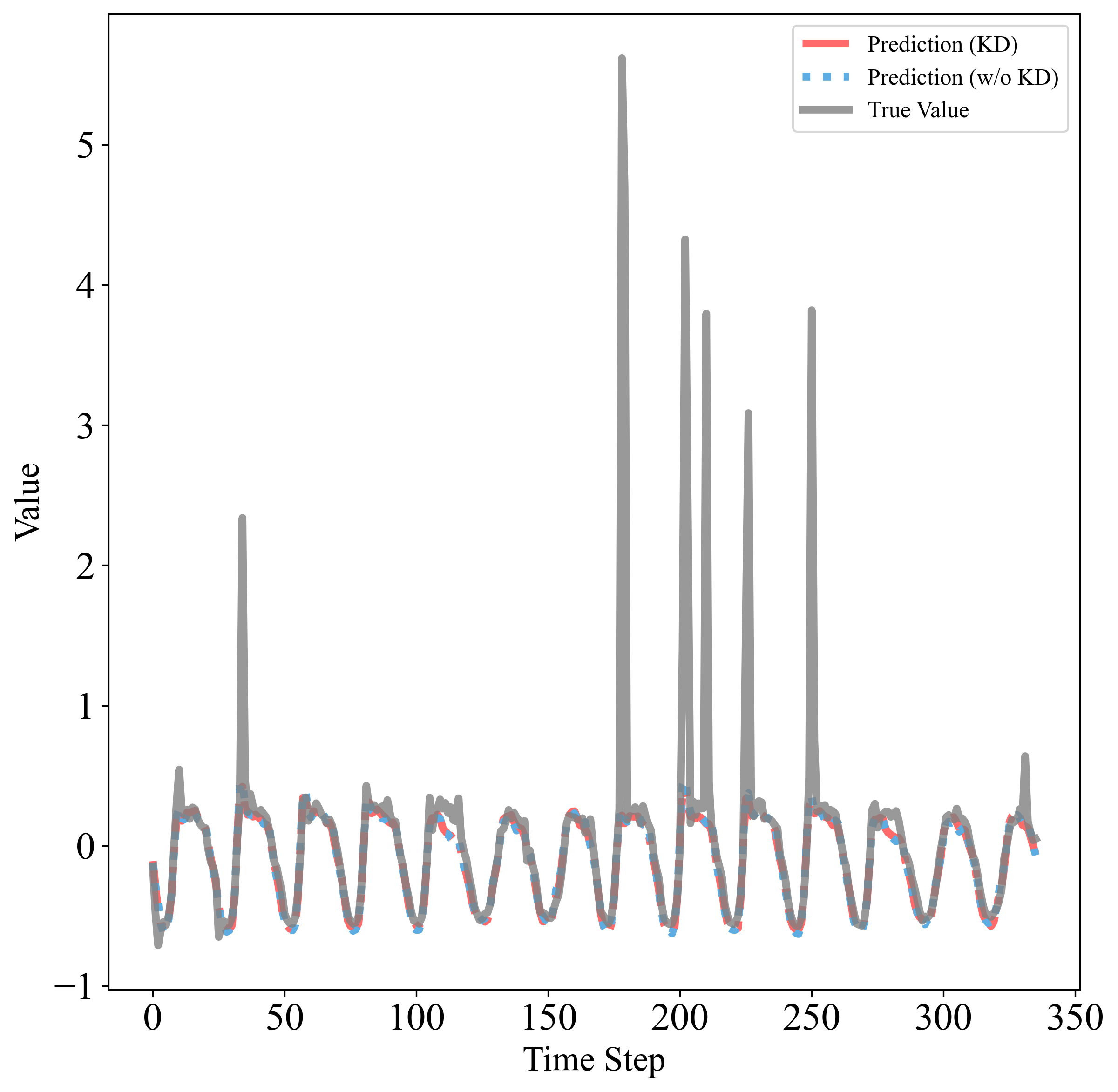}
    \caption{Traffic}
    \label{fig:traffic_336}
  \end{subfigure}
  \caption{Prediction results visualization for ETTh1, ETTm1, ECL, and Traffic datasets at 336 prediction lengths. The solid gray line shows the true values, the dashed red line shows the model’s predictions with knowledge distillation (KD), and the dotted blue line shows the teacher model's predictions without distillation (w/o KD).}
  \label{fig:visualization-336}
\end{figure}

\begin{figure}[!h]
  \centering
  \begin{subfigure}[b]{0.25\textwidth}
    \centering
    \includegraphics[width=\linewidth]{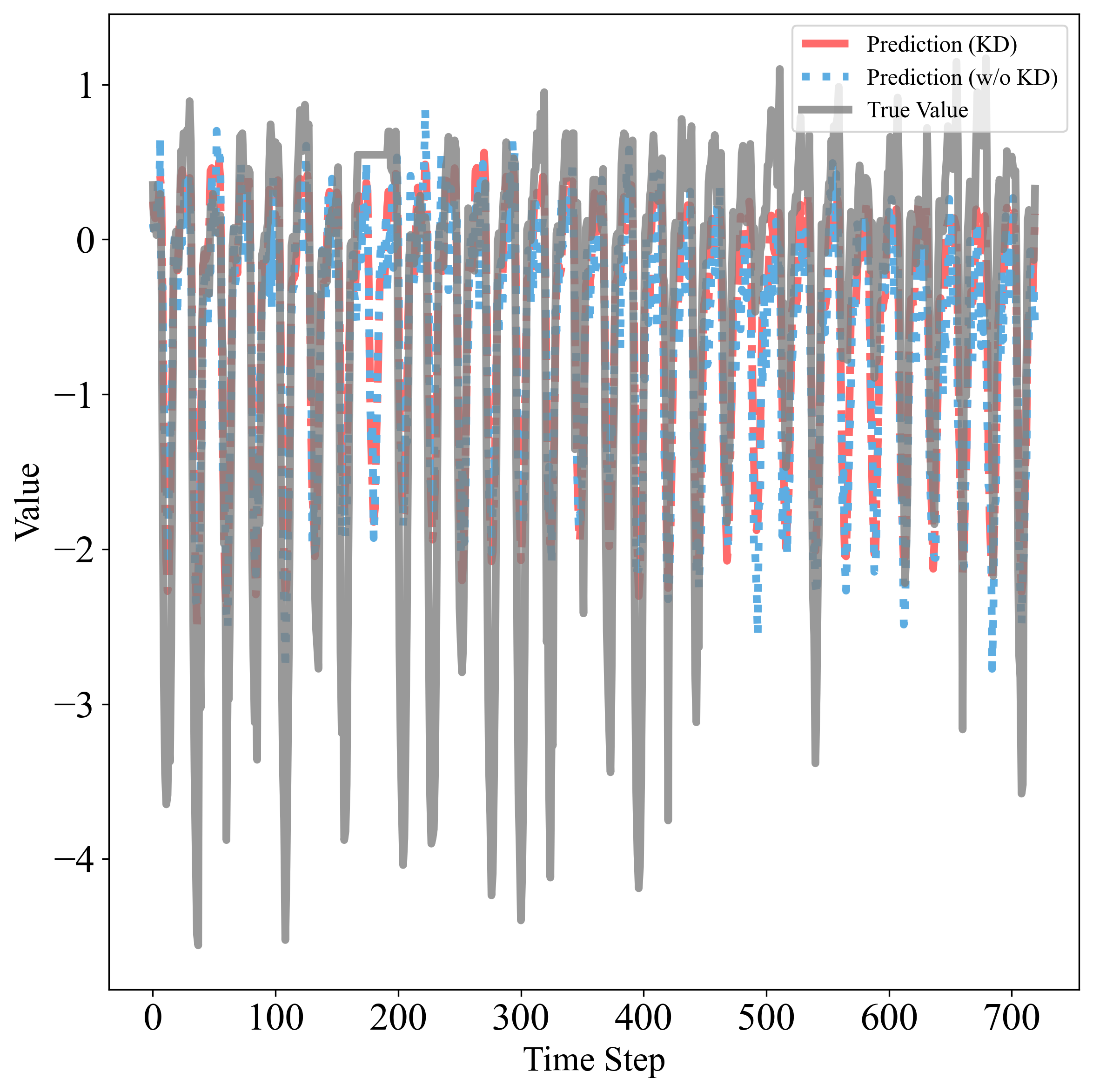}
    \caption{ETTh1}
    \label{fig:etth1_720}
  \end{subfigure}%
  \hfill
  \begin{subfigure}[b]{0.25\textwidth}
    \centering
    \includegraphics[width=\linewidth]{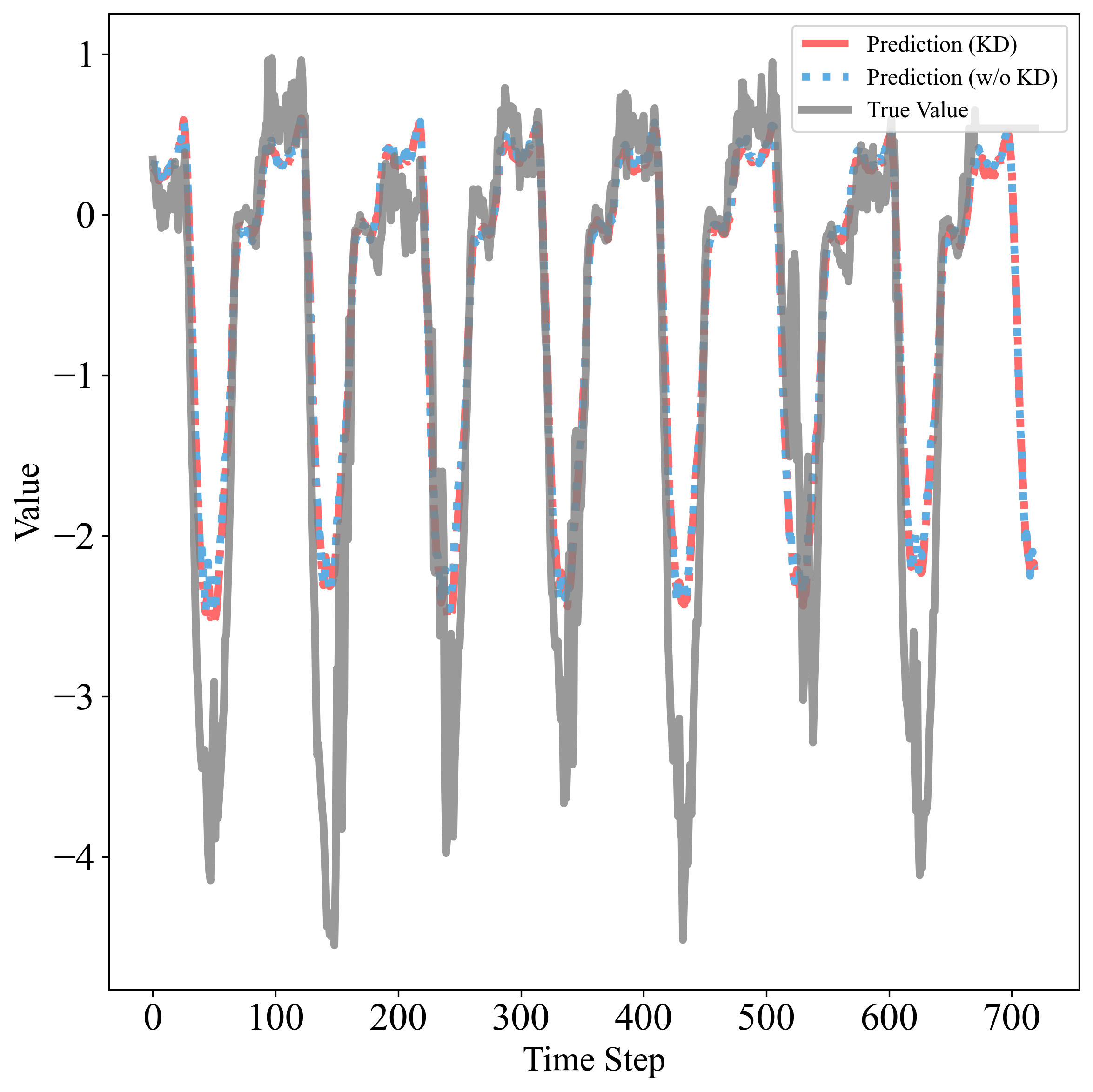}
    \caption{ETTm1}
    \label{fig:ettm1_720}
  \end{subfigure}%
  \hfill
  \begin{subfigure}[b]{0.25\textwidth}
    \centering
    \includegraphics[width=\linewidth]{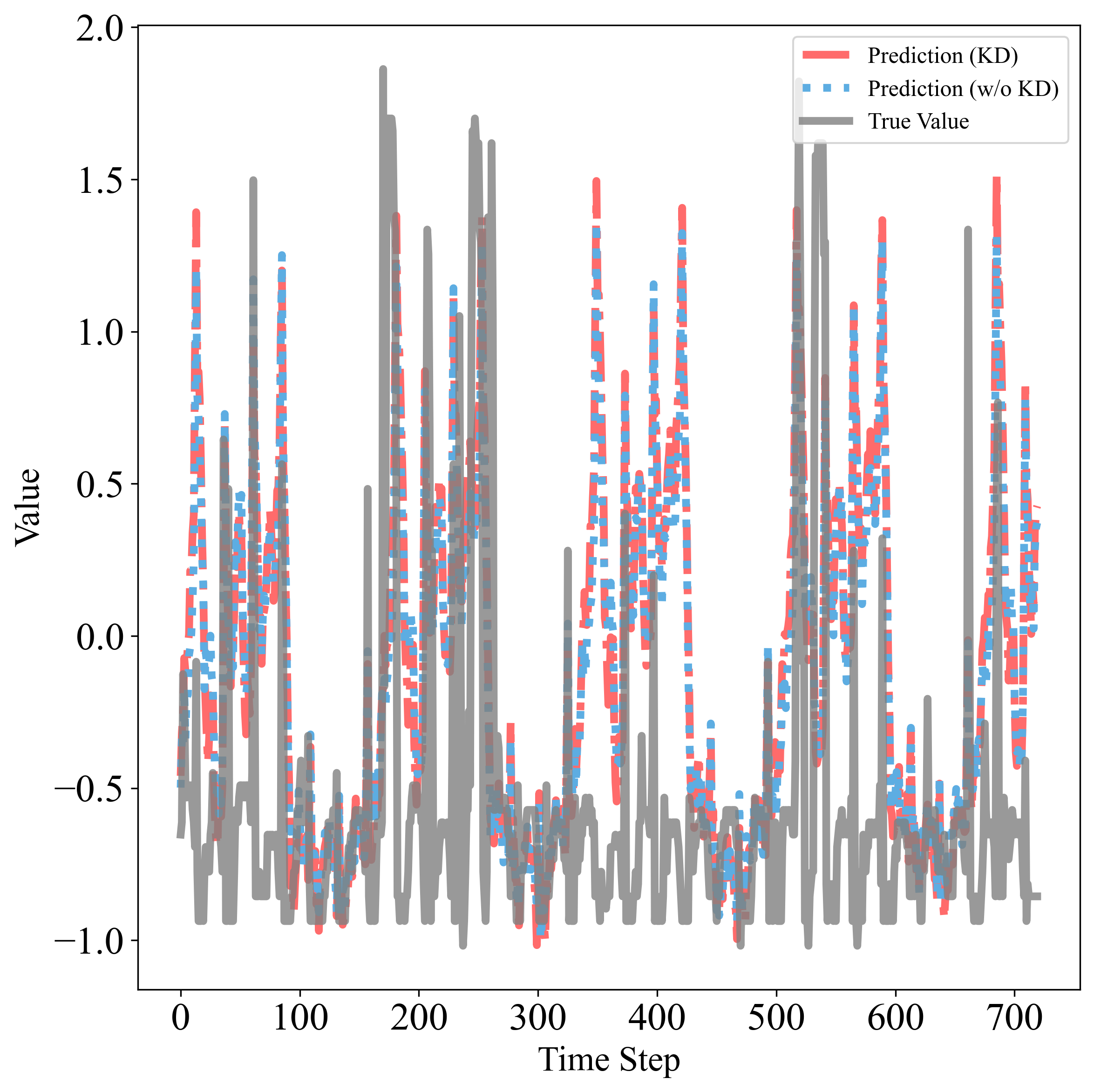}
    \caption{ECL}
    \label{fig:ecl_720}
  \end{subfigure}%
  \hfill
  \begin{subfigure}[b]{0.25\textwidth}
    \centering
    \includegraphics[width=\linewidth]{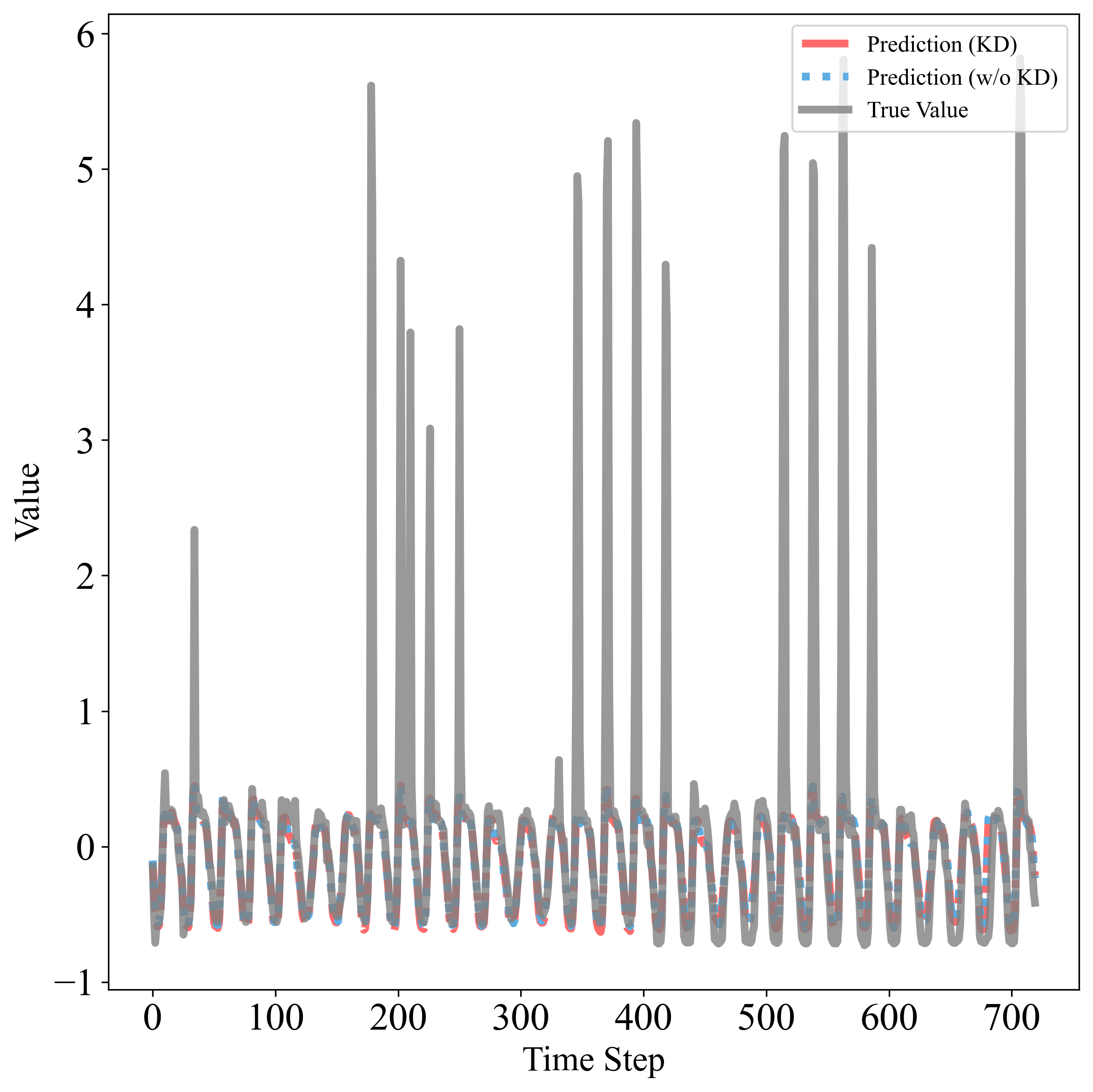}
    \caption{Traffic}
    \label{fig:traffic_720}
  \end{subfigure}
  \caption{Prediction results visualization for ETTh1, ETTm1, ECL, and Traffic datasets at 720 prediction lengths. The solid gray line shows the true values, the dashed red line shows the model’s predictions with knowledge distillation (KD), and the dotted blue line shows the teacher model's predictions without distillation (w/o KD).}
  \label{fig:visualization-720}
\end{figure}

Figures~\ref{fig:visualization-96}, \ref{fig:visualization-192}, \ref{fig:visualization-336}, \ref{fig:visualization-720} present the prediction results for horizons of 96, 192, 336, and 720 steps, comparing the knowledge distillation (KD) model with the teacher model without distillation (w/o KD). Across four datasets (ETTh1, ETTm1, ECL, Traffic), the KD-enhanced model (red dashed line) more closely follows the overall trend of the ground truth (gray solid line) compared to the teacher model (blue dash-dot line), particularly at critical inflection points and abrupt changes, where its fluctuations are markedly lower. This demonstrates that knowledge distillation not only achieves dramatic parameter reduction but also improves forecasting accuracy through effective knowledge transfer. Although error accumulation intensifies with longer horizons for both models, the KD model consistently exhibits smaller deviations and smoother curves, validating that the distillation mechanism successfully extracts and enhances the essential temporal patterns while eliminating redundant parameters, resulting in superior performance across short-, medium-, and long-term forecasting tasks.

\newpage
\section{Visualization of Time Series and ImageNet Feature Similarity}
\label{appx:visualization-of-Time-Series-and-ImageNet-Feature-Similarity}

\begin{figure}[!h]
\centering
\includegraphics[width=1\textwidth]{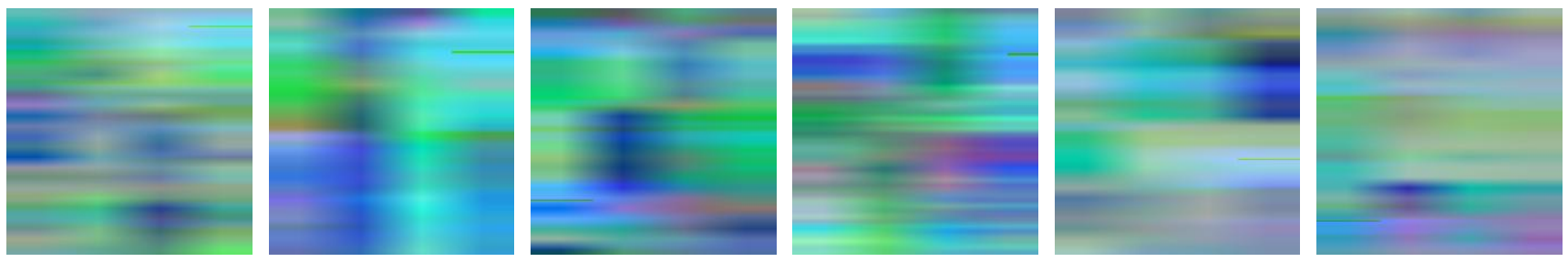}
\caption{Visual augmentations of time series data from ETT datasets generated by our visual augmentation module. These images represent the transformed temporal patterns used as input to the vision encoders.}
\label{fig:ts_visual_aug}
\end{figure}

\begin{figure}[!h]
\centering
\includegraphics[width=1\textwidth]{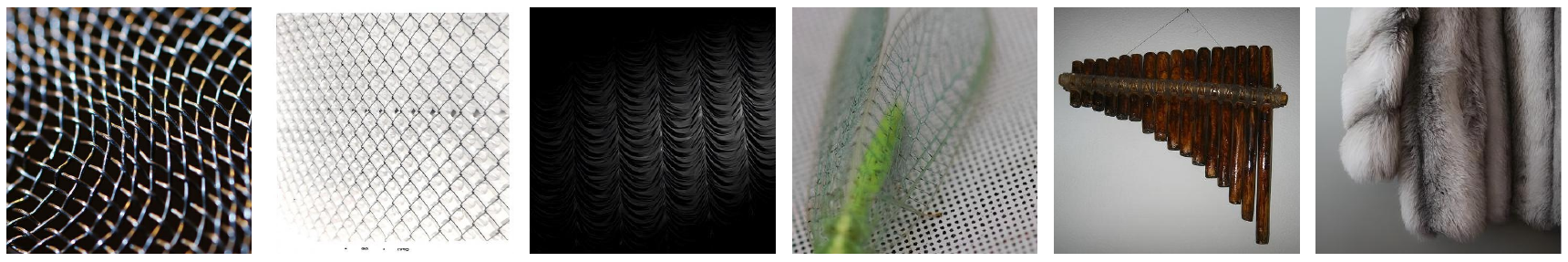}
\caption{ImageNet images with highest feature similarity to time series representations. These nearest neighbors predominantly consist of texture-rich patterns including meshes, grids, curtains, and repetitive structures, demonstrating that time series features naturally align with low-level visual patterns.}
\label{fig:imagenet_nearest}
\end{figure}

\begin{figure}[!h]
\centering
\includegraphics[width=1\textwidth]{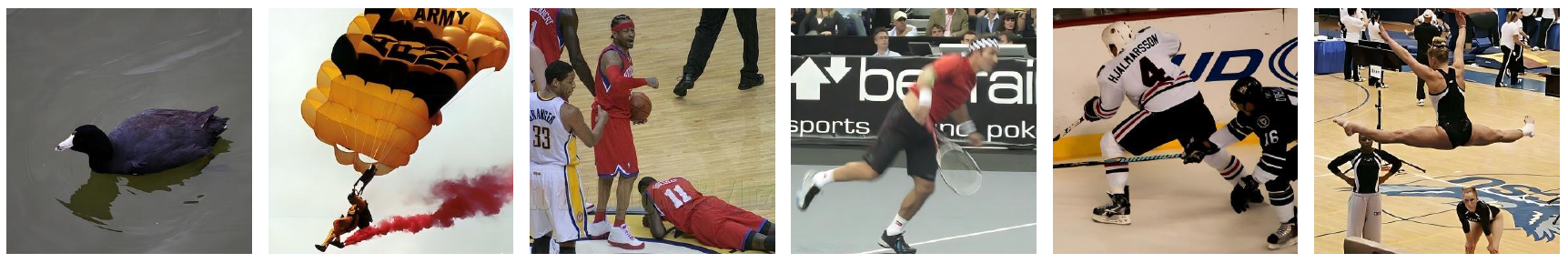}
\caption{ImageNet images with lowest feature similarity to time series representations. These distant samples contain semantically complex scenes with objects, animals, and human activities, confirming that high-level semantic features are largely irrelevant to time series forecasting tasks.}
\label{fig:imagenet_farthest}
\end{figure}

Figures~\ref{fig:ts_visual_aug}-\ref{fig:imagenet_farthest} provide compelling visual evidence for the modality gap analysis presented in Section 1. The visual augmentations in Figure~\ref{fig:ts_visual_aug} reveal how temporal patterns manifest as texture-like visual representations with gradient variations, repetitive structures, and periodic patterns. When comparing these augmentations with their feature space neighbors, a clear pattern emerges: Figure~\ref{fig:imagenet_nearest} shows that time series representations consistently align with images containing regular textures, geometric patterns, and structural repetitions—features that vision models capture in their early layers through edge detectors and texture filters. In contrast, Figure~\ref{fig:imagenet_farthest} demonstrates that semantically rich images requiring high-level understanding (object recognition, scene interpretation, activity classification) show minimal alignment with time series features. This visual analysis directly validates our core hypothesis that pre-trained vision models carry substantial redundancy for time series tasks—their deep semantic layers, which constitute the majority of parameters, contribute little to temporal pattern recognition. The clear separation between texture-aligned and semantic-divergent samples justifies our knowledge distillation approach that selectively transfers low-level visual knowledge while discarding high-level semantic parameters, enabling us to achieve superior forecasting performance with 99\% fewer parameters.

\section{Details of Cross-Modal Representation Module}
\label{appendix:cross_modal}

This appendix provides comprehensive details of the Cross-Modal Representation Module described in Section 3.1 of the main paper. We elaborate on the mathematical formulations and implementation specifics of both the temporal feature extraction and visual augmentation components that were briefly introduced in the main text.

\subsection{Temporal Feature Extraction}
\label{appendix:temporal_extraction}

This section expands on the temporal feature extraction mechanism outlined in the main paper. Our temporal feature extraction follows a transformer-based architecture designed to capture both local and global dependencies in multivariate time series. Given an input sequence $\mathbf{x}_{1:T} \in \mathbb{R}^{T \times C}$ with $T$ time steps and $C$ variables, we employ a patch-based approach to efficiently model temporal dynamics.

The input sequence is first segmented into overlapping patches of length $L$, where each patch $\mathbf{p}_i \in \mathbb{R}^{L \times C}$ is projected into a $d$-dimensional embedding space through a learnable linear transformation:
\begin{equation}
\mathbf{e}_i = \mathbf{W}_{\text{embed}} \cdot \text{flatten}(\mathbf{p}_i) + \mathbf{b}_{\text{embed}}
\end{equation}

To preserve temporal ordering, we add sinusoidal positional encodings to each patch embedding. The resulting embeddings $\{\mathbf{e}_i\}_{i=1}^{N}$ are then processed through $L_{\text{enc}}$ transformer encoder layers:
\begin{equation}
\mathbf{h}^{(\ell)} = \text{TransformerBlock}(\mathbf{h}^{(\ell-1)}), \quad \ell = 1, \ldots, L_{\text{enc}}
\end{equation}

where each transformer block consists of multi-head self-attention and feed-forward networks with residual connections. The final temporal representation $\mathbf{h}_T \in \mathbb{R}^{B \times d_{\text{model}}}$ captures hierarchical temporal patterns essential for accurate forecasting.

\subsection{Visual Augmentation for Time Series}
\label{appendix:visual_augmentation}

This section provides detailed implementation of the visual augmentation pipeline introduced in Section 3.1 of the main paper. Our empirical analysis (Figure 1 in main paper) reveals that time series features align with texture-rich visual patterns while diverging from semantic content. This insight motivates a specialized transformation that converts time series into 2D representations emphasizing low-level visual features relevant to temporal dynamics.

Specifically, our visual augmentation pipeline is designed to explicitly encode key temporal dynamics—such as frequency, periodicity, and local trends—into the visual language of textures, repetitive structures, and gradient changes in a 2D image. This approach directly aligns with our observation of LVMs' strengths in TSF, which is their capacity for low-level feature extraction rather than high-level semantic understanding. Compared to other visualization methods, our use of multi-scale convolutions provides a more flexible mechanism to capture complex, composite patterns across different time scales, from local variations to global trends.

The visual augmentation pipeline consists of three key stages:

\smallskip
\noindent\textbf{Pattern Enhancement} 

We first augment the time series with features that highlight temporal characteristics. Specifically, we apply two complementary encoding techniques:

1) \textit{Frequency Encoding.} A Fast Fourier Transform (FFT) extracts frequency components:
\begin{equation}
\text{FFT}(x_{\text{enc}}) = \Biggl\lvert\sum_{t=0}^{L-1} x_{\text{enc}}(t) \cdot e^{-2\pi i k t / L}\Biggl\lvert
\end{equation}
where $k$ is the frequency index. The magnitude of FFT coefficients reveals dominant frequencies and periodic patterns in the data.

2) \textit{Periodicity Encoding.} Temporal cycles are captured using trigonometric functions:
\begin{equation}
\text{PE}(t) = \left[ \sin\left(\frac{2\pi t}{P}\right), \cos\left(\frac{2\pi t}{P}\right) \right]
\end{equation}
where $P$ is the dataset-specific periodicity parameter. These encodings are concatenated with the original time series and frequency features, resulting in an augmented tensor $\mathbf{X}_{\text{aug}} \in \mathbb{R}^{B \times L \times D \times 3}$.

\smallskip
\noindent\textbf{Multi-Scale Transformation} 

The augmented features undergo hierarchical processing through multiple convolutional layers. A 1D convolutional layer first captures local dependencies:
\begin{equation}
\mathbf{F}_{\text{1D}} = \text{Conv1D}(\mathbf{X}_{\text{aug}}) \in \mathbb{R}^{B \times D \times H_{\text{hidden}} \times L}
\end{equation}
where $H_{\text{hidden}}$ is the hidden dimension. After averaging along the variable dimension $D$, two subsequent 2D convolutional layers progressively refine the features: the first halves the channel dimension, and the second maps to $C$ output channels, producing multi-scale representations that capture both local variations and global trends.

\smallskip
\noindent\textbf{Spatial Reorganization}

The multi-scale features are transformed into a 2D image format through bilinear interpolation. For each target pixel $(x, y)$ in the output image:
\begin{equation}
\mathbf{I}(x, y) = \sum_{i=1}^{2} \sum_{j=1}^{2} \mathbf{I}(x_i, y_j) \cdot w_{ij}
\end{equation}
where $(x_i, y_j)$ are the four nearest neighbors and $w_{ij}$ are distance-based weights. The final normalization ensures compatibility with vision encoders:
\begin{equation}
\mathbf{I}_{\text{visual}} = 255 \cdot \frac{\mathbf{I}_{\text{raw}} - \text{Min}(\mathbf{I}_{\text{raw}})}{\text{Max}(\mathbf{I}_{\text{raw}}) - \text{Min}(\mathbf{I}_{\text{raw}}) + \epsilon}
\end{equation}
where $\epsilon = 10^{-5}$ prevents division by zero. This produces $\mathbf{I}_{\text{visual}} \in \mathbb{R}^{B \times C \times H \times W}$ with pixel values in $[0, 255]$.

This visual representation serves a dual purpose: it reveals temporal patterns invisible in 1D sequences (e.g., trend changes appear as edge-like features) and provides a format compatible with pre-trained vision models. Crucially, by focusing on texture and gradient patterns rather than semantic content, we align with the 1\% of vision model capabilities actually useful for time series analysis.

\section{Details of Adaptive Knowledge Distillation Framework}
\label{appendix:adaptive_kd}

This appendix provides the complete mathematical formulation and implementation details of the Adaptive Knowledge Distillation (AKD) framework introduced in Section 3.3 of the main paper. Unlike conventional knowledge distillation approaches that rely on manually tuned fixed hyperparameters, our framework treats distillation weights, temperature parameters, and feature alignment strategies as learnable components that adapt dynamically to the specific characteristics of teacher-student model pairs and target datasets.

\subsection{Learnable Weight Optimization Module}
\label{appendix:learnable_weights}
\smallskip
\noindent\textbf{Dynamic Loss Weight Learning}  

As briefly mentioned in the main paper, traditional knowledge distillation frameworks suffer from the challenge of manually balancing multiple loss components. To address this limitation, we introduce a learnable weight optimization mechanism that automatically discovers optimal loss combinations during training.

\textit{Parameterized Weight Representation.} We parameterize the distillation weights using log-space representation to ensure positivity constraints:
\begin{equation}
\label{eq:weight_parameterization}
\boldsymbol{w} = \exp(\boldsymbol{\theta}_w) \in \mathbb{R}^4_+
\end{equation}
where $\boldsymbol{\theta}_w = [\theta_{\text{fd}}, \theta_{\text{fcst}}, \theta_{\text{recon}}, \theta_{\text{cd}}]$ represents the learnable log-weights for feature distillation, forecasting task, reconstruction, and correlation distillation losses respectively.

\textit{Weight Initialization and Constraints.} The log-weights are initialized based on empirical values:
\begin{equation}
\label{eq:weight_initialization}
\boldsymbol{\theta}_w^{(0)} = \log([\lambda_{\text{fd}}^{(0)}, \lambda_{\text{fcst}}^{(0)}, \lambda_{\text{recon}}^{(0)}, \lambda_{\text{cd}}^{(0)}])
\end{equation}
where $[\lambda_{\text{fd}}^{(0)}, \lambda_{\text{fcst}}^{(0)}, \lambda_{\text{recon}}^{(0)}, \lambda_{\text{cd}}^{(0)}] = [0.01, 1.0, 0.5, 0.01]$ serve as initialization values.

\textit{Regularization Mechanism.} To prevent weight explosion and maintain training stability, we incorporate L2 regularization:
\begin{equation}
\label{eq:weight_regularization}
\mathcal{L}_{\text{reg}} = \gamma \sum_{i=1}^{4} w_i^2
\end{equation}
where $\gamma$ is the regularization coefficient and $w_i$ represents the $i$-th component of $\boldsymbol{w}$.

\smallskip
\noindent\textbf{Adaptive Temperature Control}

\textit{Sigmoid-Constrained Temperature Learning.} The distillation temperature parameter is crucial for controlling the softness of teacher distributions. We propose an adaptive temperature mechanism:
\begin{equation}
\label{eq:adaptive_temperature}
\tau = \tau_{\min} + (\tau_{\max} - \tau_{\min}) \cdot \sigma(\theta_{\tau})
\end{equation}
where $\sigma(\cdot)$ denotes the sigmoid function, $\theta_{\tau}$ is the learnable raw temperature parameter, and $[\tau_{\min}, \tau_{\max}] = [1.0, 10.0]$ define the temperature bounds.

\textit{Temperature Evolution Strategy.} The temperature parameter evolves during training to balance between preserving teacher knowledge (high temperature) and maintaining prediction sharpness (low temperature):
\begin{equation}
\label{eq:temperature_gradient}
\frac{\partial \mathcal{L}}{\partial \theta_{\tau}} = \frac{\partial \mathcal{L}}{\partial \tau} \cdot \frac{\partial \tau}{\partial \theta_{\tau}}
\end{equation}
where the gradient flows enable automatic temperature adjustment based on distillation effectiveness.

\subsection{Pyramid-Style Feature Alignment Architecture}
\label{appendix:multiscale_alignment}
\smallskip
\noindent\textbf{Hierarchical Projection Strategy}

To address the dimensional mismatch between teacher and student representations, we develop a pyramid-style feature alignment mechanism that captures information at different granularities.

\textit{Pyramid-Style Projection Functions.} Let $\mathbf{F}^S_{\text{fus}} \in \mathbb{R}^{B \times d_s}$ and $\mathbf{F}^T_{\text{fus}} \in \mathbb{R}^{B \times d_t}$ denote student and teacher fused feature representations respectively. The pyramid-style alignment is formulated as:
\begin{equation}
\label{eq:multiscale_alignment}
\mathbf{F}^S_{\text{aligned}} = \sum_{i=0}^{N_s} w_i \phi_i(\mathbf{F}^S_{\text{fus}})
\end{equation}
where $\phi_i: \mathbb{R}^{d_s} \rightarrow \mathbb{R}^{d_t}$ represents the $i$-th scale projection function, $N_s$ is the number of scales, and $\mathbf{w} = \text{softmax}(\boldsymbol{\beta})$ are learnable scale weights with $\boldsymbol{\beta} \in \mathbb{R}^{N_s+1}$.

\textit{Scale-Specific Projections.} Each projection function operates at different hidden dimensions:
\begin{equation}
\label{eq:scale_projection}
\phi_i(\mathbf{x}) = \mathbf{W}_i^{(2)} \sigma(\mathbf{W}_i^{(1)} \mathbf{x} + \mathbf{b}_i^{(1)}) + \mathbf{b}_i^{(2)}
\end{equation}
where $\mathbf{W}_i^{(1)} \in \mathbb{R}^{h_i \times d_s}$, $\mathbf{W}_i^{(2)} \in \mathbb{R}^{d_t \times h_i}$, and $h_i = \max(d_s, d_t) / 2^i$ represents the hidden dimension for scale $i$.

\textit{Attention-Weighted Integration.} The final aligned features incorporate attention mechanisms to emphasize the most relevant scale combinations:
\begin{equation}
\label{eq:attention_integration}
\mathbf{A} = \text{softmax}\left(\frac{\mathbf{Q}\mathbf{K}^T}{\sqrt{d_k}}\right)
\end{equation}

where $\mathbf{Q} = \mathbf{F}^S_{\text{fus}}\mathbf{W}_Q$, $\mathbf{K} = \mathbf{F}^T_{\text{fus}}\mathbf{W}_K$, and the attention weights guide the scale selection process.

\subsection{Adaptive Loss Balancing Mechanism}
\label{appendix:loss_balancing}

\noindent\textbf{Exponential Moving Average Loss Tracking}

To maintain balanced contributions from different loss components, we implement an adaptive loss balancing mechanism based on exponential moving averages:
\begin{equation}
\label{eq:ema_loss_tracking}
\bar{\mathcal{L}}_i^{(t)} = \mu \bar{\mathcal{L}}_i^{(t-1)} + (1-\mu) |\mathcal{L}_i^{(t)}|
\end{equation}
where $\bar{\mathcal{L}}_i^{(t)}$ represents the moving average of the $i$-th loss component at iteration $t$, and $\mu = 0.9$ is the momentum parameter.

\textit{Scale Normalization.} The loss components are normalized using their running scales to prevent dominance by any single term:
\begin{equation}
\label{eq:scale_normalization}
\mathcal{L}_i^{\text{norm}} = \frac{\mathcal{L}_i}{\bar{\mathcal{L}}_i + \epsilon}
\end{equation}
where $\epsilon = 1e-8$ is a small constant for numerical stability.

\subsection{Comprehensive Distillation Objective}
\label{appendix:distillation_objective}
\smallskip
\noindent\textbf{Multi-Component Loss Integration}

The total distillation loss combines multiple knowledge transfer mechanisms with learnable weights:
\begin{equation}
\label{eq:total_distillation_loss}
\mathcal{L}_{\text{distill}} = w_1 \mathcal{L}_{\text{fd}} + w_2 \mathcal{L}_{\text{fcst}} + w_3 \mathcal{L}_{\text{recon}} + w_4 \mathcal{L}_{\text{cd}} + \mathcal{L}_{\text{reg}}
\end{equation}

\textit{Correlation Distillation Loss.} Aligns temporal dependency patterns between teacher and student models:
\begin{equation}
\label{eq:correlation_distillation_loss}
\mathcal{L}_{\text{cd}} = \frac{\tau^2}{B} \sum_{i=1}^{B} D_\text{KL}\left(\text{softmax}\left(\frac{\mathbf{P}_{\text{tea}}^{(i)}}{\tau}\right) \| \text{softmax}\left(\frac{\mathbf{P}_{\text{stu}}^{(i)}}{\tau}\right)\right)
\end{equation}
where $\mathbf{P}_{\text{tea}}^{(i)}$ and $\mathbf{P}_{\text{stu}}^{(i)}$ represent the attention matrices for the $i$-th sample in teacher and student models respectively, capturing temporal dependencies as described in the main paper equation (\ref{eq:correlation_distillation}).

\textit{Feature Distillation Loss.} Captures representation-level knowledge transfer using multiple similarity metrics, as introduced in equation (\ref{Lfd}) of the main paper:
\begin{equation}
\label{eq:feature_distillation_loss}
\mathcal{L}_{\text{fd}} = \lambda_{\text{MSE}} \|\mathbf{F}^S_{\text{aligned}} - \mathbf{F}^T_{\text{fus}}\|_2^2 + \lambda_{\text{cos}} \mathcal{L}_{\text{cosine}} + \lambda_{\text{KL}} \mathcal{L}_{\text{KL}}
\end{equation}
where:
\begin{align}
\mathcal{L}_{\text{cosine}} &= 1 - \frac{\mathbf{F}^S_{\text{aligned}} \cdot \mathbf{F}^T_{\text{fus}}}{\|\mathbf{F}^S_{\text{aligned}}\| \|\mathbf{F}^T_{\text{fus}}\|} \label{eq:cosine_loss} \\
\mathcal{L}_{\text{KL}} &= D_\text{KL}\left(\text{softmax}\left(\frac{\mathbf{F}^T_{\text{fus}}}{\tau}\right) \| \text{softmax}\left(\frac{\mathbf{F}^S_{\text{aligned}}}{\tau}\right)\right) \label{eq:kl_loss}
\end{align}

\smallskip
\noindent\textbf{Dual-Optimizer Training Strategy}

\textit{Separate Learning Rate Scheduling.} We employ two optimizers with different learning rates to accommodate the varying convergence characteristics:
\begin{itemize}
    \item \textit{Model Optimizer:} $\eta_{\text{model}}$ for backbone parameters
    \item \textit{Distillation Optimizer:} $\eta_{\text{distill}} = \rho \cdot \eta_{\text{model}}$ for distillation parameters
\end{itemize}
where $\rho = 0.1$ represents the learning rate ratio.

\textit{Gradient Update Protocol.} The parameter updates follow:
\begin{align}
\boldsymbol{\theta}_{\text{model}}^{(t+1)} &= \boldsymbol{\theta}_{\text{model}}^{(t)} - \eta_{\text{model}} \nabla_{\boldsymbol{\theta}_{\text{model}}} \mathcal{L}_{\text{total}} \label{eq:model_update} \\
\boldsymbol{\theta}_{\text{distill}}^{(t+1)} &= \boldsymbol{\theta}_{\text{distill}}^{(t)} - \eta_{\text{distill}} \nabla_{\boldsymbol{\theta}_{\text{distill}}} \mathcal{L}_{\text{distill}} \label{eq:distill_update}
\end{align}
where $\boldsymbol{\theta}_{\text{distill}} = [\boldsymbol{\theta}_w, \theta_{\tau}, \boldsymbol{\beta}]$ encompasses all learnable distillation parameters.

\subsection{Real-Time Monitoring and Adaptation}
\label{appendix:monitoring}
\smallskip
\noindent\textbf{Performance Tracking Metrics}

The framework continuously monitors distillation effectiveness through several indicators:
\begin{equation}
\label{eq:monitoring_metrics}
\mathcal{M}_{\text{eff}} = \left\{\frac{\partial \mathcal{L}_{\text{fcst}}}{\partial \mathcal{L}_{\text{fd}}}, \frac{\partial \mathcal{L}_{\text{fcst}}}{\partial \tau}, \|\boldsymbol{w}^{(t)} - \boldsymbol{w}^{(t-1)}\|_2\right\}
\end{equation}

\textit{Convergence Criteria.} The adaptation process stabilizes when:
\begin{equation}
\label{eq:convergence_criteria}
\frac{1}{K} \sum_{k=t-K+1}^{t} \|\boldsymbol{w}^{(k)} - \boldsymbol{w}^{(k-1)}\|_2 < \epsilon_{\text{conv}}
\end{equation}
where $K = 10$ represents the monitoring window and $\epsilon_{\text{conv}} = 1e-4$ is the convergence threshold.

This adaptive framework eliminates the need for extensive hyperparameter search while ensuring optimal knowledge transfer efficiency across diverse teacher-student configurations and datasets. The learnable components automatically discover task-specific distillation strategies, leading to improved performance and reduced manual tuning overhead.
\section{Future Work}
\label{appx:future_work}
While OccamVTS successfully demonstrates that 99\% of vision model parameters are redundant for time series forecasting, this finding opens several compelling avenues for future research. We identify four key directions that could further advance cross-modal knowledge distillation and expand its applicability to broader temporal modeling challenges.

\begin{itemize}[leftmargin=*, itemsep=0pt]
    \item \textbf{Integration with Foundation Models.} As multi-modal foundation models become increasingly prevalent, adapting OccamVTS to distill temporal knowledge from these massive architectures presents both opportunities and challenges. Future work should investigate how to extract the minimal sufficient knowledge from billion-parameter models while preserving their zero-shot generalization capabilities. This includes developing prompt-based distillation techniques that enable task-specific knowledge extraction without fine-tuning, and exploring whether the 1\% retention principle scales to foundation model architectures. Successfully addressing this challenge could democratize access to foundation model capabilities for time series applications on resource-constrained devices.
    \item \textbf{Expert-Mixture Distillation Framework.} Rather than distilling from a single vision teacher, future research could explore an ensemble approach where multiple domain-specific visual experts (e.g., satellite imagery for climate data, medical imaging for healthcare time series) provide specialized supervision. This would involve developing adaptive gating mechanisms to automatically select relevant experts based on time series characteristics, and investigating how to resolve potentially conflicting knowledge from different experts. Such an approach could achieve even higher compression ratios by focusing only on the most relevant visual patterns for each specific forecasting domain.
    \item \textbf{Multi-Modal Fusion Beyond Vision.} Extending OccamVTS to simultaneously distill knowledge from multiple pre-trained modalities—including language, audio, and sensor models—represents a natural evolution of our framework. This direction requires addressing fundamental challenges in aligning heterogeneous feature spaces and determining optimal modality-specific compression ratios. Developing unified distillation mechanisms that can selectively extract complementary knowledge from diverse modalities while maintaining computational efficiency could lead to richer temporal representations and more robust forecasting models.

\end{itemize}

These future directions collectively aim to push the boundaries of efficient cross-modal learning, from scaling to massive foundation models to incorporating diverse knowledge sources. By pursuing these research avenues, we envision a new generation of time series forecasting systems that combine the power of large-scale pre-training with the efficiency demanded by real-world applications, ultimately making advanced forecasting capabilities accessible across a broader range of computational environments.

\end{document}